\documentclass{article}

\usepackage{microtype}
\usepackage{graphicx}

\usepackage{booktabs} 

\usepackage{float} 
\usepackage{caption}
\usepackage{subcaption}
\usepackage{tcolorbox}
\usepackage{enumitem}

\usepackage{alphalph}
\usepackage{hyperref}

\usepackage[accepted]{icml2024}

\usepackage{amsmath}
\usepackage{amssymb}
\usepackage{mathtools}
\usepackage{amsthm}

\usepackage[capitalize,noabbrev]{cleveref}

\theoremstyle{plain}

\theoremstyle{definition}

\theoremstyle{remark}

\usepackage[textsize=tiny]{todonotes}

\icmltitlerunning{Overestimation, Overfitting, and Plasticity in Reinforcement Learning}

\begin{document}

\twocolumn[
\icmltitle{Overestimation, Overfitting, and Plasticity in Actor-Critic: \\ the Bitter Lesson of Reinforcement Learning}



\icmlsetsymbol{equal}{*}
\icmlsetsymbol{equalSenior}{$\dagger$}

\begin{icmlauthorlist}
\icmlauthor{Michal Nauman}{equal,ideas,uw}
\icmlauthor{Michał Bortkiewicz}{equal,pw}
\icmlauthor{Piotr Miłoś}{ideas,uw,pan}
\icmlauthor{Tomasz Trzciński}{ideas,pw,toop}
\icmlauthor{Mateusz Ostaszewski}{equalSenior,pw}
\icmlauthor{Marek Cygan}{equalSenior,uw,nomagic}
\end{icmlauthorlist}

\icmlaffiliation{ideas}{Ideas NCBR}
\icmlaffiliation{uw}{University of Warsaw}
\icmlaffiliation{pw}{Warsaw University of Technology}
\icmlaffiliation{nomagic}{Nomagic}
\icmlaffiliation{toop}{Tooploox}
\icmlaffiliation{pan}{Polish Academy of Sciences}

\icmlcorrespondingauthor{Michal Nauman}{nauman.mic@gmail.com}
\icmlcorrespondingauthor{Michal Bortkiewicz}{michalbortkiewicz8@gmail.com}

\icmlkeywords{Machine Learning, ICML}

\vskip 0.3in
]



\printAffiliationsAndNotice{* Main authors contributed equally to this work;
$\dagger$ Senior authors contributed equally to this work.} 

\begin{abstract} 
    Recent advancements in off-policy Reinforcement Learning (RL) have significantly improved sample efficiency, primarily due to the incorporation of various forms of regularization that enable more gradient update steps than traditional agents. However, many of these techniques have been tested in limited settings, often on tasks from single simulation benchmarks and against well-known algorithms rather than a range of regularization approaches. This limits our understanding of the specific mechanisms driving RL improvements. To address this, we implemented over 60 different off-policy agents, each integrating established regularization techniques from recent state-of-the-art algorithms. We tested these agents across 14 diverse tasks from 2 simulation benchmarks, measuring training metrics related to overestimation, overfitting, and plasticity loss — issues that motivate the examined regularization techniques. Our findings reveal that while the effectiveness of a specific regularization setup varies with the task, certain combinations consistently demonstrate robust and superior performance. Notably, a simple Soft Actor-Critic agent, appropriately regularized, reliably finds a better-performing policy within the training regime, which previously was achieved mainly through model-based approaches.
\end{abstract}

\section{Introduction}



In recent years, substantial improvements have been made in the domain of deep reinforcement learning, as evidenced by breakthroughs such as mastering complex games like Dota 2~\citep{openai2019dota}, Go~\citep{silver2017mastering} and achieving control over nuclear fusion plasma~\citep{Degrave2022plasma}. In particular, off-policy RL has witnessed a surge of approaches reporting state-of-the-art results \citep{li2022efficient, hafner2023mastering, lee2023plastic}, including application to real robots \citep{smith2022walk}. In general, those approaches build upon Soft Actor-Critic (SAC) algorithm with increased number of gradient steps per environment steps (Replay Ratio (RR)) used in conjunction with some form of regularization that stabilizes the learning in high RR setting \citep{janner2019trust, chen2020randomized, hiraoka2021dropout, nikishin2022primacy, li2022efficient, d2022sample, cetin2023learning}. These approaches for regularization encompass considerations of reducing overfitting \citep{li2022efficient} (\textit{network regularization}), reducing critic overestimation \citep{cetin2023learning} (\textit{critic regularization}) or reducing the rate of plasticity loss \citep{lee2023plastic} (\textit{plasticity regularization}).  




Despite significant advancements, the understanding of how different regularization techniques synergistically improve off-policy agent performance is still limited \citep{hiraoka2021dropout, lee2023plastic}. Moreover, most methods are tested in narrow contexts, mainly in locomotion or manipulation tasks, often restricted to a single simulation benchmark \citep{fujimoto2018addressing, haarnoja2018soft, chen2020randomized, moskovitz2021tactical, d2022sample}, leading to questions about their broad applicability and robustness. In this study, our goal is to consolidate these lessons and address the following research questions: \textit{Which regularization techniques lead to robust performance improvements across diverse tasks and agent designs? Can generic regularization techniques outperform domain-specific RL techniques that directly use the MDP structure?}. We extend the scope of prior research by examining over 60 design choices implemented within the Soft Actor-Critic framework. We test a diverse array of tasks, including both locomotion and manipulation, within two simulation benchmarks and two replay ratio regimes.
This comprehensive approach offers a deeper understanding of the effectiveness of these regularization techniques in various settings.

Our main result is a bitter lesson: across varied tasks, general neural network regularizers significantly outperform most RL-specific algorithmic improvements in terms of agent performance. Specifically, we find general methods that are motivated by stabilization of gradient-based learning significantly outperform RL-specific algorithmic improvements across a variety of environments. Such emphasis on generality is in line with the celebrated “Bitter Lesson” essay~\citep{sutton2019bitter}.
Notably, network regularization enables agents to find effective policies on tasks previously impossible for model-free agents, such as those in the dog domain. Our findings also show that layer normalisation is more effective in reducing overestimation than techniques specifically designed for mitigating Q-value overestimation in critic networks. Consequently, we show that replacing the ubiquitous Clipped Double Q-learning with network regularization techniques leads to significant performance gains. Our research further explores the impact of overestimation, overfitting, and plasticity loss on agent performance in a unified experimental setup. We examine the correlation between these factors and agent performance, showing a strong negative correlation for value overestimation and agent plasticity metrics. These influences vary significantly across environments, underscoring the complex nature of their effects on learning. A key observation is the environment-dependent performance of various methods. Strategies excelling in locomotion tasks may falter in manipulation scenarios and vice versa. Comparisons between experiments on the DeepMind Control Suite \citep{tassa2018deepmind} and MetaWorld \citep{Meta2019World} demonstrate the necessity for diverse benchmarking in research, highlighting the value of expansive experimental setups.


\begin{enumerate}
    \item Our study presents an extensive empirical analysis of various regularization techniques in off-policy RL. We evaluate the effectiveness, robustness, and generality of 12 SAC design choices derived from recent literature, examining their diverse interactions. This encompasses testing 64 model designs across 14 tasks from two benchmarks under two replay ratio regimes.
    \item Our findings show that combining well-established network regularization techniques with methods that prevent plasticity loss effectively addresses the value estimation problem, eliminating the need for critic regularization. Specifically, we observe that in network/plasticity regularized agents using critic regularization often leads to significant performance degradation. Leveraging these insights, we demonstrate that integrating specific regularization methods into the basic Soft Actor-Critic framework leads to state-of-the-art performance in dog domain tasks for model-free approaches.
    \item Our study investigates the correlation between overestimation, overfitting, and plasticity proxies, and their impact on agent performance. We discover that interventions aimed at one type of issue, such as full-parameter resets, significantly affect proxies for issues other than plasticity such as overestimation and overfitting, often more than interventions specifically designed for those other issues. This suggests that RL agents encounter a range of complex problems that collectively affect the learning process.
    
\end{enumerate}

\section{Background}

We consider a Markov Decision Process (MDP) \citep{puterman2014markov,sutton2018reinforcement} which is described via a tuple $(S, A, r, p, \gamma)$, where states $S$ and actions $A$ are continuous, $r(s, a)$ is the transition reward, $p(s'|s, a)$ is a transition mapping, $p_0$ is the starting state distribution and $\gamma \in (0,1]$ is the discount factor. Policy, denoted as $\pi(a|s)$ is a state-conditioned action distribution. Maximum Entropy Reinforcement Learning (MaxEnt RL) objective \citep{ziebart2008maximum, haarnoja2017reinforcement} is to find a policy that maximizes the expected sum of discounted returns and policy entropies, or equivalently expected initial state values according to $\pi^{*} = \arg \max \mathbb{E}_{p_0} V^{\pi}(s_0)$. The Q-value is defined as $Q^{\pi}(s,a) = r(s, a) + \gamma V^{\pi}(s')$. State value is defined by $V^{\pi}(s) = \mathbb{E}_{\pi} (Q^{\pi}(s,a) - \alpha \log \pi(a|s))$, where $\alpha \log \pi(a|s)$ is the maximum entropy term. In actor-critic, policy and Q-value functions are represented by parameterized function approximators \citep{silver2014deterministic}. Policy parameters $\theta$ are updated to maximize the value approximation at sampled states $s$ from an off-policy replay buffer $\mathcal{D}$ \citep{fujimoto2018addressing, haarnoja2018soft}:
\begin{equation}
\label{eq:actor_objective}
\begin{split}
& \theta^{*} = \arg \max_{\theta} \underset{\mathcal{D}}{\mathbb{E}}~ Q_{\phi}(s,a) - \alpha \log \pi_{\theta}(a|s), ~ a \sim \pi_{\theta}.
\end{split}
\end{equation}
The critic parameters $\phi$ are updated by minimizing the temporal-difference \citep{silver2014deterministic}:
\begin{equation}
\label{eq:critic_objective}
\begin{split}
& \phi^{*} = \arg \min_{\phi} \underset{\mathcal{D}}{\mathbb{E}} \bigl(Q_{\phi}(s, a) - r_(s,a) - \gamma \Bar{V}_{\phi}(s') \bigr)^2,
\end{split}
\end{equation}
where $\Bar{V}_{\phi}(s')$ is the target network \citep{mnih2015human}.

\subsection{Overestimation}

Q-learning methods employing function approximation have been observed to exhibit a bias toward overestimation, a phenomenon critical to the training process~\citep{thrun2014issues, fujimoto2018addressing}. Positive bias stems from the policy being trained to locally maximize action-value estimates, leading its actions to exploit potential model errors for higher scores. Modern actor-critic algorithms leverage a variety of countermeasures to overestimation of Q-value targets, with Clipped Double Q-learning (\textbf{CDQ}) \citep{fujimoto2018addressing} being most used by many other algorithms \citep{haarnoja2018soft, chen2020randomized, hiraoka2021dropout}. In CDQ, the algorithm maintains two critics and uses their minimum as an approximate Q-value lower bound. The CDQ was generalized to the following pessimistic objective \citep{ciosek2019better, moskovitz2021tactical, cetin2023learning}: 
\begin{equation}
\label{eq:cdql_general}
\begin{split}
    & Q_{\phi}^{\beta}(s,a) = Q^{\mu}_{\phi}(s,a) - \beta Q^{\sigma}_{\phi}(s,a).
\end{split}
\end{equation}
We denote the level of pessimism as $\beta$, and the critic ensemble mean and standard deviation as $Q^{\mu}_{\phi}$ and $Q^{\sigma}_{\phi}$ respectively. In particular, for $\beta = 1$, the above rule is exactly equal to the CDQ minimum \citep{ciosek2019better, cetin2023learning}. The success of pessimistic updates led to various methods for adjusting $\beta$ online. A recent approach, Generalized Pessimism Learning (\textbf{GPL})~\cite{cetin2023learning}, estimates the critic approximation error and modifies $\beta$ accordingly. A different strategy, Tactical Optimism and Pessimism (\textbf{TOP}) \cite{moskovitz2021tactical}, adjusts pessimism independent of the estimated approximation error. Specifically, TOP uses an external bandit controller to maximize online episodic rewards. Whereas this controller is aligned with the RL objective, it only allows for discrete values of pessimism.

\subsection{Overfitting}

Overfitting, while not commonly scrutinized in reinforcement learning, has gained attention in recent discussions~\citep{li2022efficient} as a phenomenon correlated with performance decline in models characterized by a high ratio of updates to data. To evaluate overfitting in agents, \citet{li2022efficient} utilizes a validation dataset that consists of samples gathered using the same policy as the canonical replay buffer. The validation buffer is established to provide an unbiased assessment of the critic error in experiences that were not used in the learning. Although there are many strategies to deal with overfitting in supervised learning, only a few of them were applied in the context of RL. To this end, application of Weight Decay (\textbf{WD}) \citep{schwarzer2023bigger}, Layer Normalization (\textbf{LN}) \cite{ball2023efficient} or Spectral Normalization (\textbf{SN}) \cite{cetin2023learning} was shown to greatly effect the performance of the underlying agent.


\subsection{Plasticity}
Plasticity, in the context of models, refers to their ability to learn new information. The concept of plasticity loss has recently gained prominence in the deep learning community, particularly in supervised learning~\citep{achille2017critical, NEURIPS2020_288cd256, dohare2021continual} and RL~\citep{nikishin2022primacy, dohare2021continual, pmlrv202lyle23b, lee2023plastic, kumar2023maintaining, nikishin2023deep}. Numerous hypotheses have been proposed regarding the sources of plasticity loss, including dead or dormant units, rank collapse, and divergence due to large weight magnitudes~\citep{lyle2022understanding, sokar2023dormant, kumar2020implicit, dohare2021continual}. However, none of these mechanisms alone is sufficient to explain the phenomenon of plasticity loss. Whereas the cause of plasticity loss remains to be discovered, various approaches for regularizing the model plasticity have been proposed. For example, full-parameter \textbf{resets} of actor-critic modules were shown to greatly improve the agent's ability to learn \citep{nikishin2022primacy, d2022sample}. The problem of plasticity was also tackled at the level of the activation function with Concatenated ReLU (\textbf{CRLU}) \citep{abbas2023loss} or the optimizer with the Sharpness-Aware Minimization (\textbf{SAM}) \citep{foret2020sharpnessaware}.

\section{Study Design}

In this paper, we analyze the impact of various interventions on SAC performance across seven DeepMind Control Suite~\citep{tassa2018deepmind} (DMC) tasks: acrobot-swingup, hopper-hop, humanoid-walk, humanoid-run, dog-trot, dog-run, quadruped-run and seven MetaWorld~\citep{Meta2019World} (MW) tasks: Hammer, Push, Sweep, Coffee-Push, Stick-Pull, Reach, Hand-Insert. We chose a wide spectrum of tasks, ranging from easy (acrobot-swingup, Reach) to barely solvable (dog-run) for generic insights that are not overfitted to only a specific group. We choose tasks that are not easily solved by the baseline high replay SAC, as presented in \citet{d2022sample} and \citet{hansen2022temporal}, with added dog tasks (which are generally unsolved in state-based representation). Finally, following \citep{li2022efficient}, we conduct experiments in low 2 and high replay regimes. Such experimental design allows us to pinpoint if specific regularization targets issues associated with high replay, or if it is universally applicable across varying replay regimes. Categorizing them based on current state-of-the-art methods, we identify three intervention groups:
\begin{itemize}
\item Critic Regularizations (CR),
\begin{itemize}
\item Clipped Double Q-learning (CDQ)~\citep{fujimoto2018addressing},
\item Tactical Optimism Pessimism (TOP)~\citep{moskovitz2021tactical},
\item Generalized Pessimism Learning (GPL)~\citep{cetin2023learning},
\end{itemize}
\item Network Regularizations (NR),
\begin{itemize}
\item Layer Norm {(LN)}~\citep{ba2016layer},
\item Spectral Norm {(SN)}~\citep{miyato2018spectral,zhang2018selfattention,brock2018large},
\item Weight Decay {(WD)}~\citep{loshchilov2017decoupled},
\end{itemize}
\item Plasticity Regularizations (PR),
\begin{itemize}
\item Resets {(Res)}~\citep{nikishin2022primacy},
\item Concatenated ReLU activations (CRLU)~\citep{abbas2023loss},
\item Sharpness-Aware Minimization Optimizer (SAM)~\citep{foret2020sharpnessaware}.
\end{itemize}
\end{itemize}
To explore the interactions between interventions, we systematically run all possible combinations of methods across groups, ensuring that methods from the same group are not combined. Each configuration is evaluated on 10 seeds. In the results analysis, we categorize marginalization into three levels:

\textit{First-order marginalization} combines all results for a specific intervention. For instance, the marginalized performance of layer norm will be computed as the average performance across all combinations with interventions from other groups (in this example, Critic Regularizations and Plasticity Regularizations). 

\textit{Second-order marginalization} involves evaluating the performance of fixed pairs of methods from two groups and marginalizing results from the third group.

\textit{Third-order} results involve no intervention marginalization and represent the performance of a specific combination, including one method from each group. The only marginalization is over all tested environments (we present these results in Appendix \ref{appendix:third_order} due to space constraints). This cumulative result provides insight into the overall impact of a given intervention.  

Furthermore, we conduct an analysis of various proxy metrics associated with the problems of overestimation, overfitting, and plasticity loss. For overestimation, we evaluate the state-action critic approximation error, denoted as $b_{\phi}(s,a)$, is quantified as the disparity between the critic output and the true on-policy Q-value according to $b_{\phi}(s,a) = Q_{\phi}(s, a) - Q^\pi(s, a)$, where $Q_{\phi}$ denotes the critic Q-value approximation and $Q^{\pi}$ represents the on-policy Q-value which we estimate via a Monte-Carlo rollout with 5 samples. To calculate overfitting, we compare average TD errors on evaluation trajectories (which are not used for learning) to average TD errors observed in training according to $o_{\phi} = \frac{\mathbb{E}_{\mathcal{D}_v} TD_{\phi}}{\mathbb{E}_{\mathcal{D}} TD_{\phi}}$, where $o_{\phi}$ denotes critic overfitting, $\mathcal{D}_v$ denotes validation data, and $TD_{\phi}$ denotes the temporal difference loss. As such, the extent of overestimation is then quantified by the ratio of validation TD error to training TD error. We monitor plasticity loss by the rank of penultimate layer representations~\citep{kumar2020implicit}, dormant neurons or dead units~\citep{sokar2023dormant}, {the L2 norm
of weights~\citep{nikishin2022primacy,pmlrv202lyle23b}, and gradient norm~\citep{nikishin2022primacy,pmlrv202lyle23b} as a proxy for plasticity loss.}
\section{Experiments}

\subsection{Combination of interventions -- First-order marginalization}
\label{sec:first_margin}
\paragraph{Study description:}

First-order marginalization provides insights into the robust impact of a given intervention on model performance, irrespective of what other type of regularization it is paired with. To measure such robustness, we compare the performance of the baseline SAC model augmented with one specific regularization (e.g., SAC + WD) to the performance of SAC augmented with this regularization paired with some other technique (e.g. SAC + WD + Resets).
\begin{figure}[ht!]
\begin{center}
\begin{minipage}[h]{1.0\linewidth}
			\centering
			\begin{subfigure}{0.88\linewidth}
				\includegraphics[width=\textwidth]{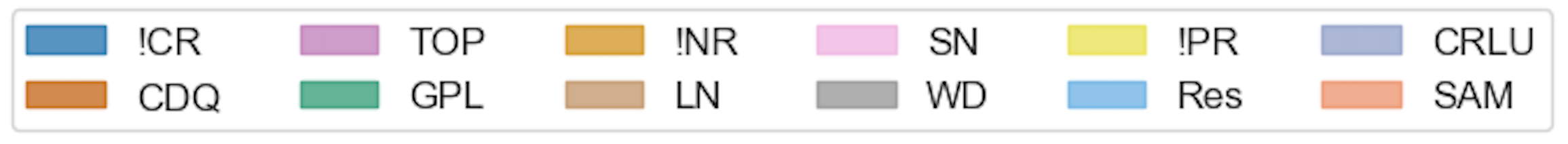}
			\end{subfigure}
		\end{minipage}
\begin{minipage}[h]{1.0\linewidth}
\begin{subfigure}{0.99\linewidth}
             \includegraphics[width=0.49\linewidth]{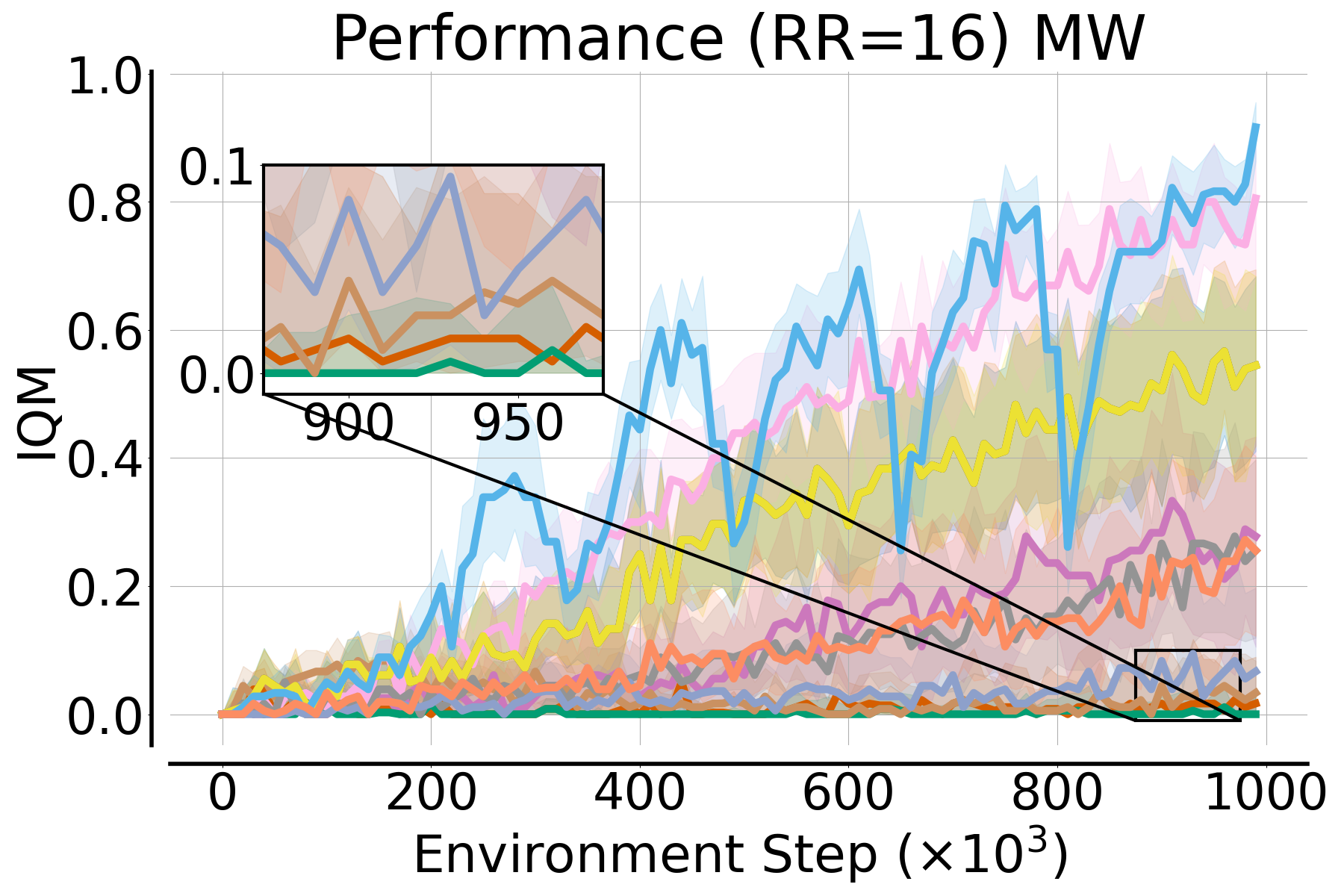}
         \includegraphics[width=0.49\linewidth]{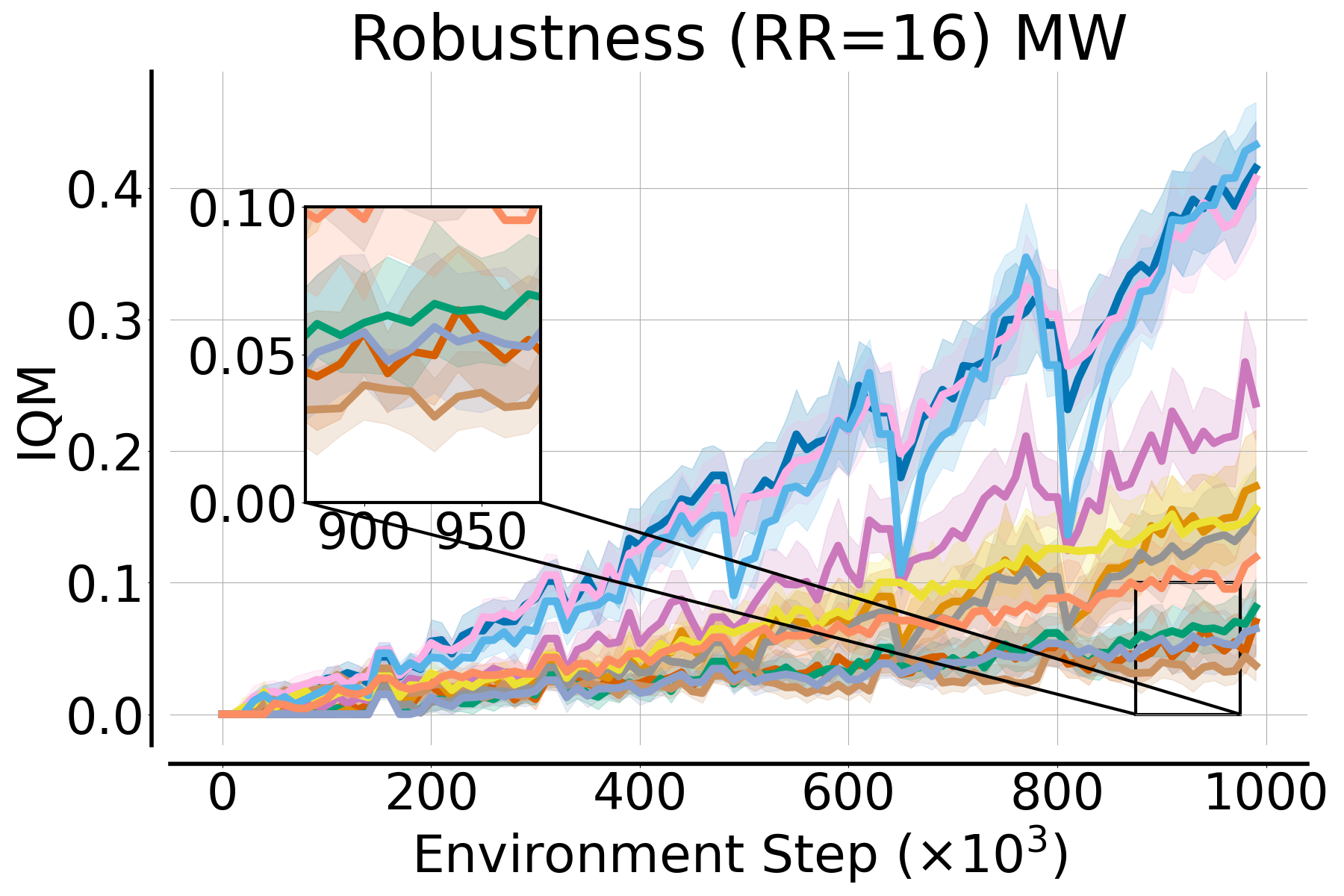}
\end{subfigure}
  \begin{subfigure}{0.99\linewidth}
        \includegraphics[width=0.49\linewidth]{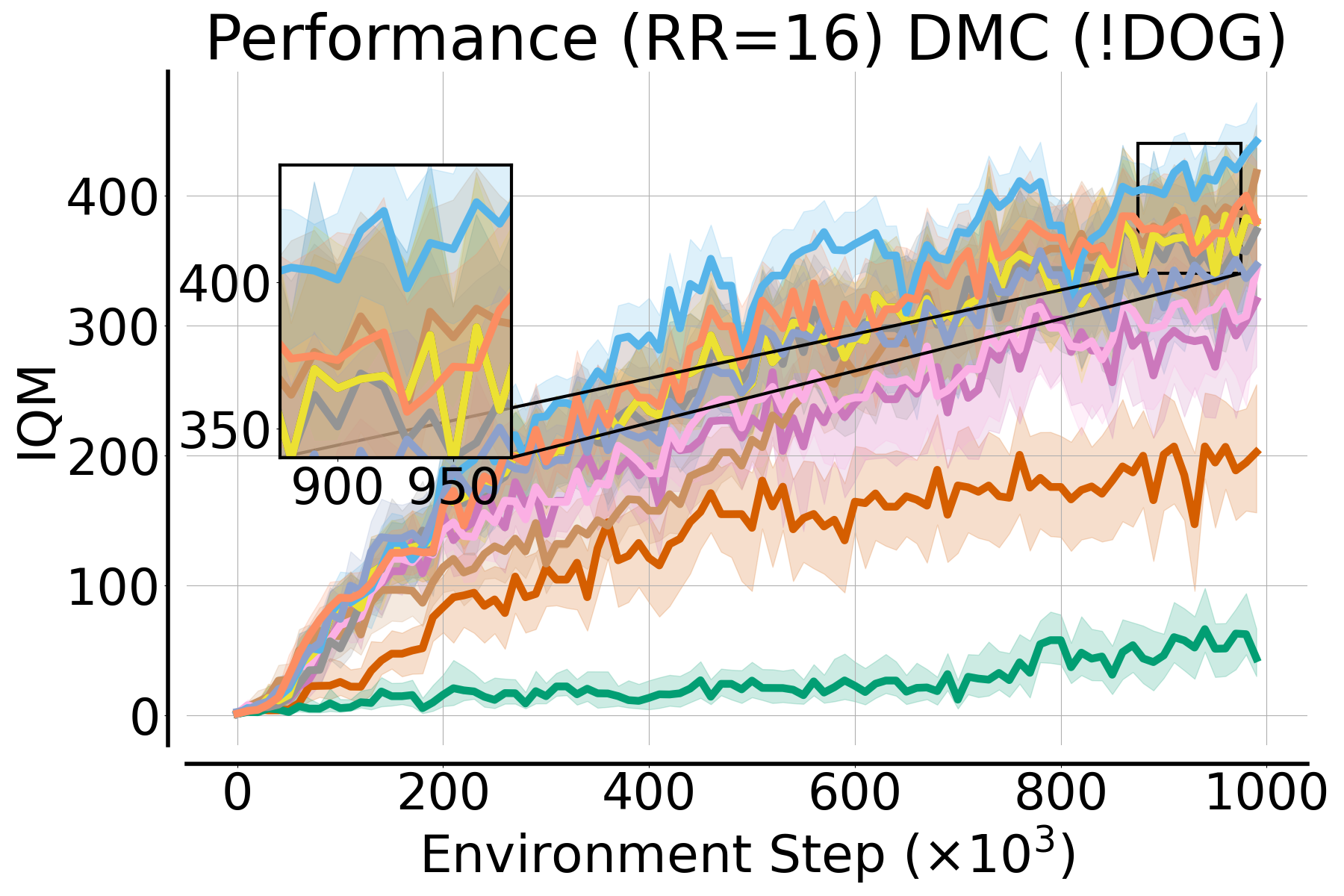}
         \includegraphics[width=0.49\linewidth]{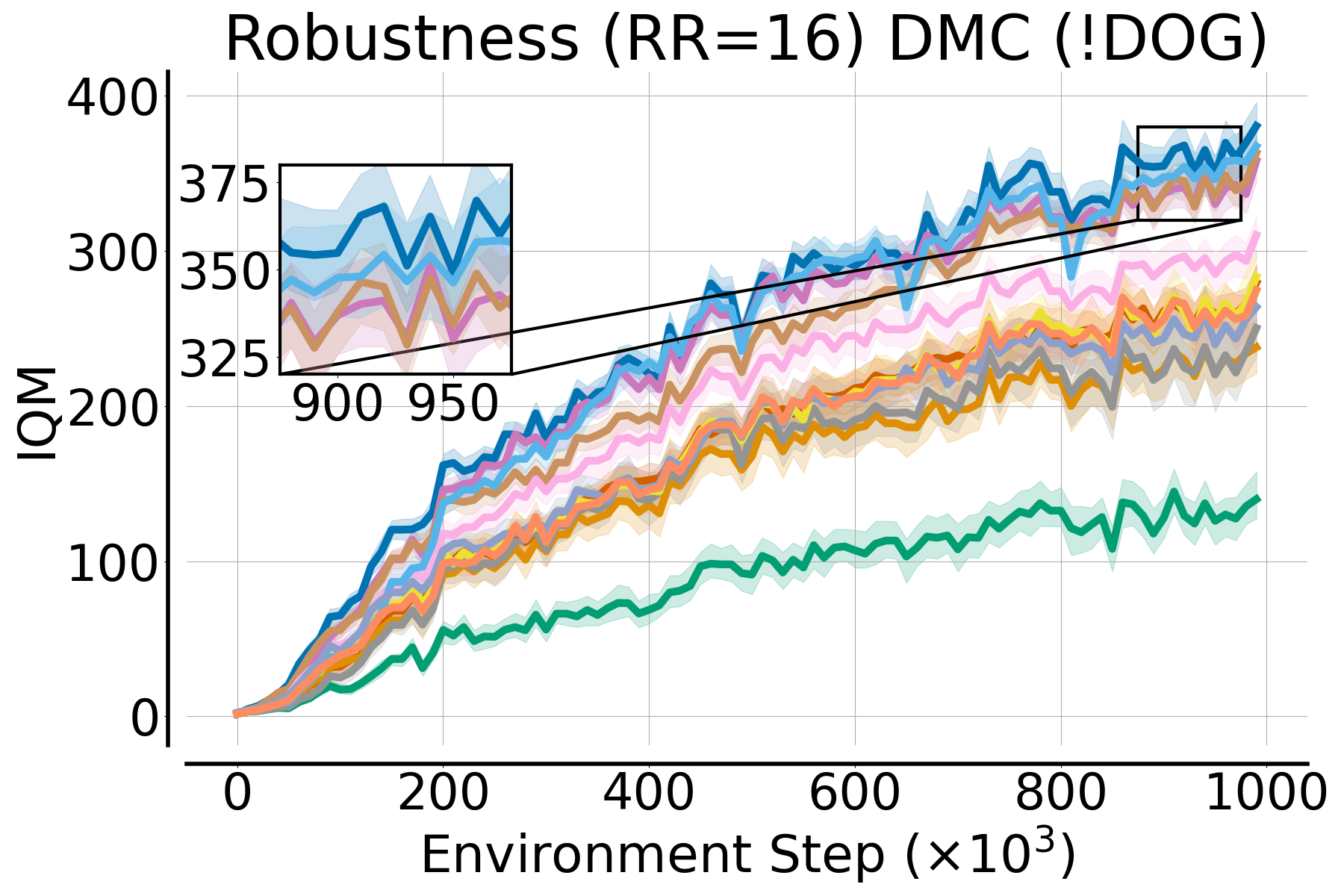}
 \end{subfigure}
 \begin{subfigure}{0.99\linewidth}
        \includegraphics[width=0.49\linewidth]{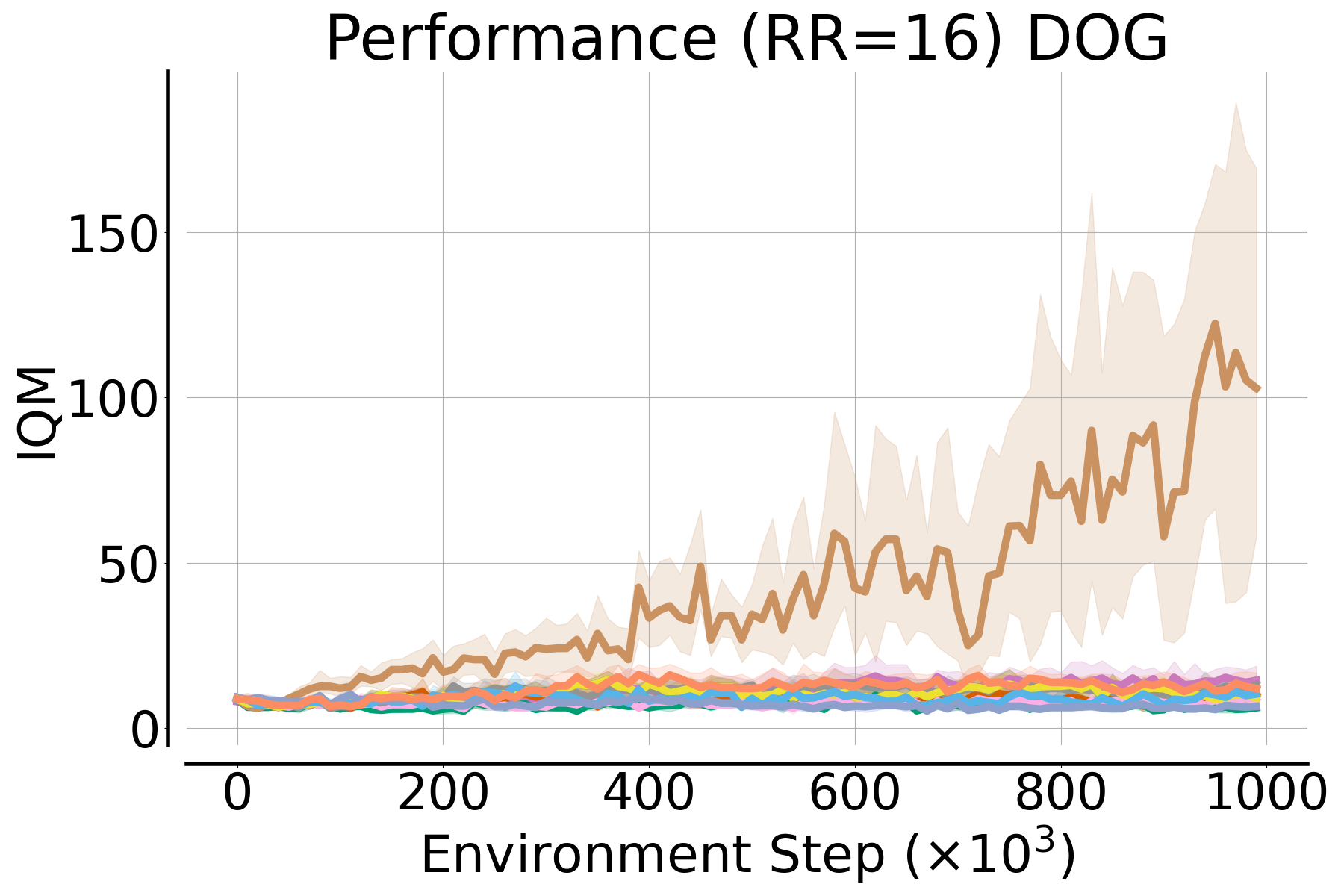}
         \includegraphics[width=0.49\linewidth]{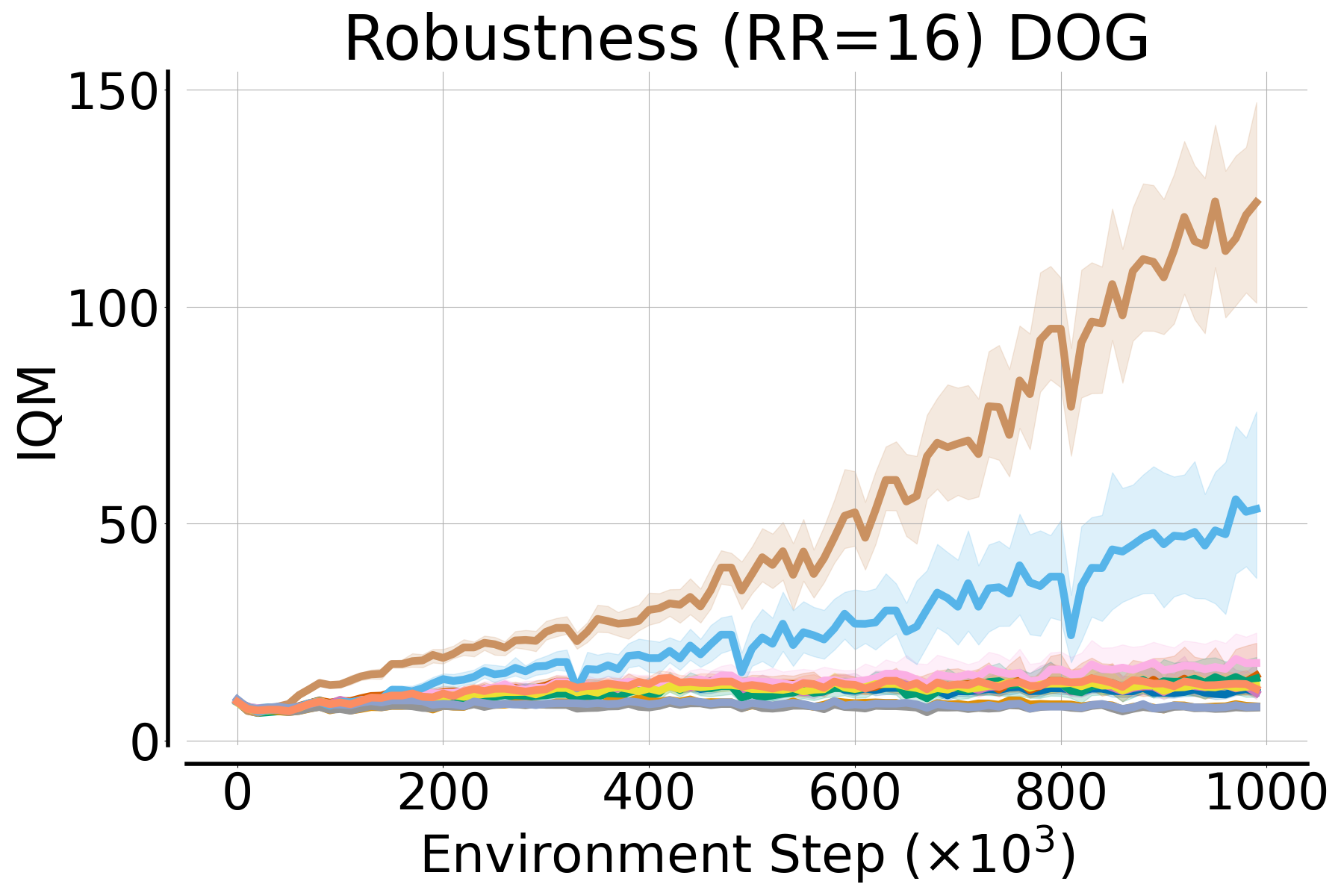}
 \end{subfigure}

\end{minipage}
			\caption{
   IQM performance of First-Order Marginalization. The left column presents results for baseline SAC augmented with a single regularization technique (and thus uses 10 seeds per task), and the right column presents the aggregate performance of a specific regularization technique when paired with other regularizations (and thus uses 640 seeds per task). {Results are presented for MW (top row), DMC without Dog environments (middle row) and only Dog-run and Dog-trot (bottom row) benchmarks. 14 tasks.}
   }
			\label{fig:no_interaction}
   \vspace{-0.2in}
   \end{center}
   
\end{figure}
\paragraph{Results:}
Examining the plots in Figure~\ref{fig:no_interaction} shows that network and plasticity regularization techniques are generally more effective than critic regularization -- dark blue line (!CR) on Robustness plots on MW and DMC (!DOC). We observe these results for both simple models using one regularization technique and more complex agents leveraging many regularizations at once. Most notably, avoiding the use of critic regularization interventions (!CR) proves advantageous for both DMC and MW, especially if some other type of regularization is used (such as layer norm or full-parameter resets). This result is somewhat surprising, as critic regularization methods were designed specifically for off-policy actor-critic agents, whereas the network and plasticity regularization techniques are general. Notably, TOP \citep{moskovitz2021tactical} emerges as an exception, particularly showcasing its effectiveness on the DMC benchmark. Conversely, the GPL intervention exhibits the least robust performance in both DMC and MW. {Upon further analysis of the impact of CDQL presence (see Section~\ref{sec:CDQL_negative}), it becomes apparent that in certain environments, such as Hopper Hop, the adverse effects of this intervention cannot be mitigated even with additional regularizations.}
Full-parameter resets \citep{nikishin2022primacy}, tailored for high replay ratio regimes, prove to be one of the most robust approaches in this RR regime. Further analysis reveals discrepancies in conclusions between benchmarks. {Clearly, LN is the most effective approach in the DMC Dog environments (bottom row in Figure~\ref{fig:no_interaction}), but it also ranks among the top four interventions in the remaining DMC environments. However, it exhibits very poor performance on the MW benchmark. Therefore, we find SN to be more robust, significantly aiding the MW benchmark and providing moderate assistance in the DMC scenarios.}

\begin{tcolorbox}
\paragraph{Takeaways:}
\begin{itemize}
    \item Critic regularization methods exhibit limited effectiveness in enhancing performance. When using network or plasticity regularization, critic regularization leads to reduced performance.
    \item Periodical network resetting is the most robust intervention across two benchmarks in a high replay ratio regime, and highly surpasses other plasticity regularization techniques in both robustness and performance.
    \item {Layer norm is essential for Dog environments.}
    \item When considering network regularization approaches, layer norm is generally recommended for DMC, while {spectral} norm is more effective for MW benchmarks. When considering a {diverse range of tasks}, we find spectral norm to be more robust than layer norm. Weight decay has generally low performance when used alone with SAC.
\end{itemize}
\end{tcolorbox}

\subsection{Combination of interventions -- Second-order marginalization}
\label{sec:second_order}
\paragraph{Study description:}
This study delves into second-order marginalization to pinpoint the most effective combinations. Results are presented across various replay ratios (2 and 16) and benchmarks (DMC or MW). Given the limited impact of critic regularizations like CDQ or GPL in the first order experiments, our focus is on discerning the most advantageous combinations involving of regularization.

					

\begin{figure}[ht!]
\centering
    \begin{subfigure}{0.98\linewidth}
        \includegraphics[width=0.49\linewidth]{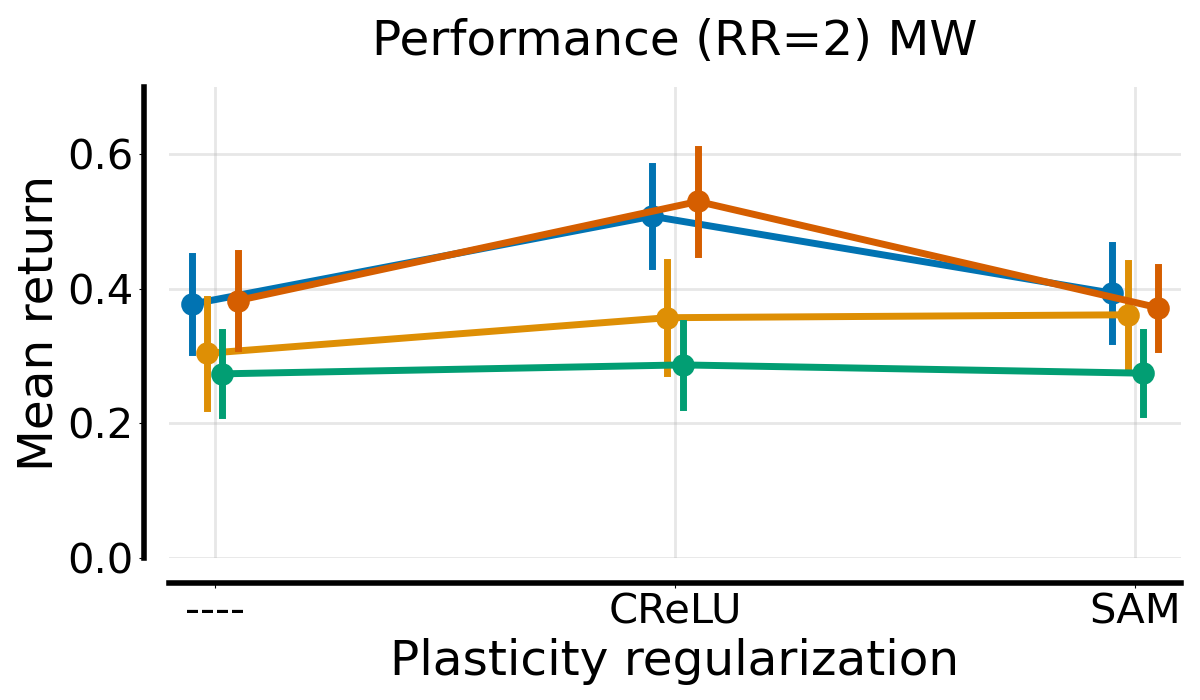}
        \hfill
        \includegraphics[width=0.49\linewidth]{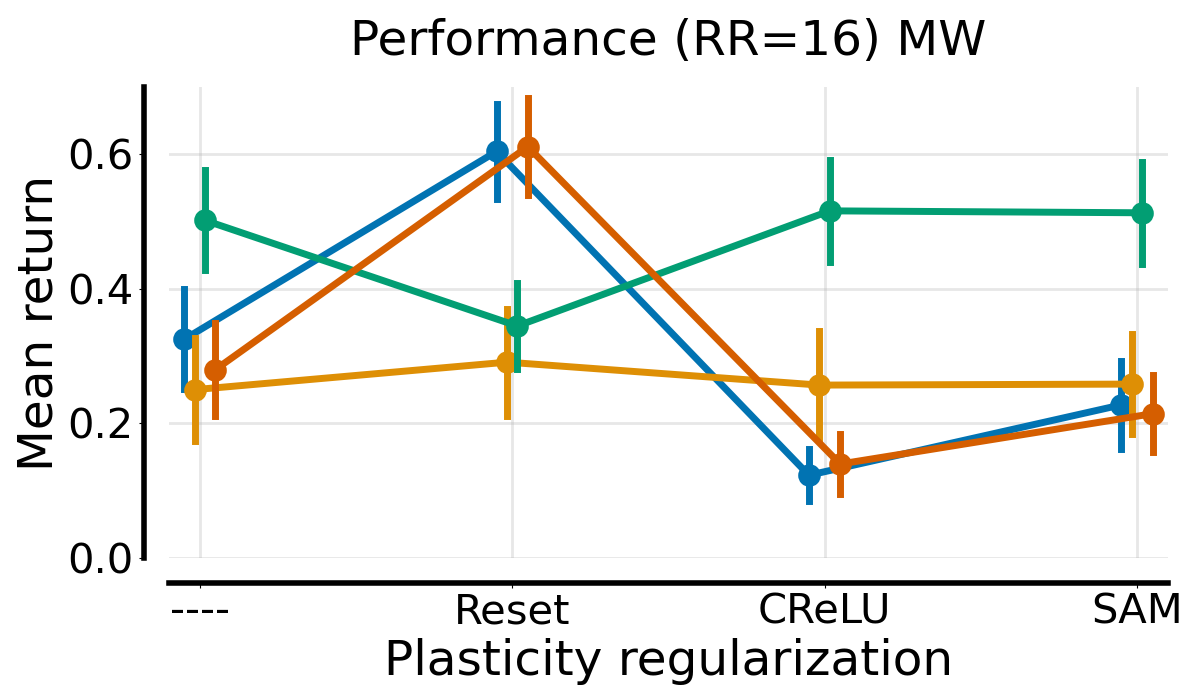}
    \end{subfigure}
    \hfill
    \begin{subfigure}{0.98\linewidth}
        \includegraphics[width=0.49\linewidth]{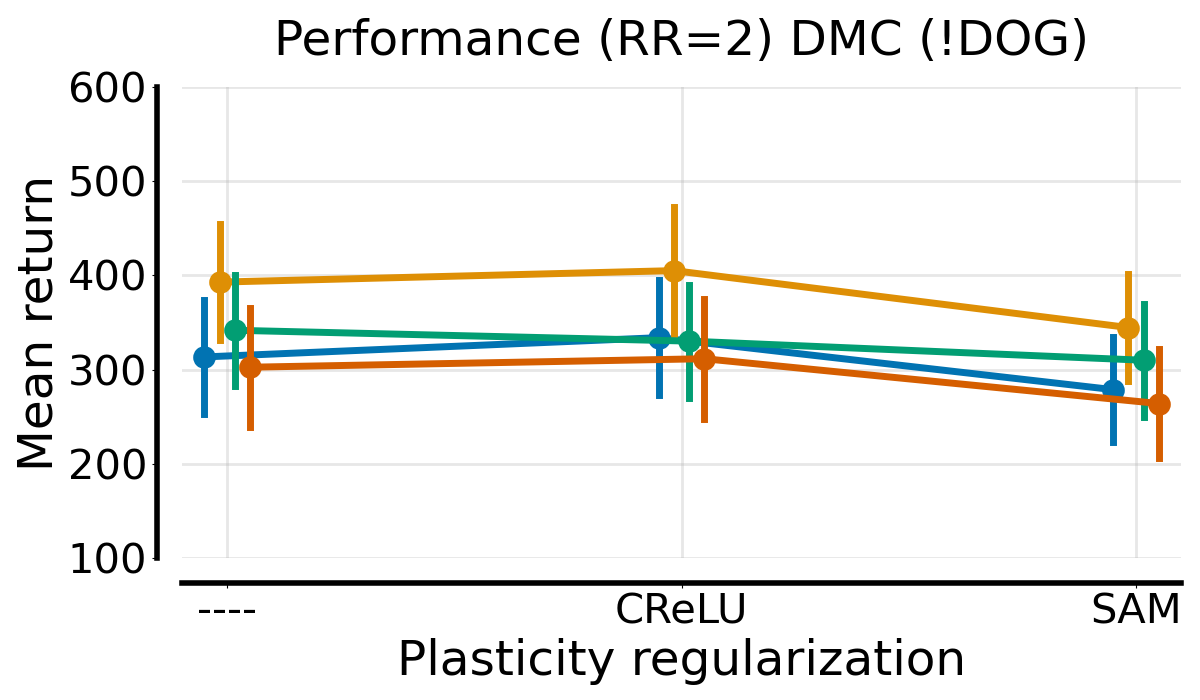}
        \hfill
        \includegraphics[width=0.49\linewidth]{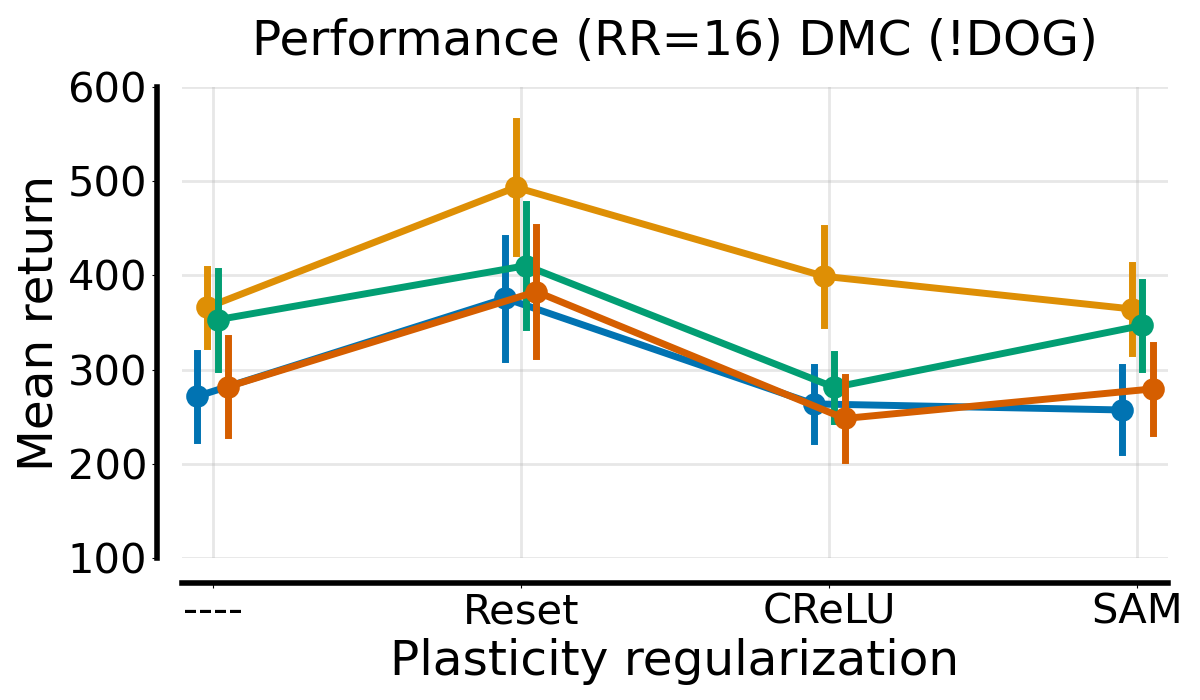}
    \end{subfigure}
    \hfill
    \begin{subfigure}{0.98\linewidth}
        \includegraphics[width=0.49\linewidth]{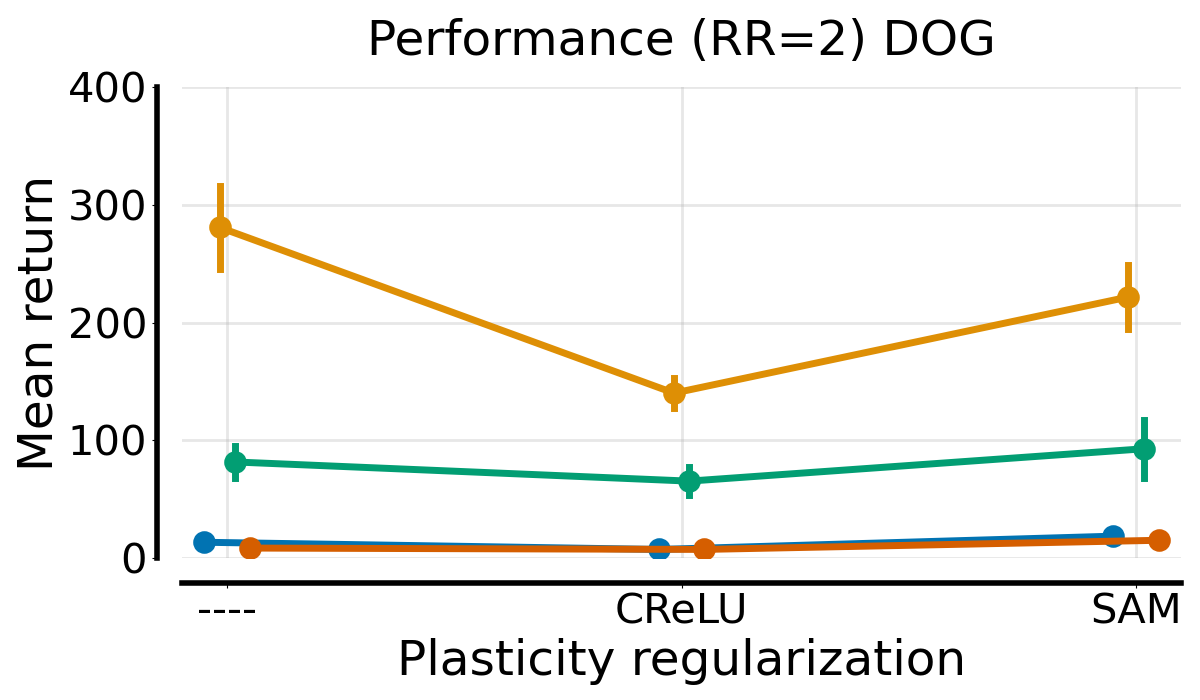}
        \hfill
        \includegraphics[width=0.49\linewidth]{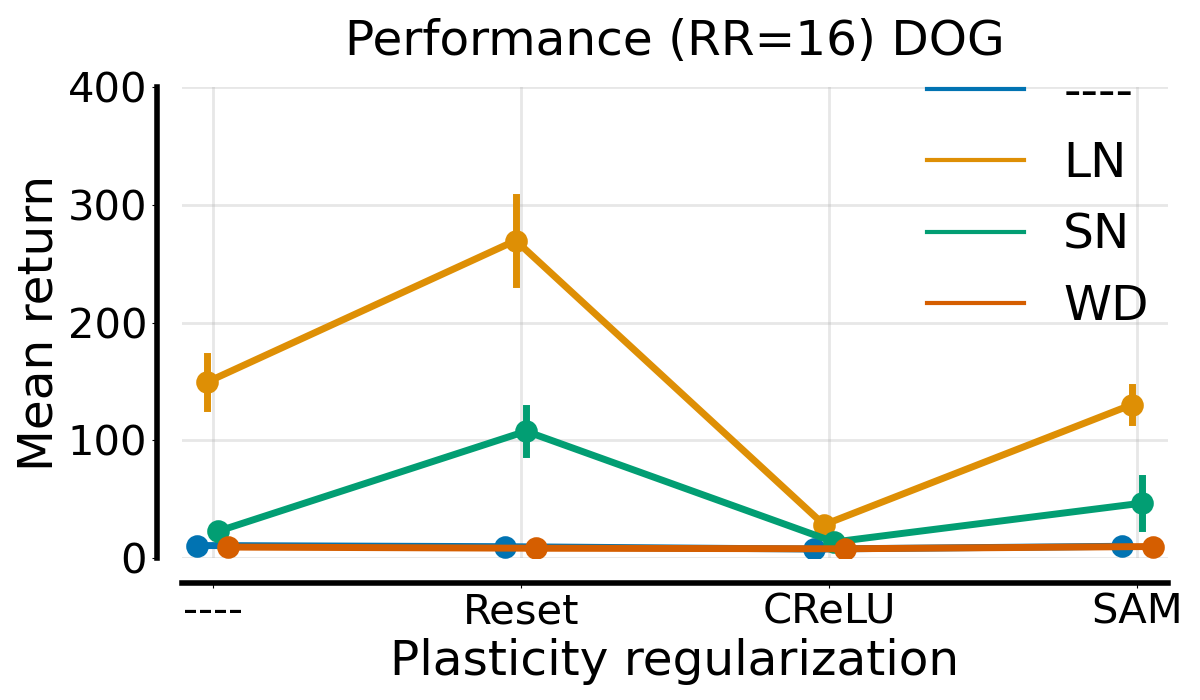}
    \end{subfigure}
\caption{
   Second-order results marginalizing critic regularization methods. On the x-axis, we have different types of plasticity regularization, and each colour denotes network regularization. For better readability, points within one plasticity regularization are spaced slightly horizontally. Vertical lines indicate standard error.
   }
\label{fig:marginalized}
\end{figure}

\paragraph{Results:}
On the DMC with RR=2 and RR=16 (top row in Figure~\ref{fig:marginalized}) a clear hierarchy of interventions is observed:  layer norm and right below it, spectral norm consistently outperforms others in mean return, irrespective of the plasticity regularization (x-axis). Notably, the combination of layer norm and resets in RR=16 ({middle and bottom-right plot} in Figure~\ref{fig:marginalized}) demonstrates exceptional performance across all critic regularization variations{ on DMC Dog and without Dog environments.} In contrast, the hierarchy of interventions on the MW benchmark ({top} row in Figure~\ref{fig:marginalized}) diverges significantly from the DMC setup. Furthermore, a higher replay ratio introduces shifts in training dynamics, as evidenced by SN transitioning from a lower position on RR=2 (bottom-left plot in Figure~\ref{fig:marginalized}) to nearly the most versatile intervention on RR=16 (bottom-right plot in Figure~\ref{fig:marginalized}). The results generally align with first-order marginalization findings, emphasizing the positive impact of SN, as well as using many different types of regularization at once in general. Deeper analysis (see Appendix~\ref{app:Grad_analysis}) reveals that on MW, indeed, the gradient norm in a higher RR regime is orders of magnitude bigger. The finding that most contrasts with the first-order experiments, is that we observe that weight decay can actually yield significant performance benefits, under the condition that it is paired with other specific methods, namely full-parameter resets. In particular, we observe that this combination yields synergies surpassing using any of these methods alone.

\begin{tcolorbox}
\paragraph{Takeaways:}
\begin{itemize}
    \item The DMC benchmark can be largely trivialized by using high RR agents combined with layer norm and full-parameter resets.
    \item Spectral normalization intervention ranks best for the Meta World benchmark, but it's not universally applicable. Whereas weight decay does not perform when used alone, it seems to have high synergy with full-parameter resets. 
    \item Resetting the network significantly outperforms other plasticity-inducing interventions such as CRLU and SAM. 
\end{itemize}
\end{tcolorbox}

\subsection{A closer look on Dog environment {performance}}
\label{sec:dog}
\paragraph{Study description:}

In this study, we delve into the intricacies of two challenging Dog tasks, Dog-Trot and Dog-Run, included in our DMC setup. These environments present considerable difficulties for model-free approaches relying on proprioceptive states, making them of particular interest within the research community. {Due to the inherent difficulty of these tasks, we conducted additional experiments using the top three methods identified in Figure~\ref{fig:task_specific_rr16} (Appendix) for 4 million steps, akin to approaches used in model-based~\citep{hansen2022temporal} or pixel-based studies~\citep{ji2024seizing}. For the rest of the detailed experimental information, please refer to Appendix~\ref{app:experim_details}.}

\paragraph{Results:}


\begin{figure}[ht!]
\centering
\begin{subfigure}{0.49\linewidth}
        \includegraphics[width=0.95\linewidth]{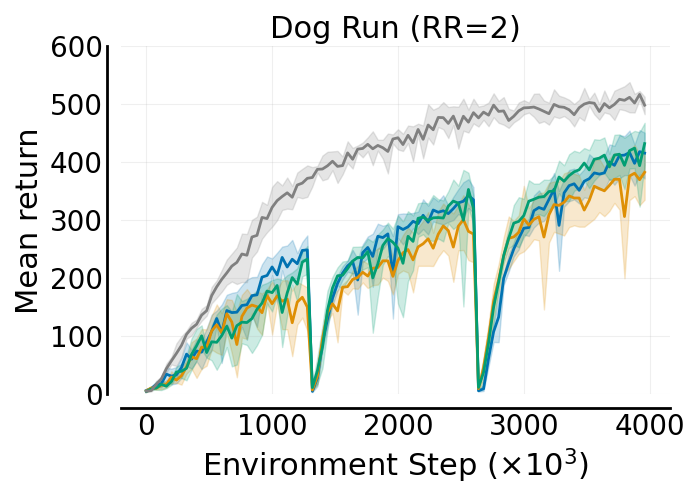}
\end{subfigure}
\begin{subfigure}{0.49\linewidth}
        \includegraphics[width=0.95\linewidth]{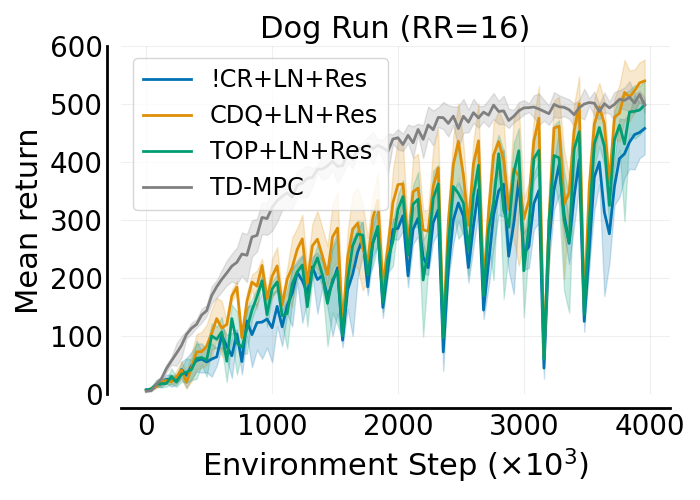}
\end{subfigure}
\begin{subfigure}{0.49\linewidth}
        \includegraphics[width=0.95\linewidth]{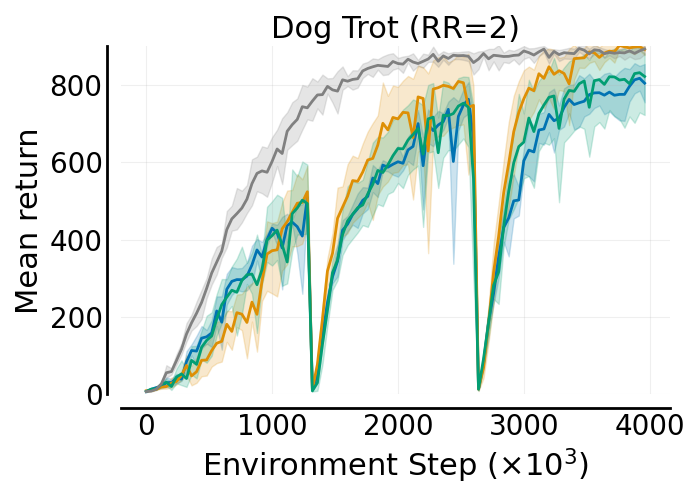}
\end{subfigure}
\begin{subfigure}{0.49\linewidth}
        \includegraphics[width=0.95\linewidth]{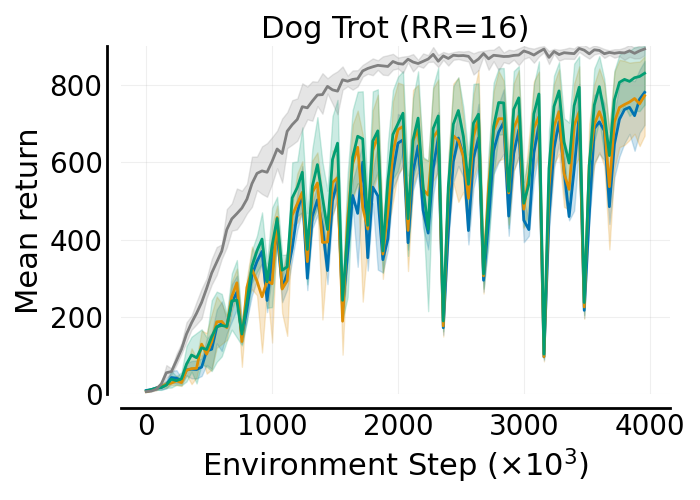}
\end{subfigure}
\caption{Mean return evolution across {4 million }timesteps for Dog-Run (top row) and Dog-Trot (bottom row) environments. Gray plot depicts model-based agent performance. Each plot showcases the top three combinations.}
\label{fig:dmc-dogs}
\end{figure}

\begin{figure}[ht!]
\centering
\begin{subfigure}{0.49\linewidth}
        \includegraphics[width=0.95\linewidth]{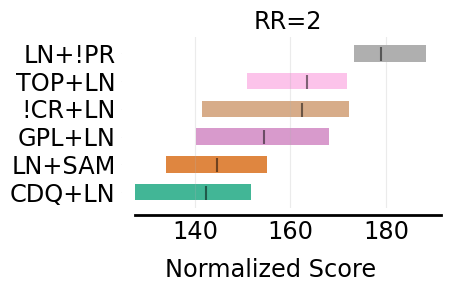}
\end{subfigure}
\begin{subfigure}{0.49\linewidth}
        \includegraphics[width=0.95\linewidth]{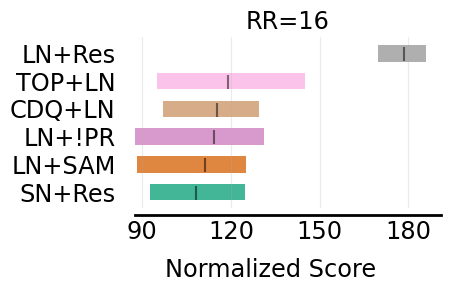}
\end{subfigure}
\begin{subfigure}{0.49\linewidth}
        \includegraphics[width=0.95\linewidth]{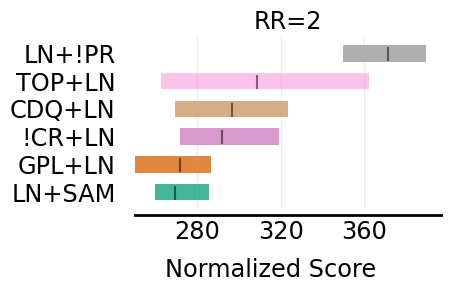}
\end{subfigure}
\begin{subfigure}{0.49\linewidth}
        \includegraphics[width=0.95\linewidth]{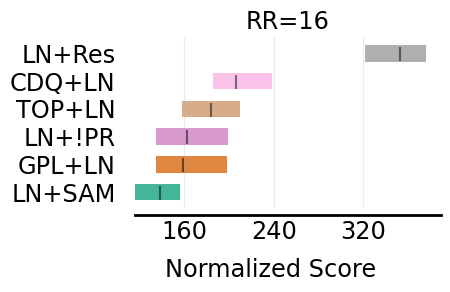}
\end{subfigure}
\caption{IQM performance of the top six intervention pair combinations based on 1 million steps experiments. The IQM is calculated based on the average of the last ten evaluation points in each run, not the last evaluation point. Results come from 1 million steps experiments. Top row: Dog-Run. Bottom row: Dog-Trot.}
\label{fig:iqm-dogs}
\end{figure}

Specific intervention combinations effectively tackle the challenges posed by the Dog environment, as depicted in Figure~\ref{fig:dmc-dogs}. Analyzing the top {three} approaches for Dog-Run and Dog-Trot tasks reveals the prevalence of layer norm in nearly all combinations. Additionally, each critic regularization approach contributes to the leading group. Notably, in scenarios with high replay ratios, resets emerge as a crucial intervention. 
These observations are further substantiated by the analysis of second-order marginalization IQM plots (see~\ref{fig:iqm-dogs}). Indeed, layer norm without critic regularization excels in RR=2, and layer norm with resets outperforms all others convincingly. This achievement is particularly notable as, to our best knowledge, no model-free agent has previously {find a better-performing policy within the training regime} the Dog environments using proprioceptive states. {Notably, there is a recent study\citep{ji2024seizing} where a model-free agent achieved comparable results on the Dog environments but using pixel-based inputs instead. Additionally, our results demonstrate that while a model-based approach on proprioceptive states~\citep{hansen2022temporal} outperforms slightly, the above model-free approach with simple regularization techniques achieves performance very close to that of the model-based approach. This suggests the efficacy and competitiveness of our approach in challenging environments.}

\begin{tcolorbox}
\paragraph{Takeaways:}
\begin{itemize}
    \item Well-established network regularization techniques such as layer norm and spectral norm enable {finding high-performing policies for} Dog-Trot and Dog-Run effectively.
    \item The choice of domain-specific RL critic regularization has little significance in dog environments when layer norm and resetting interventions are employed.
\end{itemize}
\end{tcolorbox}

\subsection{Correlation of Overestimation, Overfitting and Plasticity metrics with Performance}
\label{sec:correlations}
\paragraph{Study description:}
This study analyses the relationships between Overestimation, Overfitting, Plasticity, and model performance. Overestimation is quantified as an approximation error, overfitting as the ratio of TD error on the validation set to TD error on the training set. Expressing Plasticity loss is challenging, so we utilize proxy metrics, {including the percentage of Dormant neurons~\citep{sokar2023dormant}, representations rank~\citep{kumar2020implicit}, gradient norm~\citep{nikishin2022primacy,pmlrv202lyle23b}, and parameters norm~\citep{nikishin2022primacy,pmlrv202lyle23b}.}

We employ a Spearman correlation matrix to scrutinize these dependencies. This statistic is chosen because we observe non-linear yet monotonic dependencies between the mentioned metrics. We employ it on the data from all performed experiments, i.e., form runs with different combinations of interventions. Moreover, we do not have a division into RR=2 and RR=16, only the results from both setups are combined, and analyses are made on them. Overestimation, and gradient norm, and parameters norm are analyzed in a logarithmic scale for precision. We exclude metric pairs where the p-value of correlation is above $5\%$ by whitening tiles in the correlation matrix. 

\begin{figure}[ht!]
\centering
        \includegraphics[width=0.8\linewidth]{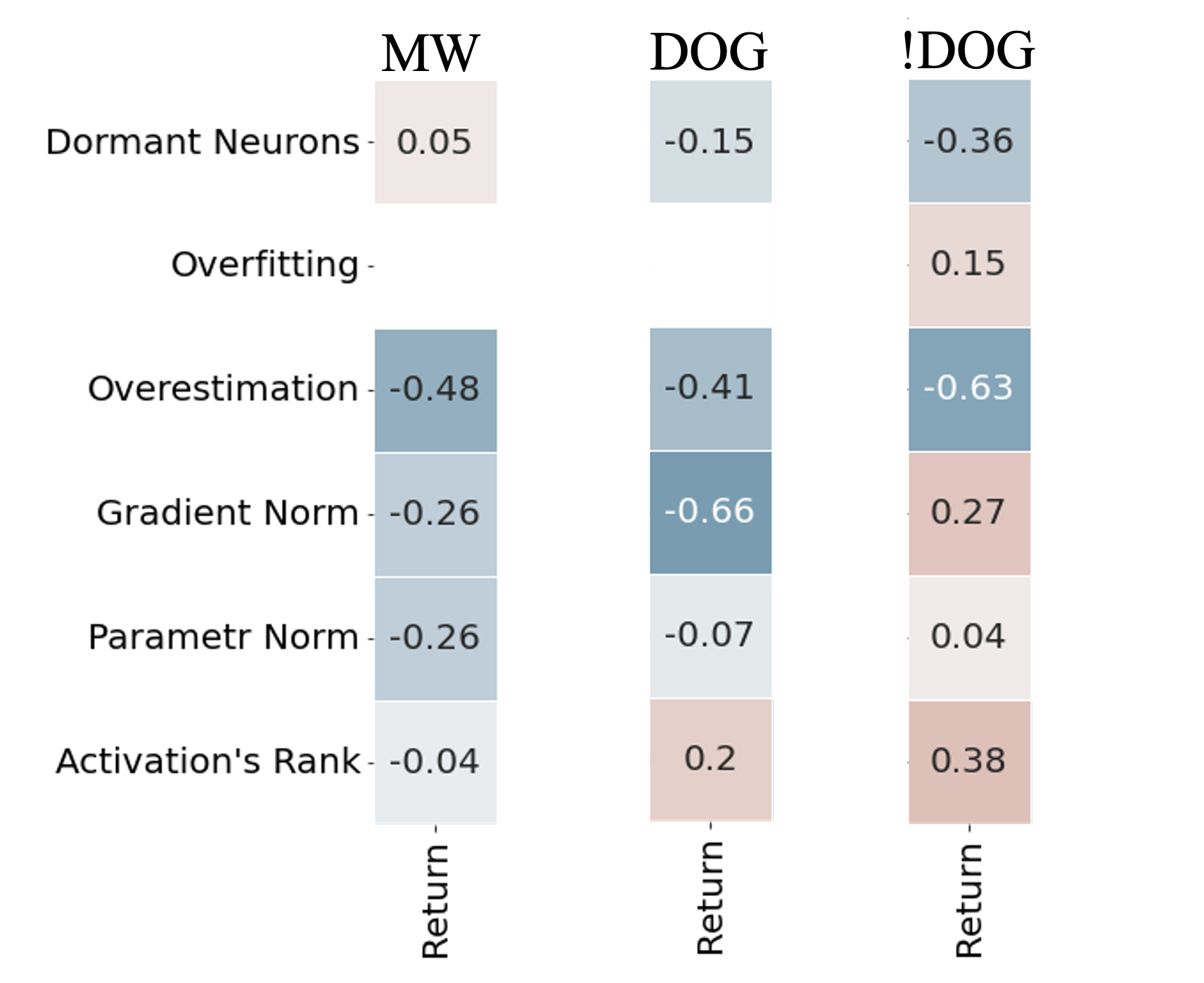}

\caption{Explanatory metrics correlations for three different groups of environment, namely: MetaWorld, DMC Dog environments, and DMC environments without Dog environments. It's important to observe that not only does the main explanatory metric, gradient norm, vary for dog environments, but the remaining DMC environments also exhibit a different correlation sign for this metric.}
\label{fig:dog_no_dog}
\end{figure}

\begin{figure}[ht!]
\centering
\begin{subfigure}{0.48\linewidth}
        \includegraphics[width=0.93\linewidth]{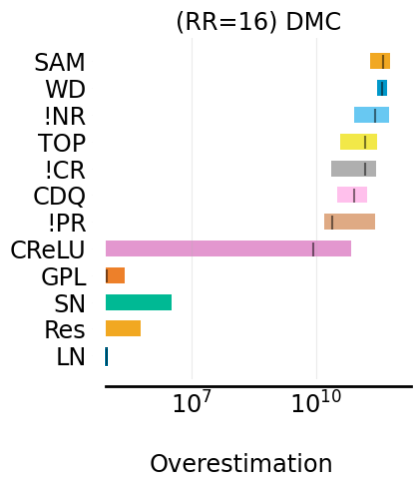}
\end{subfigure}
\begin{subfigure}{0.48\linewidth}
        \includegraphics[width=0.93\linewidth]{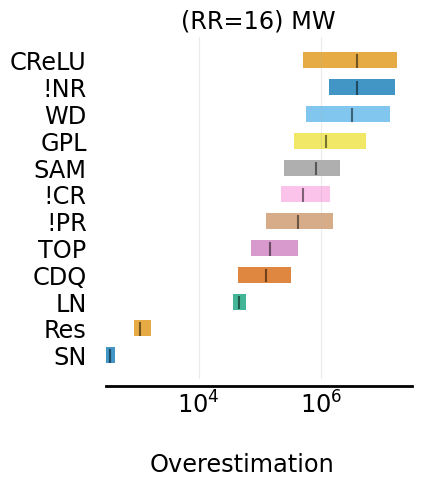}
\end{subfigure}
\caption{IQM overestimation in logarithmic scale. Each plot presents sorted results. The IQM is calculated based on the average of the last ten evaluation points in each run, not the last evaluation point. Left Figure: DMC benchmark. Right Figure: MW benchmark. Colours indicate hierarchy on the plot, not specific names.}
\label{fig:iqm-overestimation_1st_order}
\end{figure}

\paragraph{Spearman correlation:}
{In Figure~\ref{fig:dog_no_dog}, we investigate the relations between plasticity loss, overestimation and overfitting metrics and agents return, separately for every benchmark with special separation for Dog environments. Notably, overestimation exhibits the strongest correlation with agent returns on both benchmarks, offering insights into the findings of previous sections regarding the limited robustness of critic regularization methods designed to minimize overestimation. }
{Interestingly, as shown in Figure~\ref{fig:iqm-overestimation_1st_order}, layer norm and spectral norm effectively mitigate approximation errors, outperforming CR methods. 
However, further investigation on the DMC benchmark (Figure~\ref{fig:dog_no_dog}) uncovers a strong negative correlation between the percentage of dormant neurons and performance, closely tied to the rank of representations on the critic's penultimate layer. What is more, the overestimation is not the best predictor for Dog environments where we observe the highest values of overestimation (Figure~\ref{fig:dog_no_dog}). We hypothesize that overestimation becomes a good predictor of performance only when more fundamental issues, such as plasticity, are mitigated, indicating a multifaceted learning problem in harder environments. }

{
As an observation that supports this hypothesis, we refer to the critic gradient norm, which exhibits the most monotonic relation with the return in dog environments, as indicated by Spearman correlation (Figure \ref{fig:dog_no_dog}). 
Analysing Figure~\ref{fig:iqm-GradNorms3} one can see, that Dog environments especially with high replay ratio experience exploding gradients. 
The high gradient norm directly points to high curvature of the loss landscape, which, as indicated by \cite{lee2023plastic}, describes low input plasticity. For this reason, layer norm primarily smoothens the activation distribution and plays a critical role in making SAC work in dog environments. Environments from the MW benchmark also encounter challenges with high curvature loss landscapes (as indicated by the Spearman correlation between gradient norm and return). This suggests a resemblance between MW and DMC dog environments.}

Moreover, on the MW benchmark (Figure~\ref{fig:dog_no_dog}), the second-best correlated metrics with performance are the critic gradient norm and the critic parameters norm. This aligns with the results from section~\ref{sec:second_order}, highlighting the significant performance boost provided by spectral norm and weight decay, particularly in the RR=16 setup. An in-depth analysis of how RR increases the negative correlation of gradient norm and return can be found in sections \ref{app:Grad_analysis} and \ref{app:Scatter_regression} of Appendix.

\paragraph{Dog environments through gradients prism}

\begin{figure}[ht!]
\centering
\begin{subfigure}{0.49\linewidth}
        \includegraphics[width=0.95\linewidth]{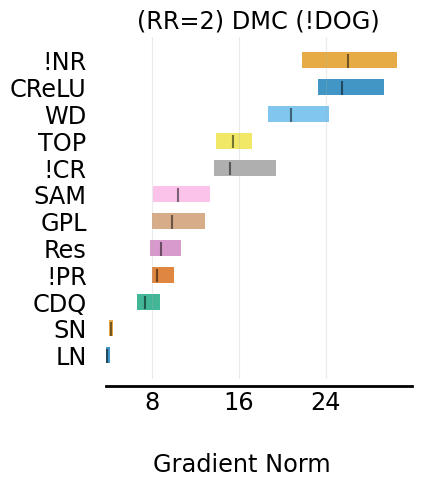}
\end{subfigure}
\begin{subfigure}{0.49\linewidth}
        \includegraphics[width=0.95\linewidth]{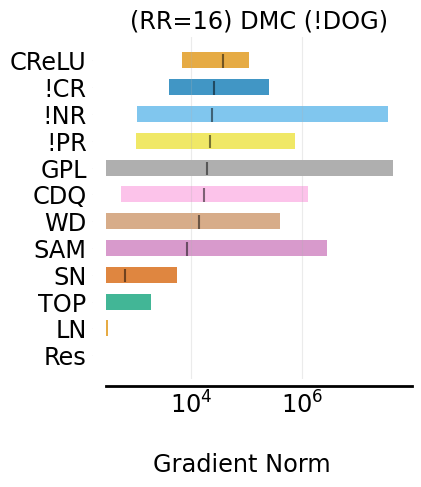}
\end{subfigure}
\begin{subfigure}{0.49\linewidth}
        \includegraphics[width=0.95\linewidth]{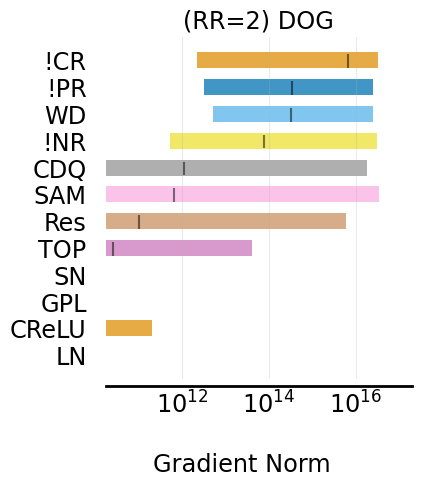}
\end{subfigure}
\begin{subfigure}{0.49\linewidth}
        \includegraphics[width=0.95\linewidth]{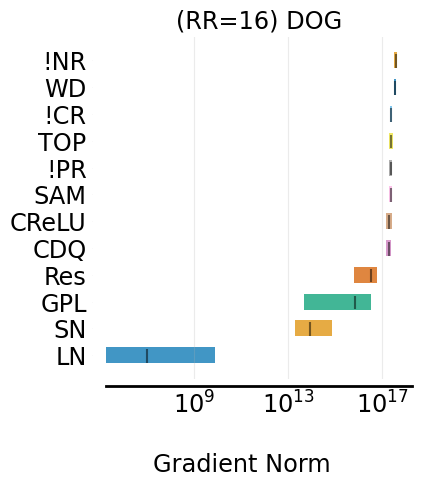}
\end{subfigure}
\caption{IQM gradient norm of first-order results. The IQM is calculated based on the average of the last ten evaluation points in each run, not the last evaluation point.}
\label{fig:iqm-GradNorms3}
\end{figure}



\begin{tcolorbox}

\paragraph{Takeaways:}
\begin{itemize}
    \item There are distinctive correlations between plasticity loss, overestimation, overfitting metrics, and agent returns in various benchmark suites. These underscore the importance of considering environment-specific factors when assessing model performance and designing effective regularization strategies.
\end{itemize}

\end{tcolorbox}

\begin{tcolorbox}
\begin{itemize}
    \item Techniques like layer or spectral norm and resets are particularly effective in mitigating overestimation also compared to methods specifically designed for that purpose.
    \item The negative correlation of the critic's gradient norm and return becomes more apparent in challenging environments and mainly in a high replay ratio regime.
\end{itemize}

\end{tcolorbox}

\section{Related Works}
The literature on deep reinforcement learning has long explored various factors contributing to performance challenges, approaching the issue from different perspectives. A notable study, akin to our pragmatic approach, investigates the impact of diverse design choices in the training process of on-policy methods~\citep{andrychowicz2021matters}.

In the realm of off-policy methods, numerous hypotheses have been proposed regarding crucial factors. One key focus is on addressing the overestimation problem, with attempts to harness its potential benefits~\citep{ciosek2019better} and, more prominently, to mitigate the phenomenon by introducing novel loss functions~\citep{fujimoto2018addressing, moskovitz2021tactical, cetin2023learning}.

Regularization schemes have proven effective in enhancing deep reinforcement learning methods. Neural network regularizations, such as Spectral Norm~\citep{gogianu2021spectral, BjorckGW21}, Layer Norm~\citep{BjorckGW21, ball2023efficient}, or weight decay~\citep{liu2020regularization}, have yielded significant improvements in results. Notably, periodic resets of critic weights proposed by~\citep{nikishin2022primacy} constitute a strong baseline in robotics control. Several regularization schemes have been proposed to address the issue of discarding knowledge caused by fully resetting the critic network. Of particular interest is Shrink and Perturb~\citep{NEURIPS2020_288cd256} and L2 Init~\citep{kumar2023maintaining}. In both of these methods the benefit comes from the regularization towards the distribution of "freshly" initialized weights. A concurrent work finds that using unit-ball normalization allows for learning with a high-replay ratio without full-parameter resets \citep{hussing2024dissecting}. An ensemble approach was also suggested to address the challenge of determining the optimal number of gradients per environment step, where decisions are based on the validation TD error~\citep{li2022efficient}.

\section{Limitations}
In this work, we found crucial choices that drive SAC effectiveness in a wide range of control tasks from two popular benchmarks. Through extensive experiments, we uncovered perplexities concerning explanatory metrics correlations and complex dynamics of overestimation, which is successfully mitigated by widely used regularizations. Nevertheless, our study has certain constraints. Our empirical evaluations were limited to proprioceptive tasks on DMC and MetaWorld benchmarks and only for SAC method. 

\section{Conclusions}

This study explored different common RL design choices, as well as interactions thereof, evaluating their impact on agent learning. Specifically, we consider three types of regularization families: critic regularization (motivated by value overestimation); network regularization (motivated by model overfitting); and plasticity regularization (motivated by plasticity loss). Our analysis revealed that generic network regularization methods such as layer normalization, especially when paired with full-parameter resets, can have a vastly greater impact on the final performance than domain-specific RL approaches. To this end, the same network regularization methods can lead to strong performing policies on domains previously solved by model-based agents, such as the dog domain. Furthermore, we studied a variety of metrics that were shown to co-occur with the deterioration of learning in low and high replay regimes. Our analysis revealed the complex interactions of the considered metrics with agent performance. Surprisingly, we found that interventions motivated by a specific problem, for example, overfitting, can have a pronounced impact on the metrics associated with overestimation or plasticity. Finally, we found that the effectiveness of considered critic, network, and plasticity regularization techniques is not only highly dependent on the simulation benchmark but also type of simulated task. The most prominent example is Clipped Double Q-learning, a technique used in a majority of modern actor-critic algorithms, which is effective in DMC locomotion tasks, but leads to significant performance deterioration on the MetaWorld manipulation tasks. To this end, we highlighted the need to test new algorithms on a diverse set of tasks, preferably stemming from more than one suite. 

\section*{Acknowledgements}

Marek Cygan was partially supported by an NCBiR grant POIR.01.01.01-00-0433/20. Mateusz Ostaszewski was funded by the National Science Center Poland under the grant agreement 2020/39/B/ST6/01511. Michał Bortkiewicz was funded by the Warsaw University of Technology within the Excellence Initiative: Research University (IDUB) programme. This research was also supported by National Science Centre, Poland grant no 2020/39/B/ST6/01511 and grant no 2022/45/B/ST6/02817. This work was partially funded by the European Union under the Horizon Europe grant OMINO (grant number 101086321) and Horizon Europe Program (HORIZON-CL4-2022-HUMAN-02) under the project "ELIAS: European Lighthouse of AI for Sustainability", GA no. 101120237. We also gratefully acknowledge Polish high-performance computing infrastructure PLGrid (HPC Center: ACK Cyfronet AGH) for providing computer facilities and support within computational grant no. PLG/2023/016783. 


\section*{Impact Statement}

This paper focuses on the issue of interplay between different design choices within Reinforcement Learning (RL) algorithms. While the successful application of RL has the potential to influence society in many ways, our work, focused primarily on a technical advancement in RL algorithms, does not introduce novel ethical considerations beyond those already inherent in the broader field of RL.

\bibliography{bib}
\bibliographystyle{icml2024}

\newpage
\appendix
\onecolumn

\section{Details of experiments}
\label{app:experim_details}

Results reporting Interquartile mean (IQM) are based on 500 bootstrapping points as calculated by \textit{rliable} package~\citep{agarwal2021deep}. The final performance is defined as the average of the last 10 policy evaluations.

\section{Architecture details}
\label{app:architecture_details}
In all experiments, the Actor and Critic are represented by three-layer MLP networks, each containing 256 neurons in the hidden layers, utilizing the ReLU activation function (except for the CReLU variant, which effectively doubles the number of activations).

In the scenario involving Layer Norm, it is applied to each hidden layer~\citep{li2022efficient,ball2023efficient}, while the spectral norm is applied exclusively to the last hidden layer~\citep{gogianu2021spectral,li2022efficient}. Weight decay is uniformly applied across all layers~\citep{li2022efficient}. It's important to note that all network regularizations are exclusively applied to the Critic network.

\subsection{Hyperparameters}
All hyperparameters are taken from original papers introducing the given intervention. 
\begin{table}[ht!]
\centering
\caption{Hyperparameter values used in the experiments.}
\label{table:3}
{\renewcommand{\arraystretch}{1.03}%
 \begin{tabular}{||c |c | c ||} 
 \hline
 \textsc{Hyperparameter} & \textsc{Notation} & \textsc{Value} \\
 \hline \hline
  \multicolumn{3}{||c||}{\textsc{Joint}} \\
 \hline \hline
 \textsc{Network Size} & \textsc{na} & $(256, 256)$ \\
 \textsc{Action Repeat} & \textsc{na} & $1$ \\
 \textsc{Optimizer} & \textsc{na} & \textsc{Adam} \\
 \textsc{Learning Rate} & \textsc{na} & $3e-4$ \\
 \textsc{Batch Size} & $B$ & $256$ \\
 \textsc{Discount} & $\gamma$ & $0.99$ \\
 \textsc{Initial Temperature} & $\alpha_{0}$ & $1.0$ \\
 \textsc{Initial Steps} & $\textsc{na}$ & $10000$ \\
 \textsc{Target Entropy} & $\mathcal{H}^{*}$ & $|\mathcal{A}|/2$ \\
 \textsc{Polyak Weight} & $\tau$ & $0.005$ \\
 
 \hline \hline
 \multicolumn{3}{||c||}{\textsc{TOP}} \\
 \hline \hline
 \textsc{Pessimism Values} & $\beta$ & $\{0,1\}$ \\
 \textsc{Bandit Learning Rate} & \textsc{na} & $0.1$ \\
 \hline \hline
 \multicolumn{3}{||c||}{\textsc{GPL}} \\
 \hline \hline
 \textsc{Pessimism Learning Rate} & \textsc{na} & $1e-5$ \\
 \hline \hline
 \end{tabular}}
\end{table}
\label{appendix1}

\section{Further Experiments}

\subsection{Third-order marginalization}
\label{appendix:third_order}
\begin{figure*}[ht!]
	\begin{center}
        \begin{minipage}[h]{1.0\linewidth}
                \begin{subfigure}{1.0\linewidth}
                        \includegraphics[width=0.49\linewidth]{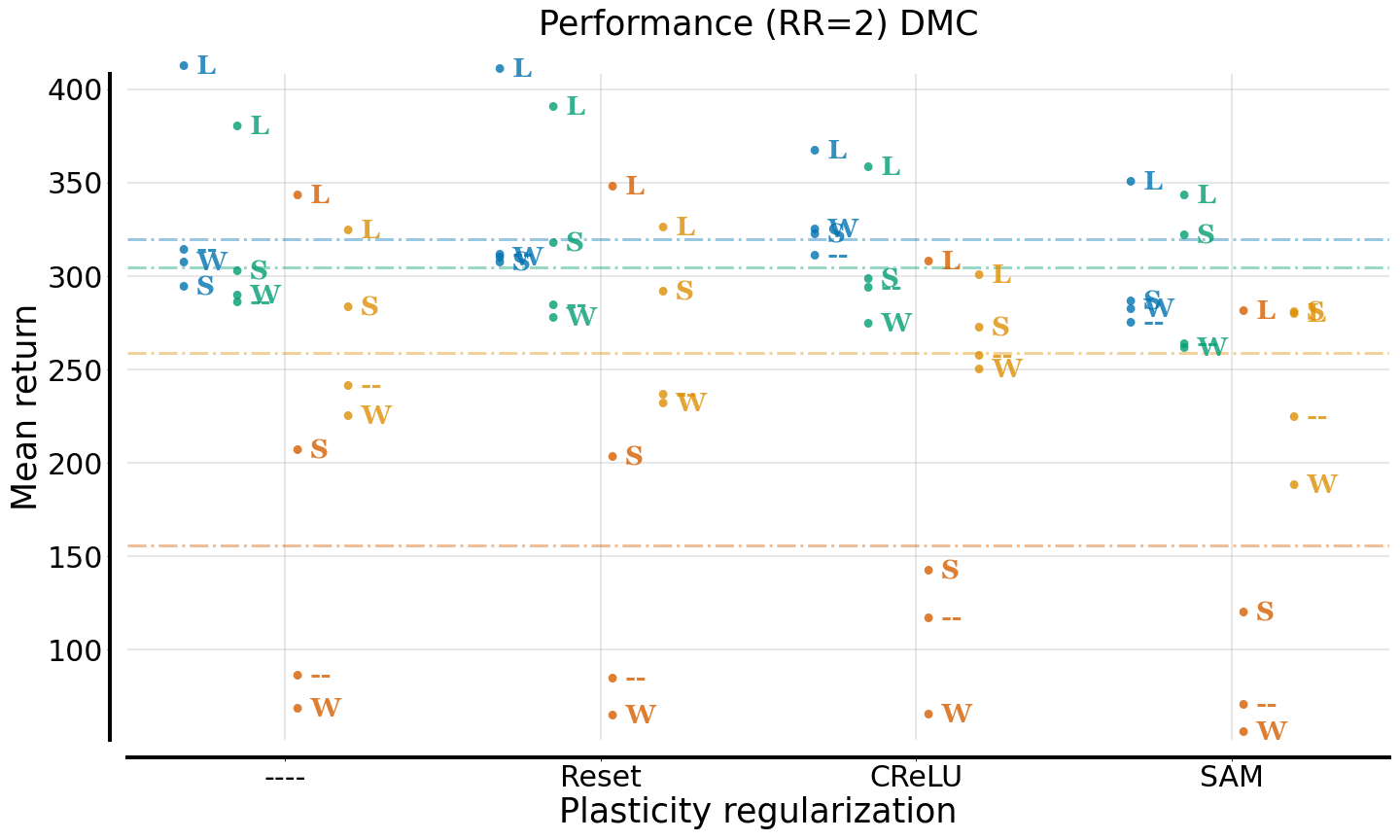}
                        \hfill
                        \includegraphics[width=0.49\linewidth]{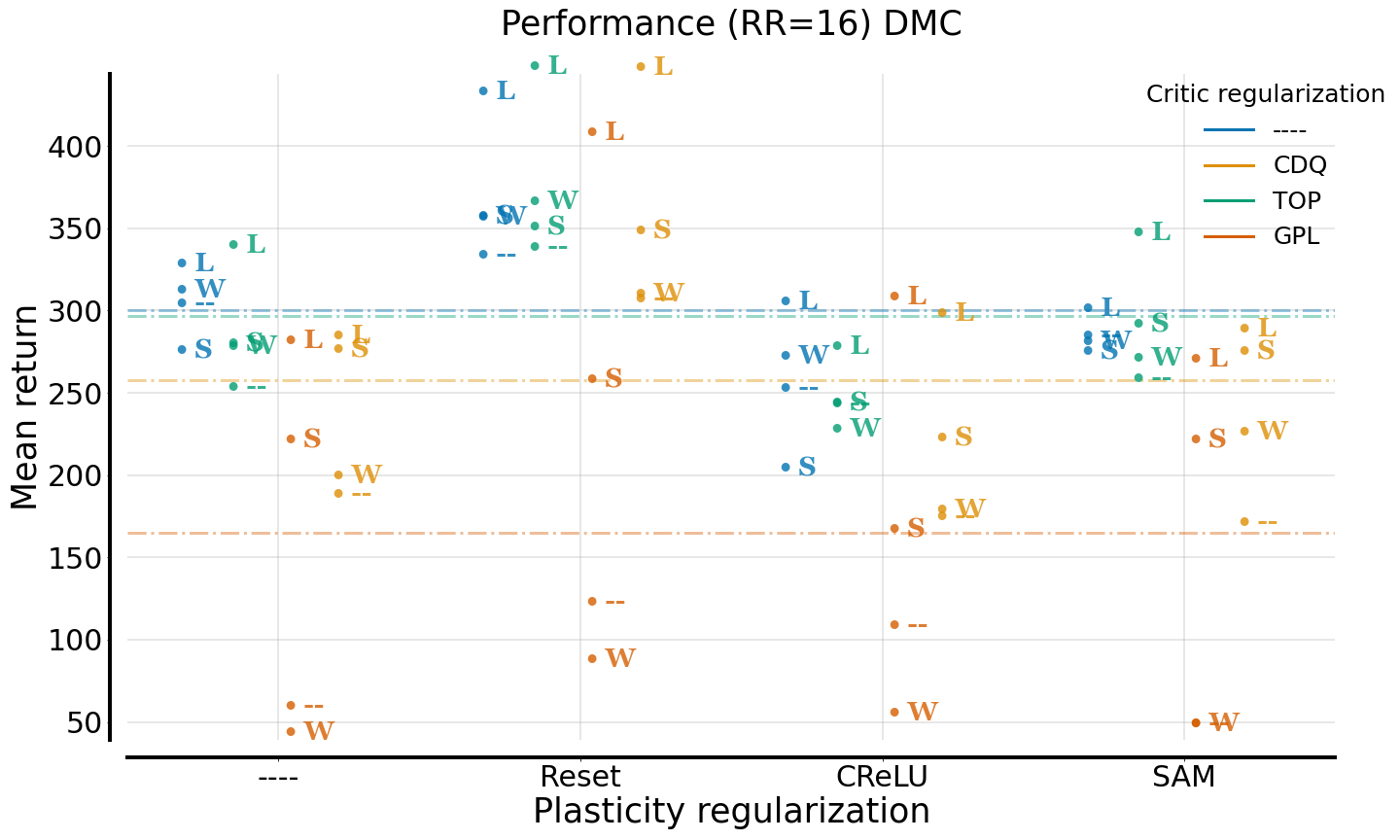}
                    \end{subfigure}
            \end{minipage}
        \begin{minipage}[h]{1.0\linewidth}
                \begin{subfigure}{1.0\linewidth}
                        \includegraphics[width=0.49\linewidth]{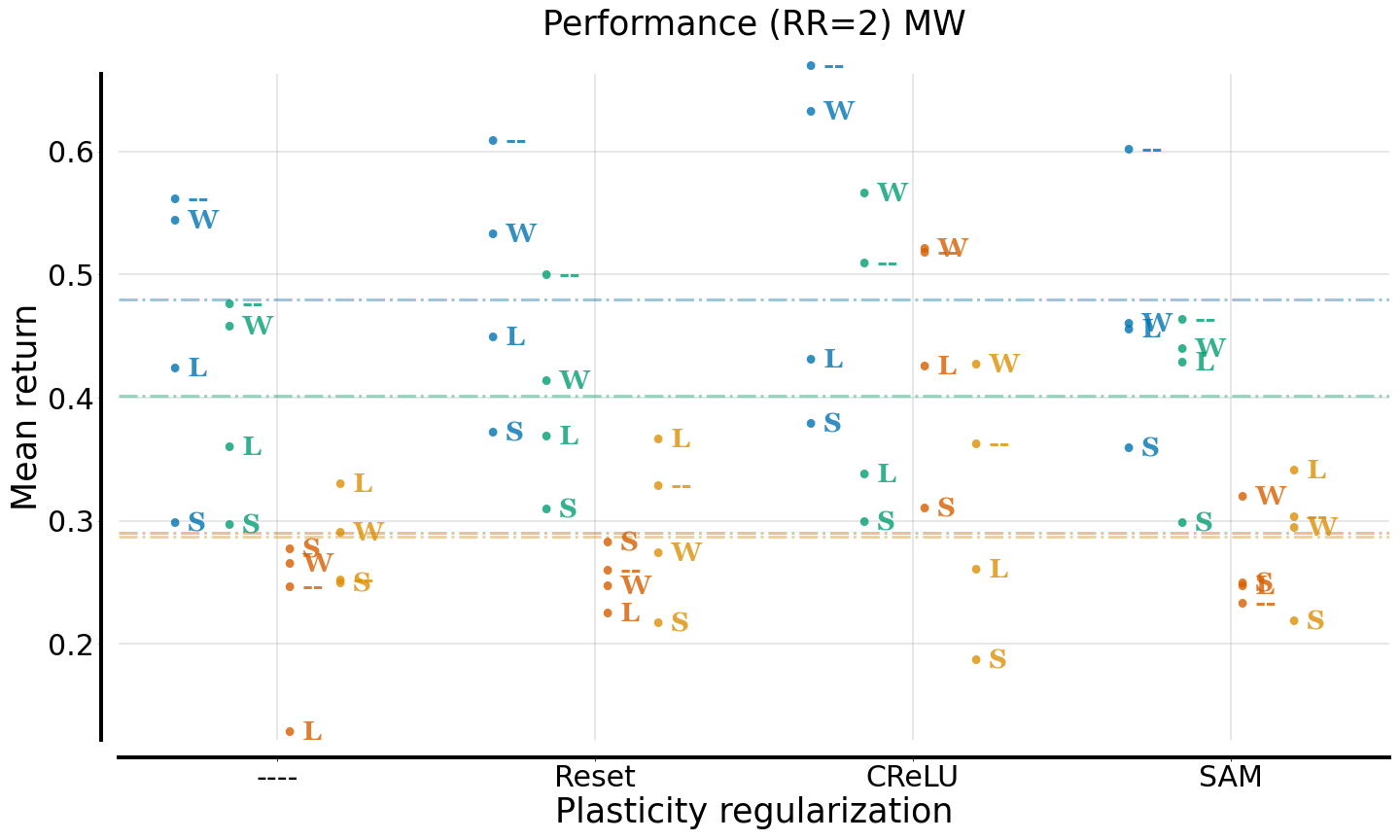}
                        \hfill
                        \includegraphics[width=0.49\linewidth]{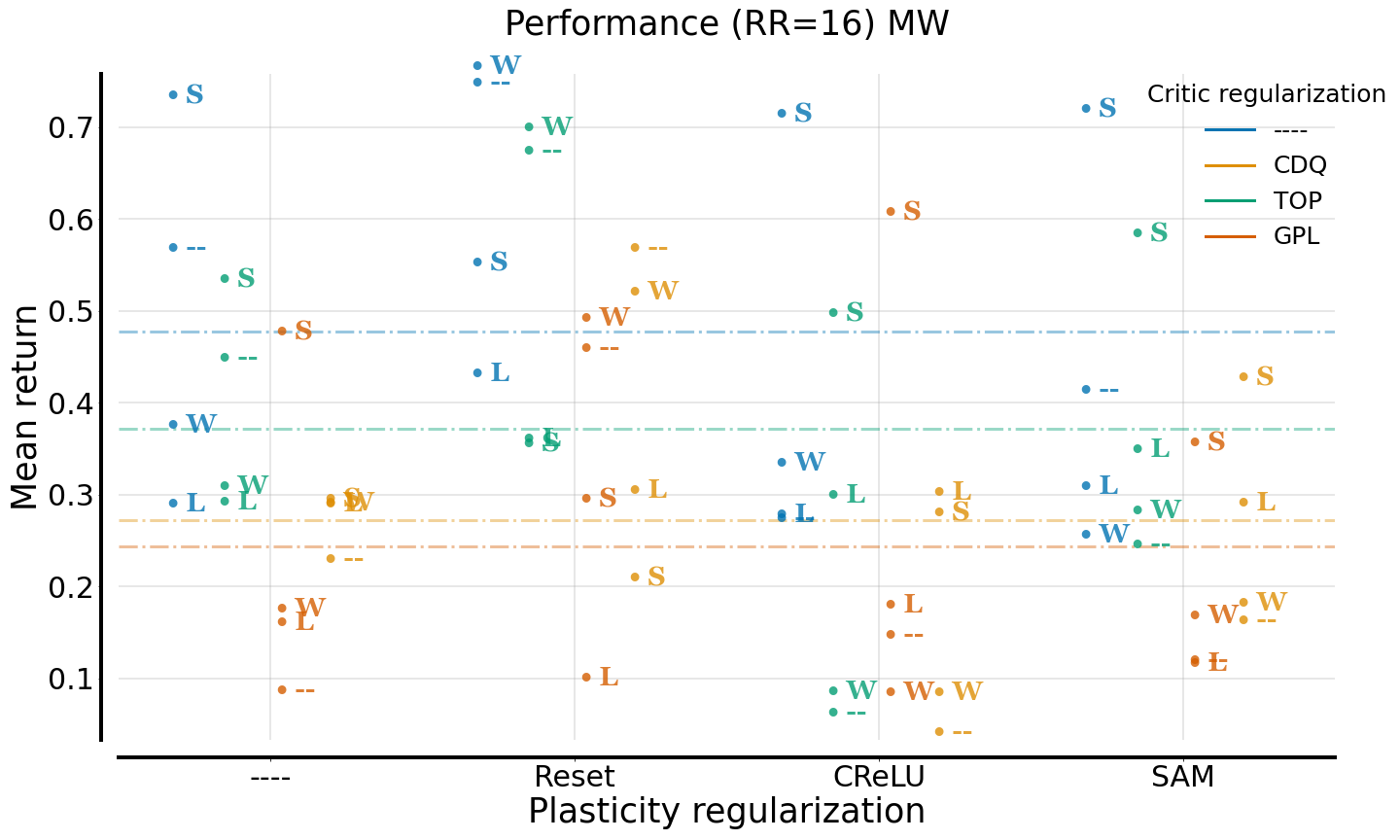}
                    \end{subfigure}
            \end{minipage}
        \caption{DMC (top) MW (bottom).}
        \label{fig:second_interv}
    \end{center}
\end{figure*}

Examining the plots without marginalization (Figure~\ref{fig:second_interv}) provides further insights into the conclusions drawn from the previous experiment. Specifically, most combinations without critic regularization (red points) consistently perform well across all setups. Additionally, the results for GPL (pink points) affirm the overall subpar performance of this method.
On the DMC with RR=2 plot (top-left in Figure~\ref{fig:second_interv}), layer norm (points labeled "L") consistently outperforms others in mean return, irrespective of the Plasticity regularization (x-axis) or Critic Regularization (color). Notably, the combination of layer norm and resets in RR=16 (top-right plot in Figure~\ref{fig:second_interv}) demonstrates exceptional performance across all critic regularization variations.

For the MW benchmark with RR=16 (bottom-right plot in Figure~\ref{fig:second_interv}), the results align with first-order marginalization findings, highlighting the positive impact of Spectral Normalization. Particularly interesting is the role of Weight Decay, forming optimal combinations with the Reset method on RR=16 and consistently performing well across various combinations on RR=2 (bottom-left plot in Figure~\ref{fig:second_interv}). 

\begin{figure*}[ht!]
\begin{center}
\begin{minipage}[h]{1.0\linewidth}
    \begin{subfigure}{1.0\linewidth}
    \includegraphics[width=0.19\linewidth]{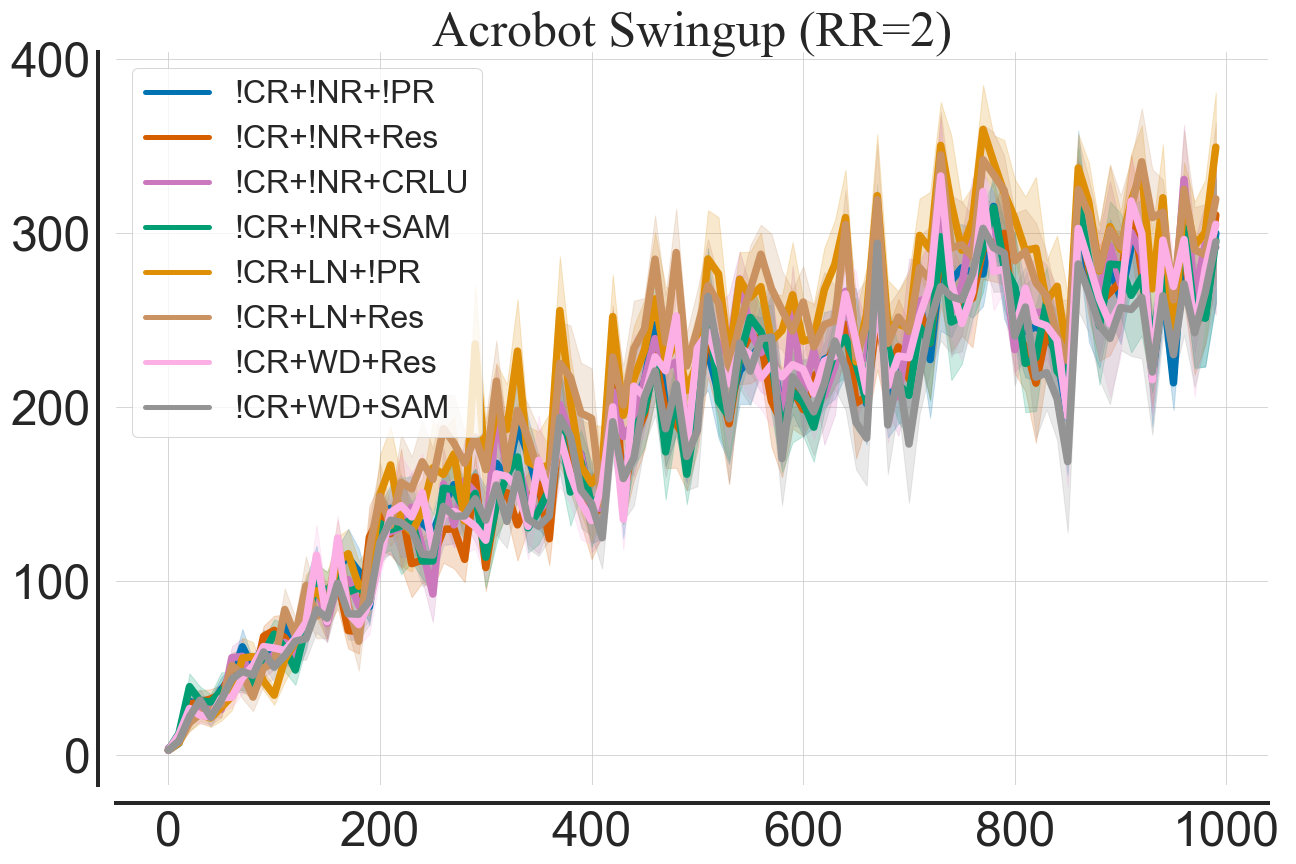}
    \hfill
    \includegraphics[width=0.19\linewidth]{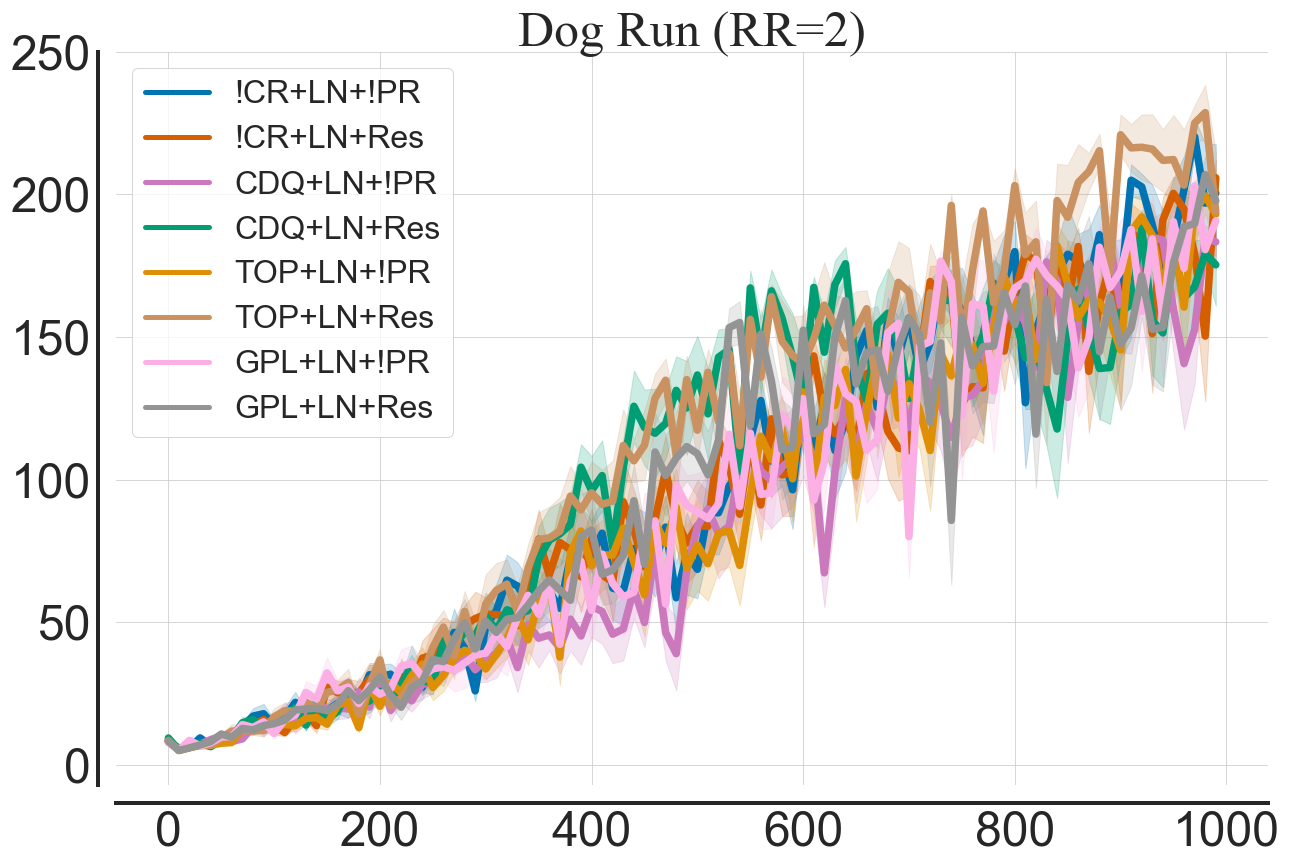}
    \hfill
    \includegraphics[width=0.19\linewidth]{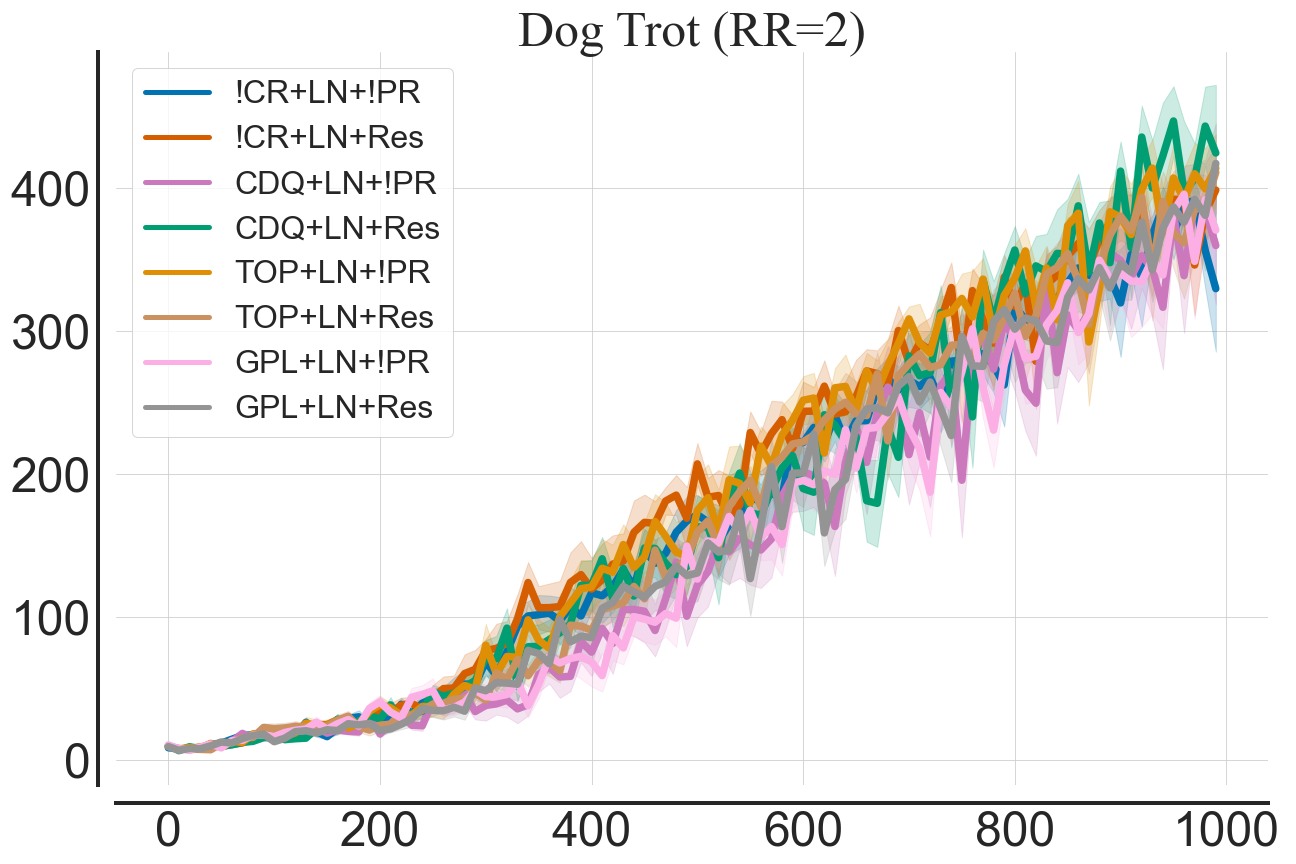}
    \hfill
    \includegraphics[width=0.19\linewidth]{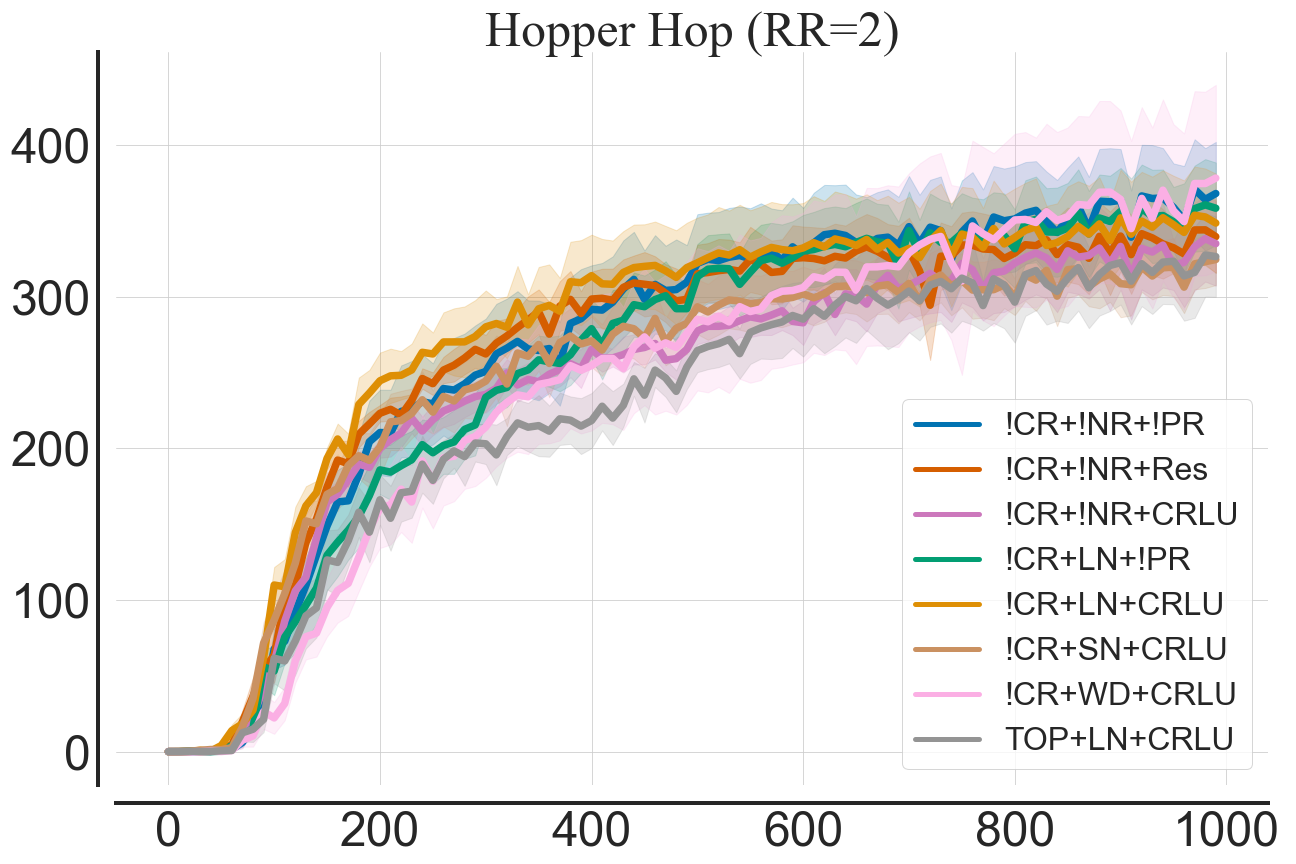}
    \hfill
    \includegraphics[width=0.19\linewidth]{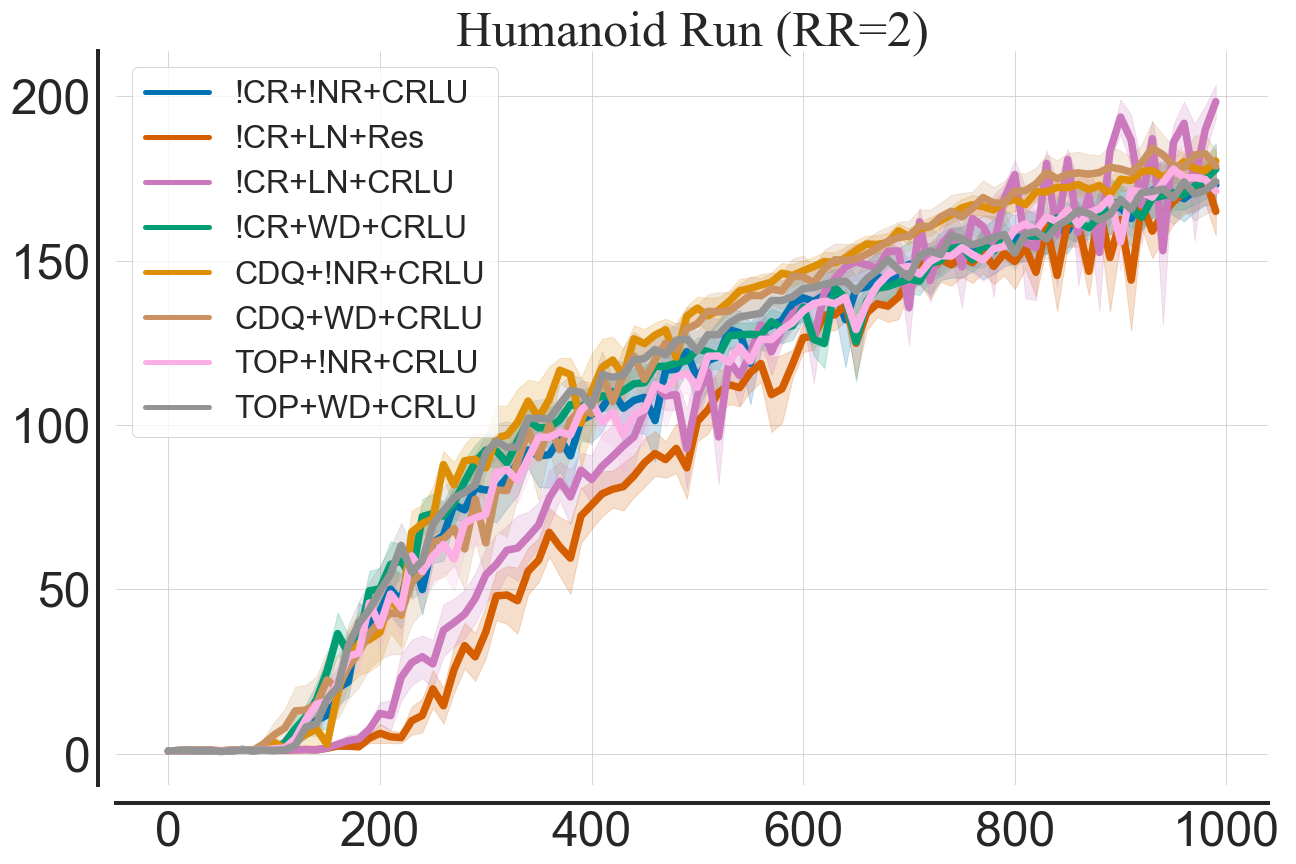}
    \end{subfigure}
\end{minipage}
\begin{minipage}[h]{1.0\linewidth}
    \begin{subfigure}{1.0\linewidth}
    \includegraphics[width=0.19\linewidth]{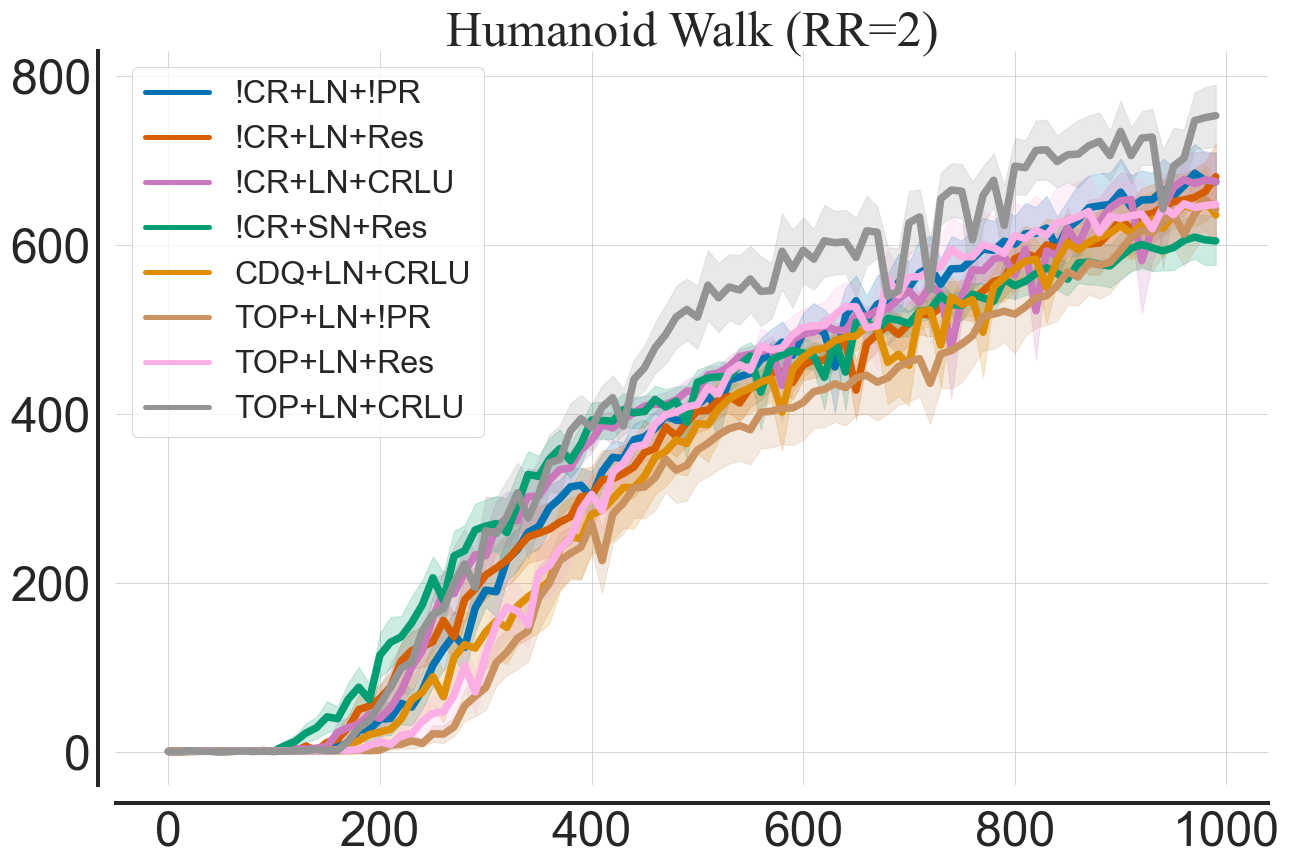}
    \hfill
    \includegraphics[width=0.19\linewidth]{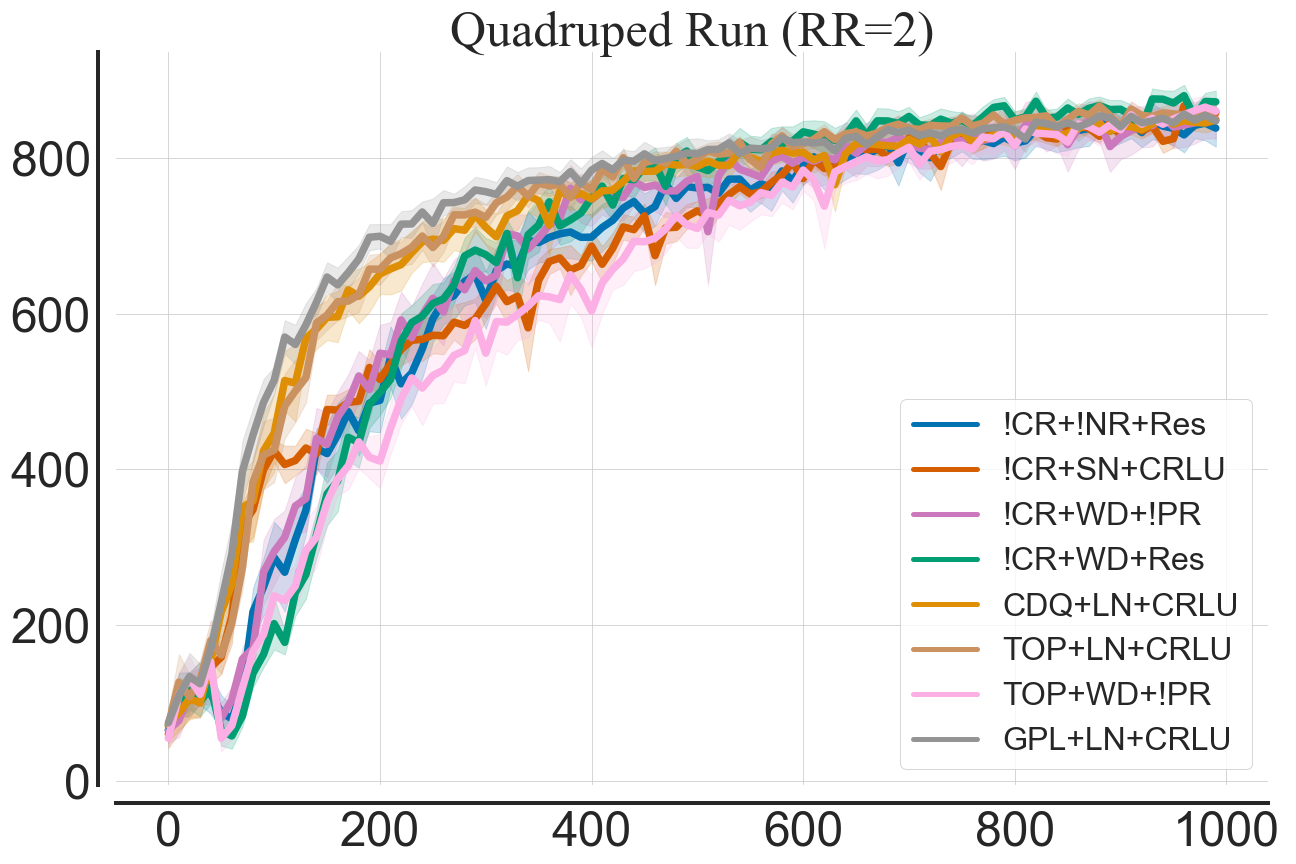}
    \hfill
    \includegraphics[width=0.19\linewidth]{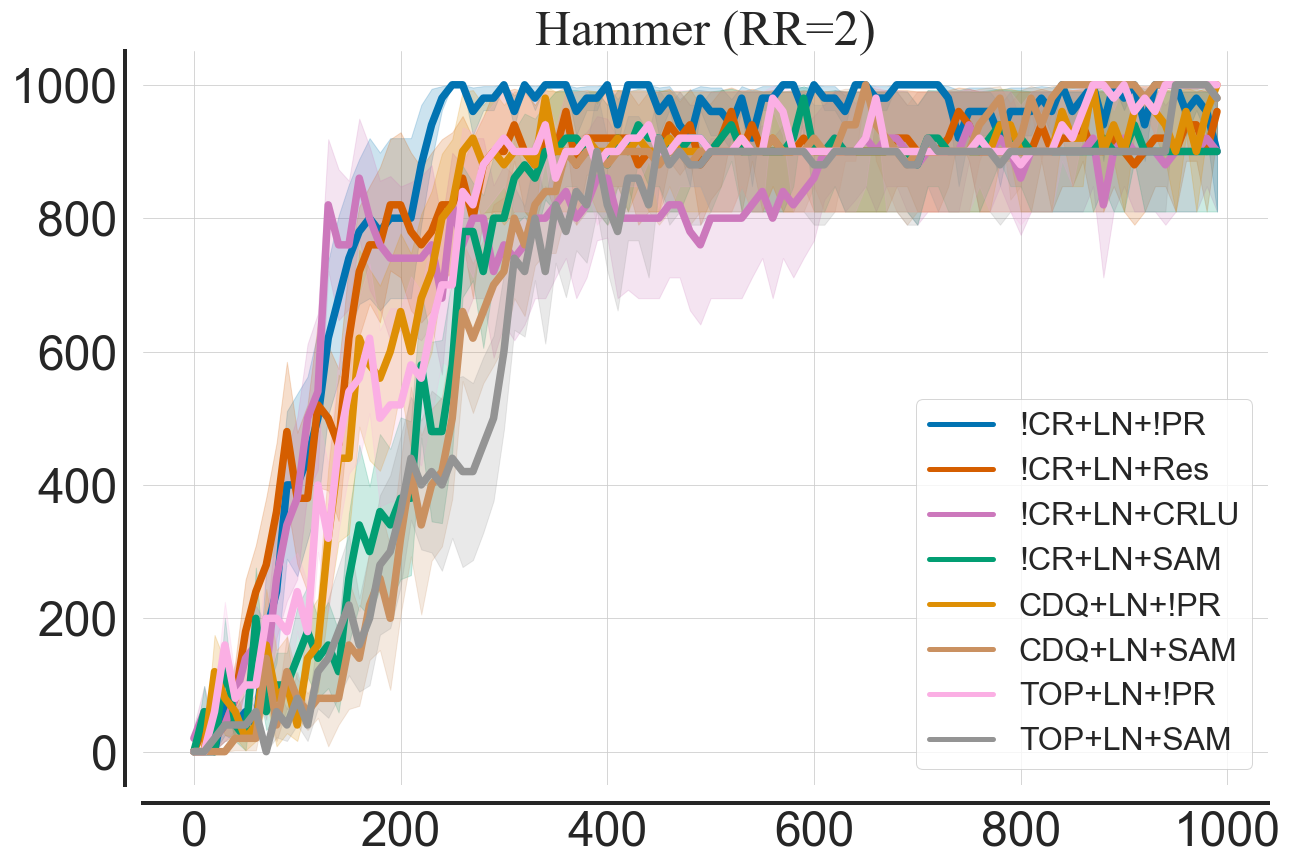}
    \hfill
    \includegraphics[width=0.19\linewidth]{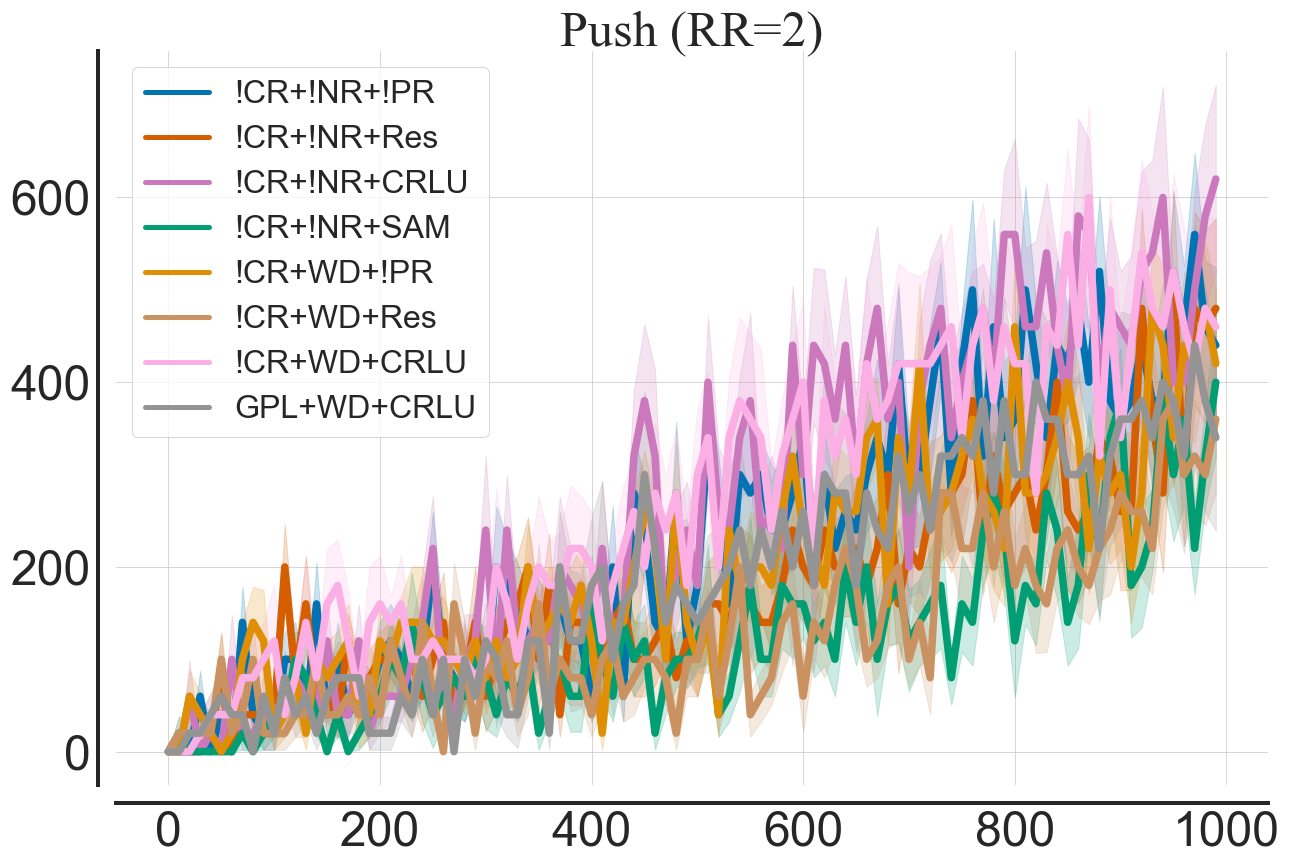}
    \hfill
    \includegraphics[width=0.19\linewidth]{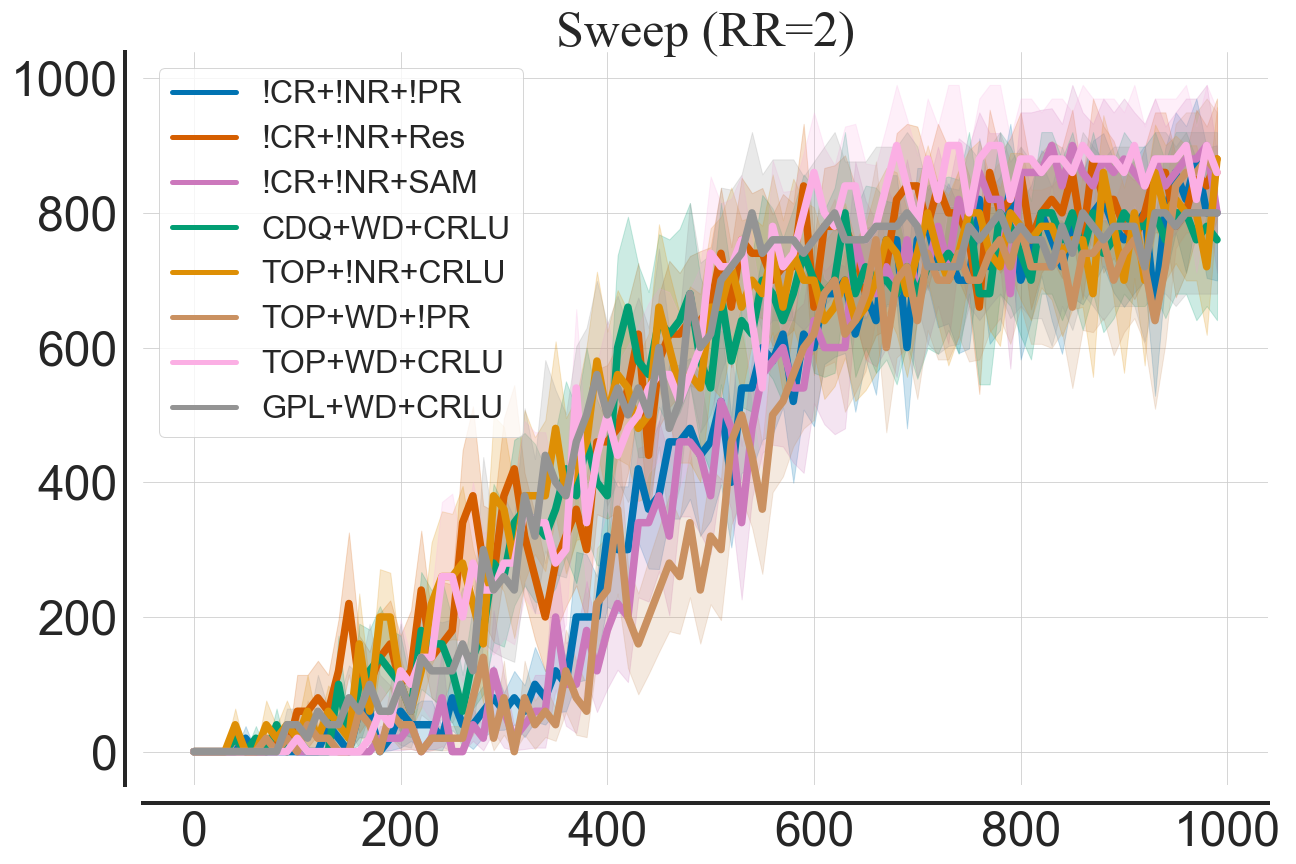}
    \end{subfigure}
\end{minipage}
\begin{minipage}[h]{1.0\linewidth}
    \begin{subfigure}{1.0\linewidth}
    \includegraphics[width=0.19\linewidth]{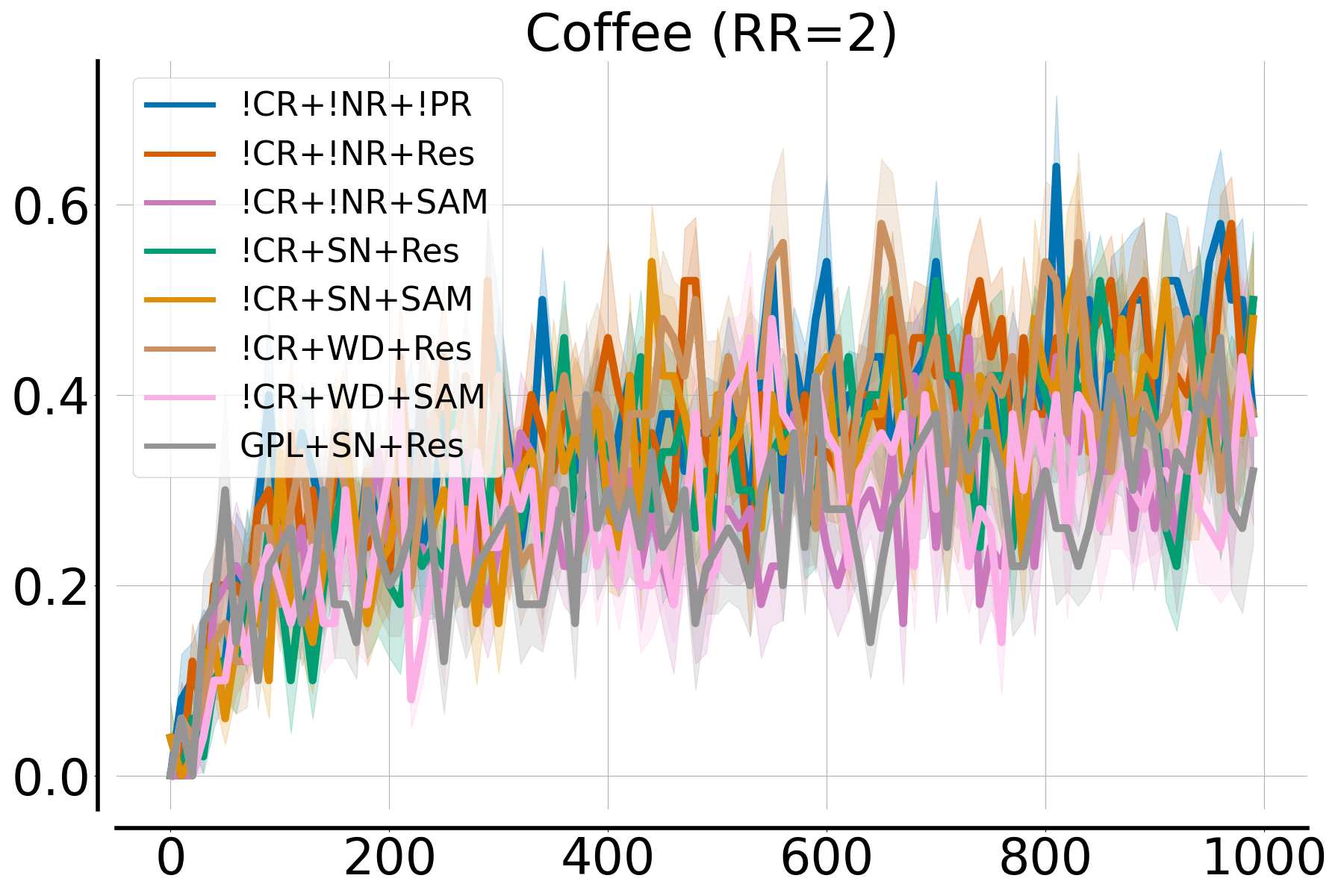}
    \hfill
    \includegraphics[width=0.19\linewidth]{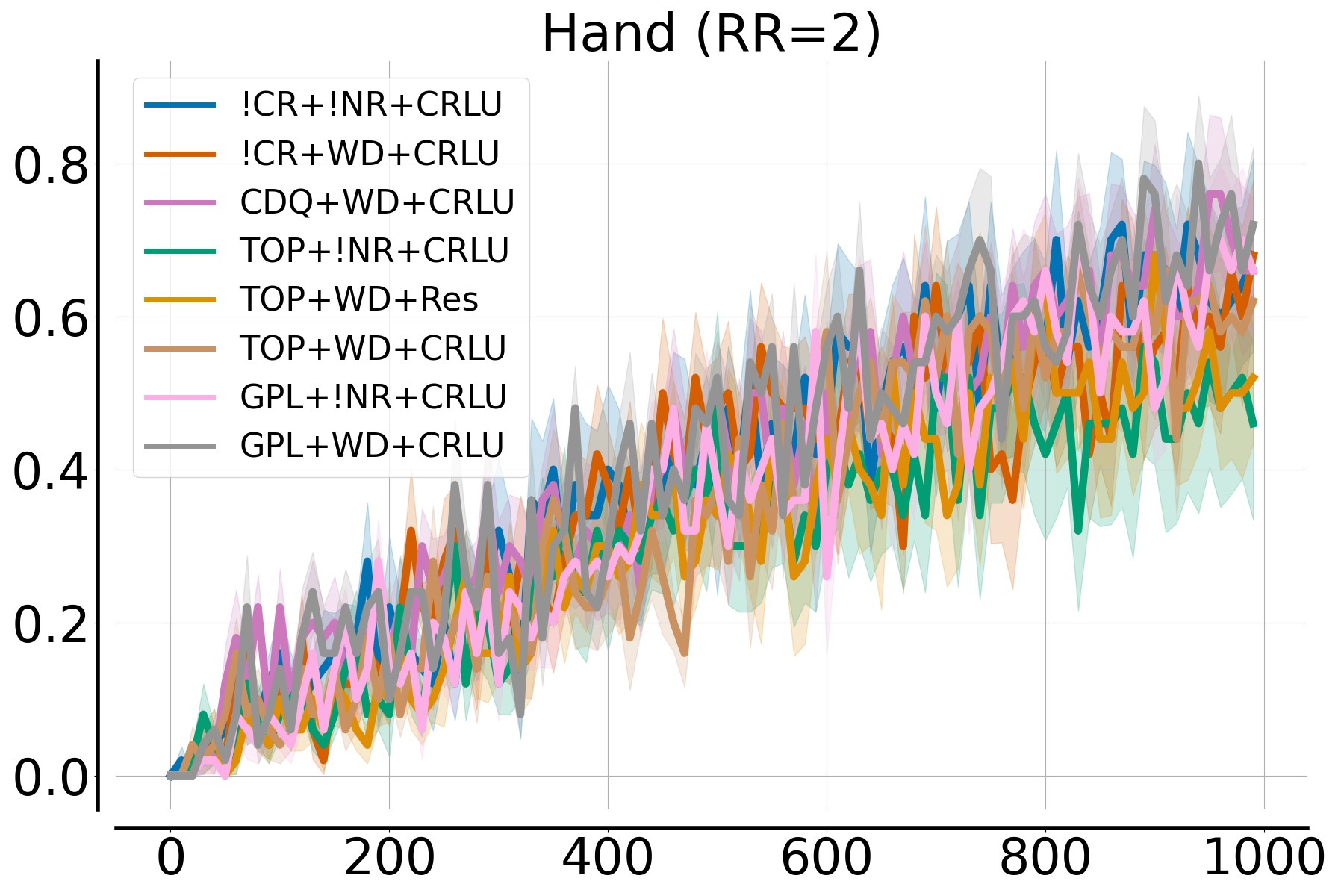}
    \hfill
    \includegraphics[width=0.19\linewidth]{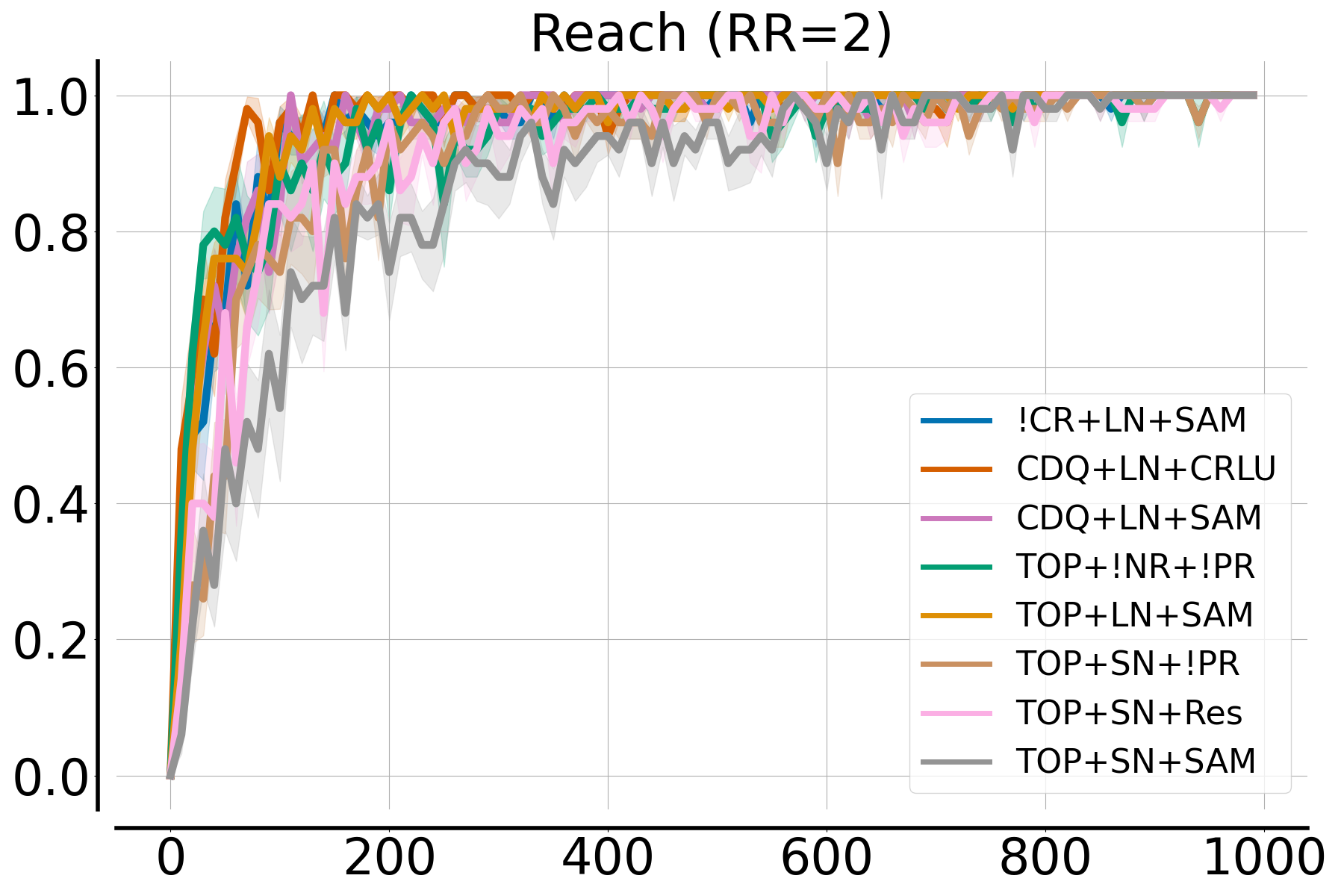}
    \hfill
    \includegraphics[width=0.19\linewidth]{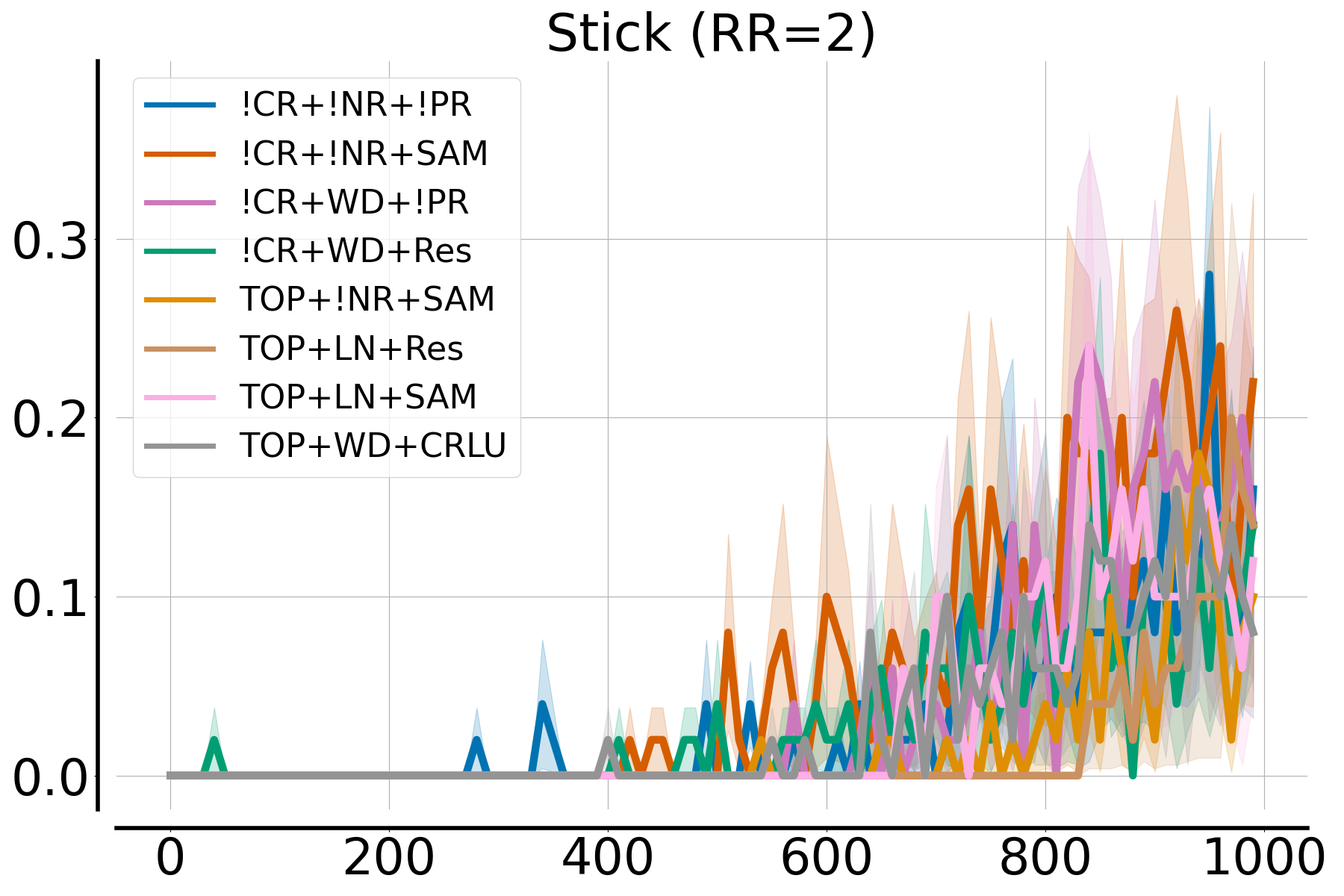}
    \end{subfigure}
\end{minipage}
\caption{Top performing configuration in the low replay regime. 10 seeds per task per algorithm.}
\label{fig:task_specific_rr2}
\end{center}
\end{figure*}

\begin{figure*}[ht!]
\begin{center}
\begin{minipage}[h]{1.0\linewidth}
    \begin{subfigure}{1.0\linewidth}
    \includegraphics[width=0.19\linewidth]{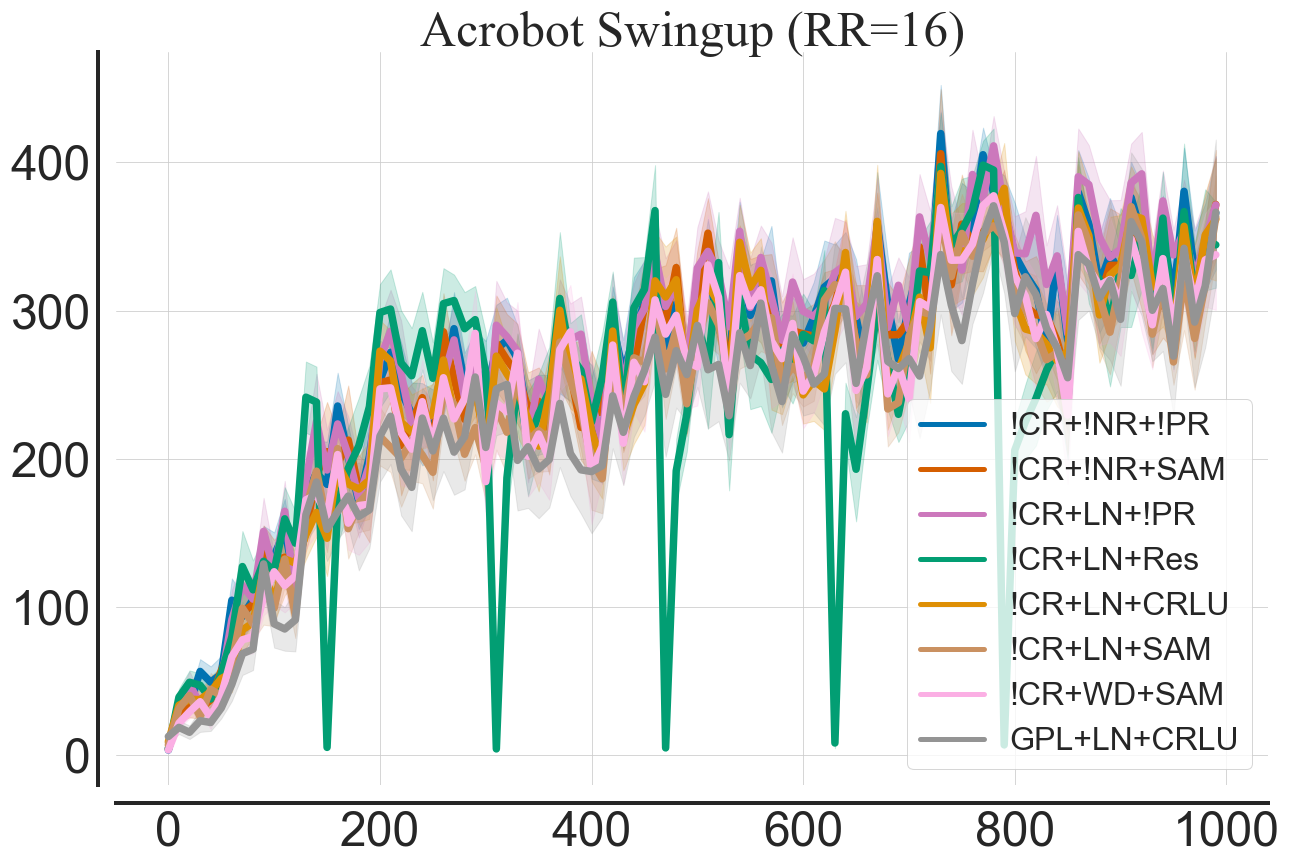}
    \hfill
    \includegraphics[width=0.19\linewidth]{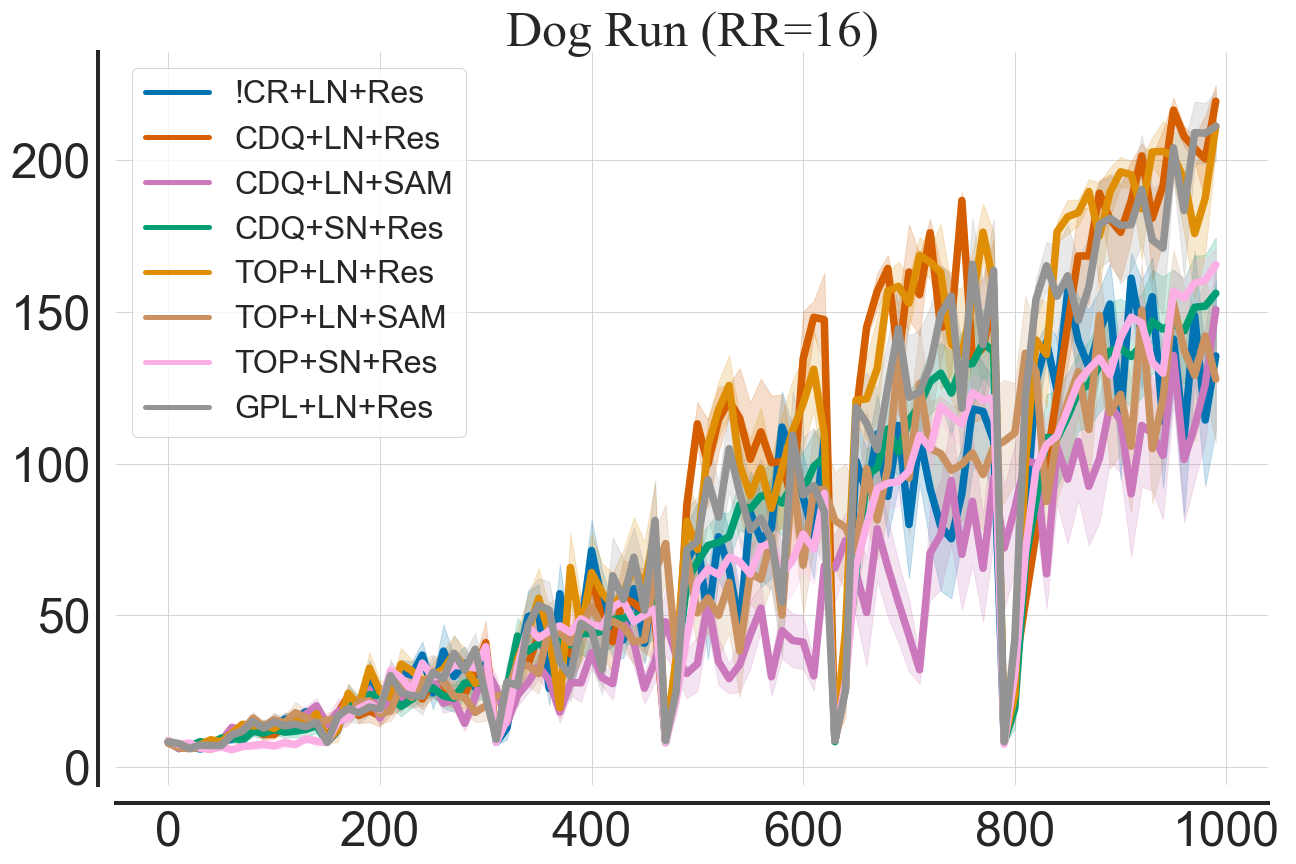}
    \hfill
    \includegraphics[width=0.19\linewidth]{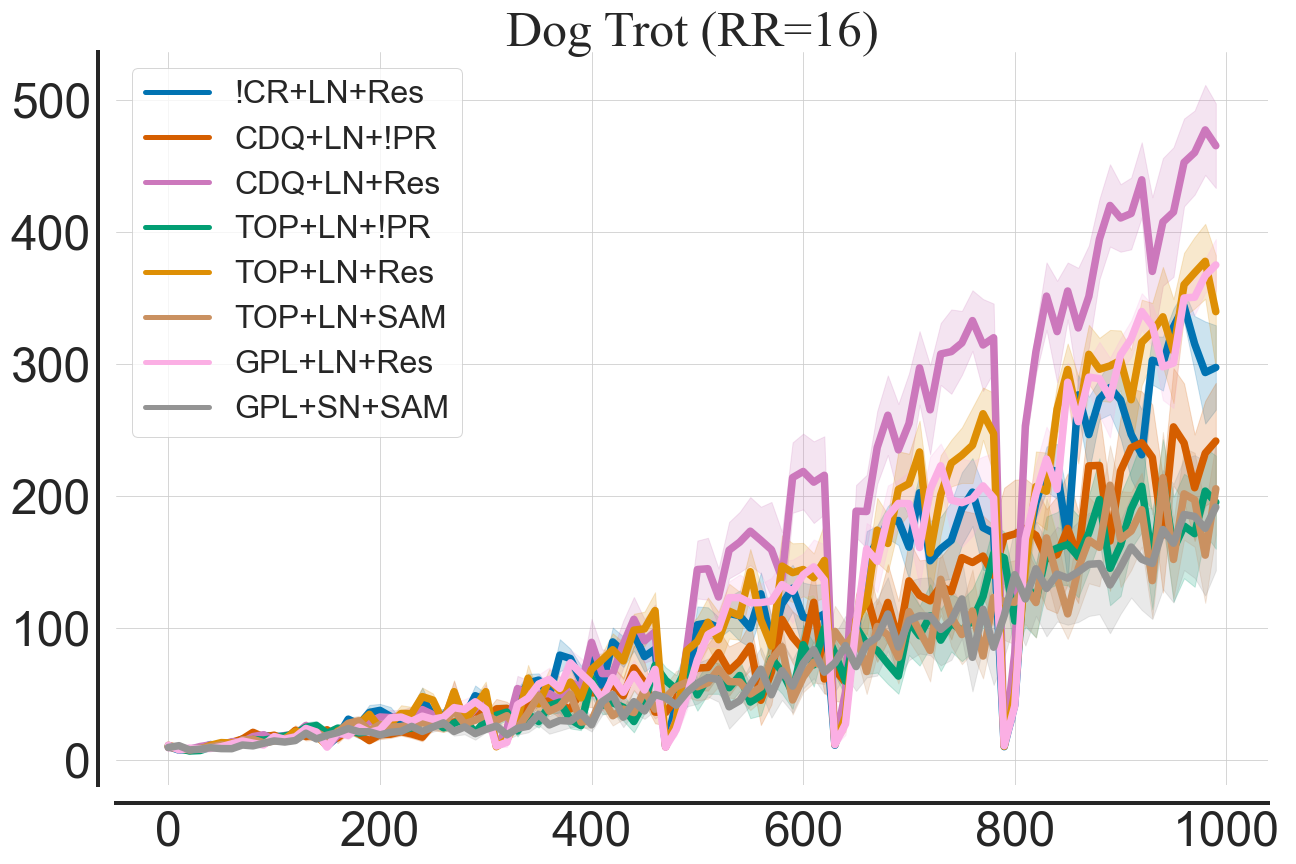}
    \hfill
    \includegraphics[width=0.19\linewidth]{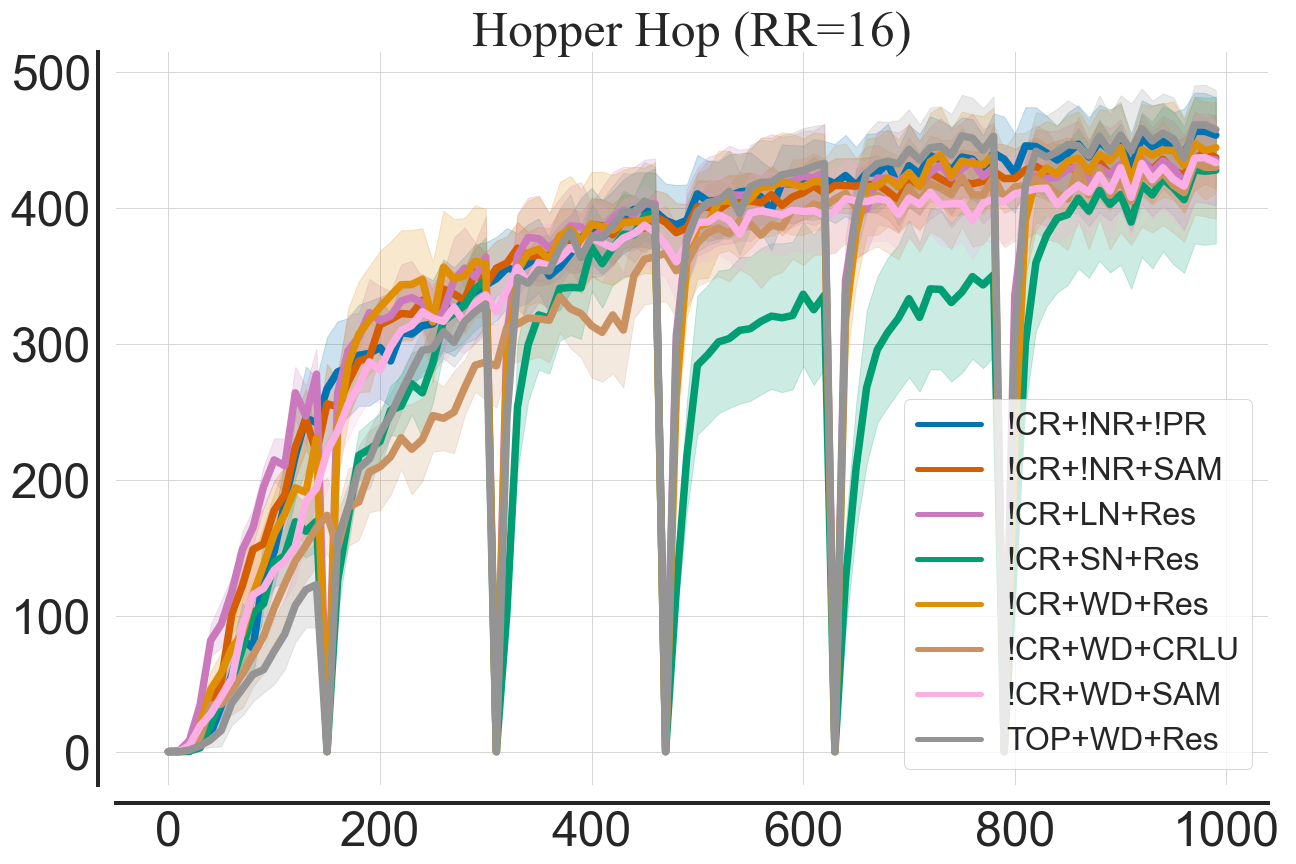}
    \hfill
    \includegraphics[width=0.19\linewidth]{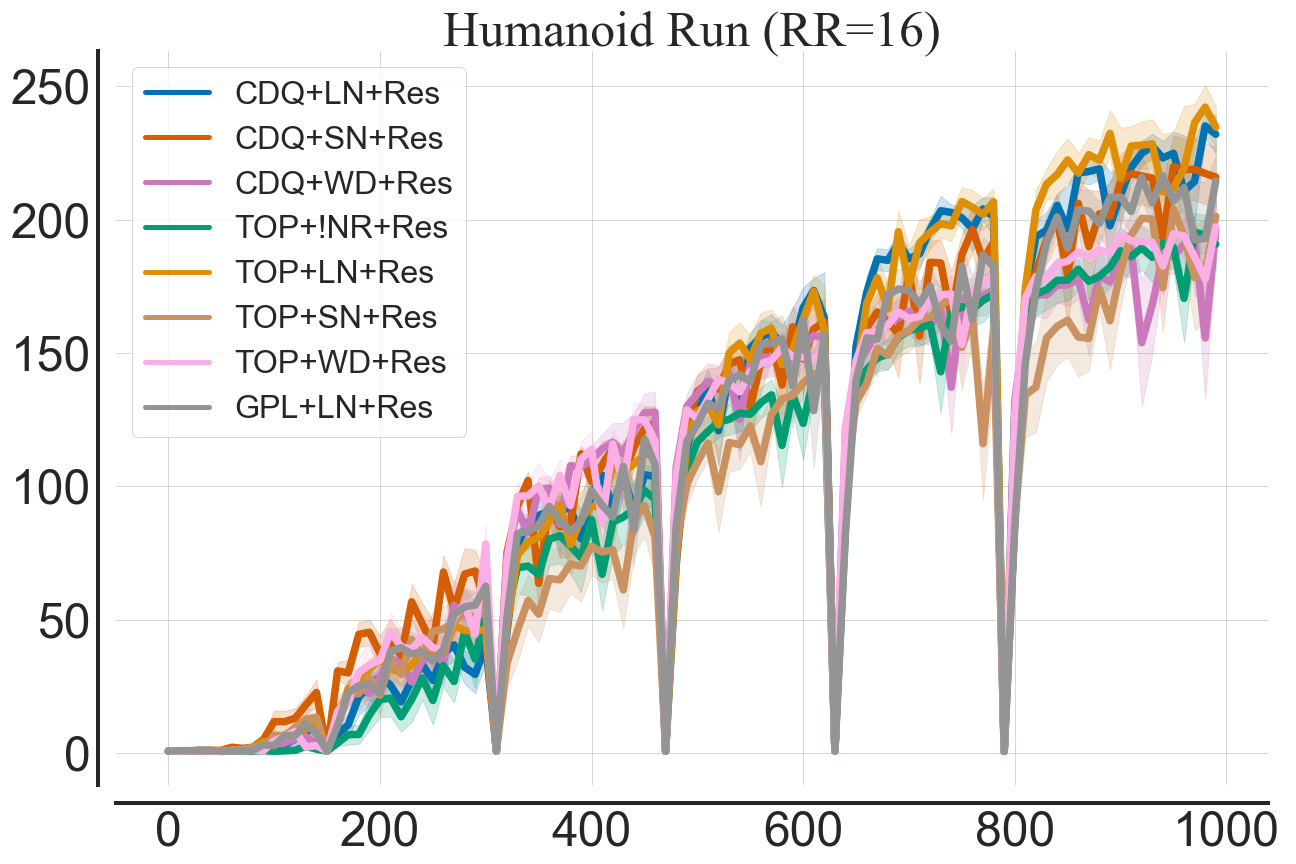}
    \end{subfigure}
\end{minipage}
\begin{minipage}[h]{1.0\linewidth}
    \begin{subfigure}{1.0\linewidth}
    \includegraphics[width=0.19\linewidth]{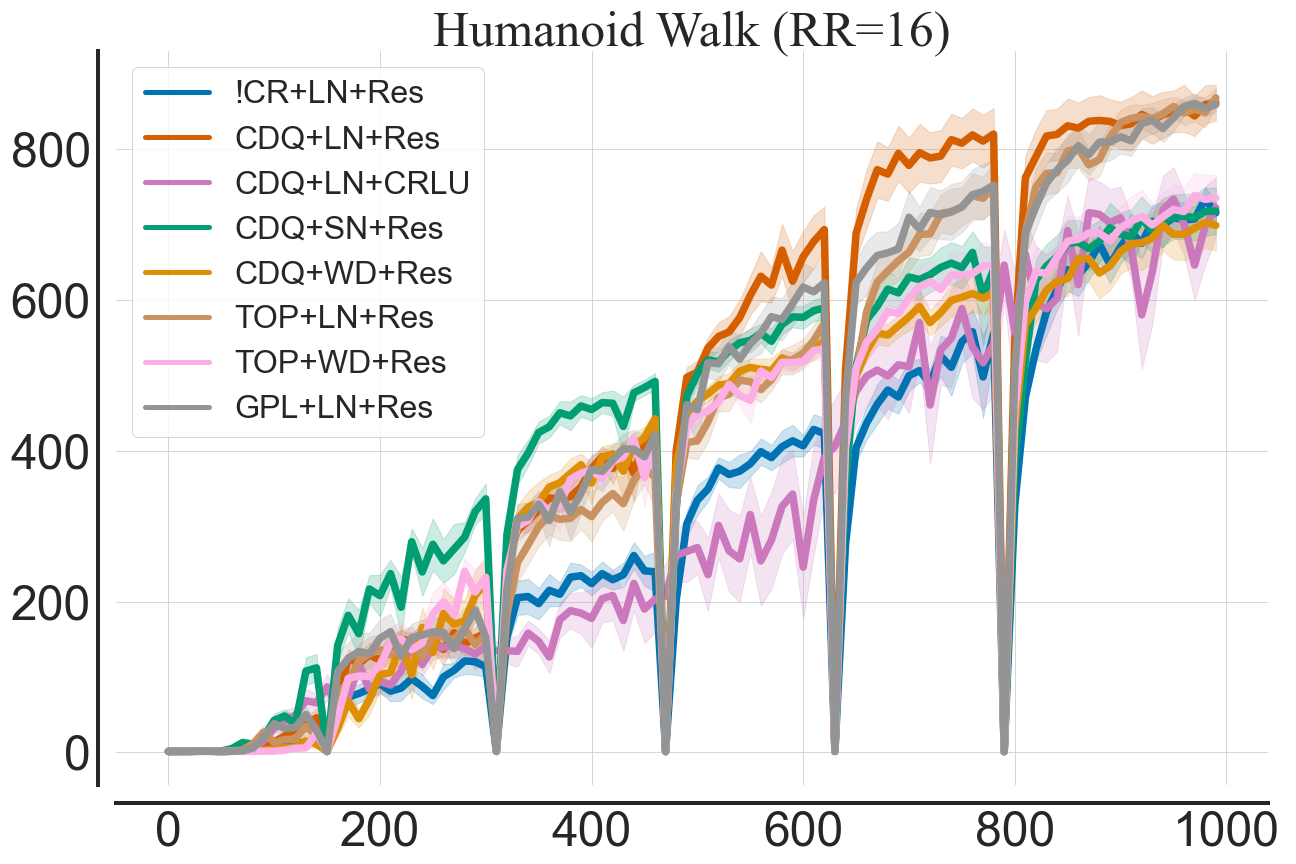}
    \hfill
    \includegraphics[width=0.19\linewidth]{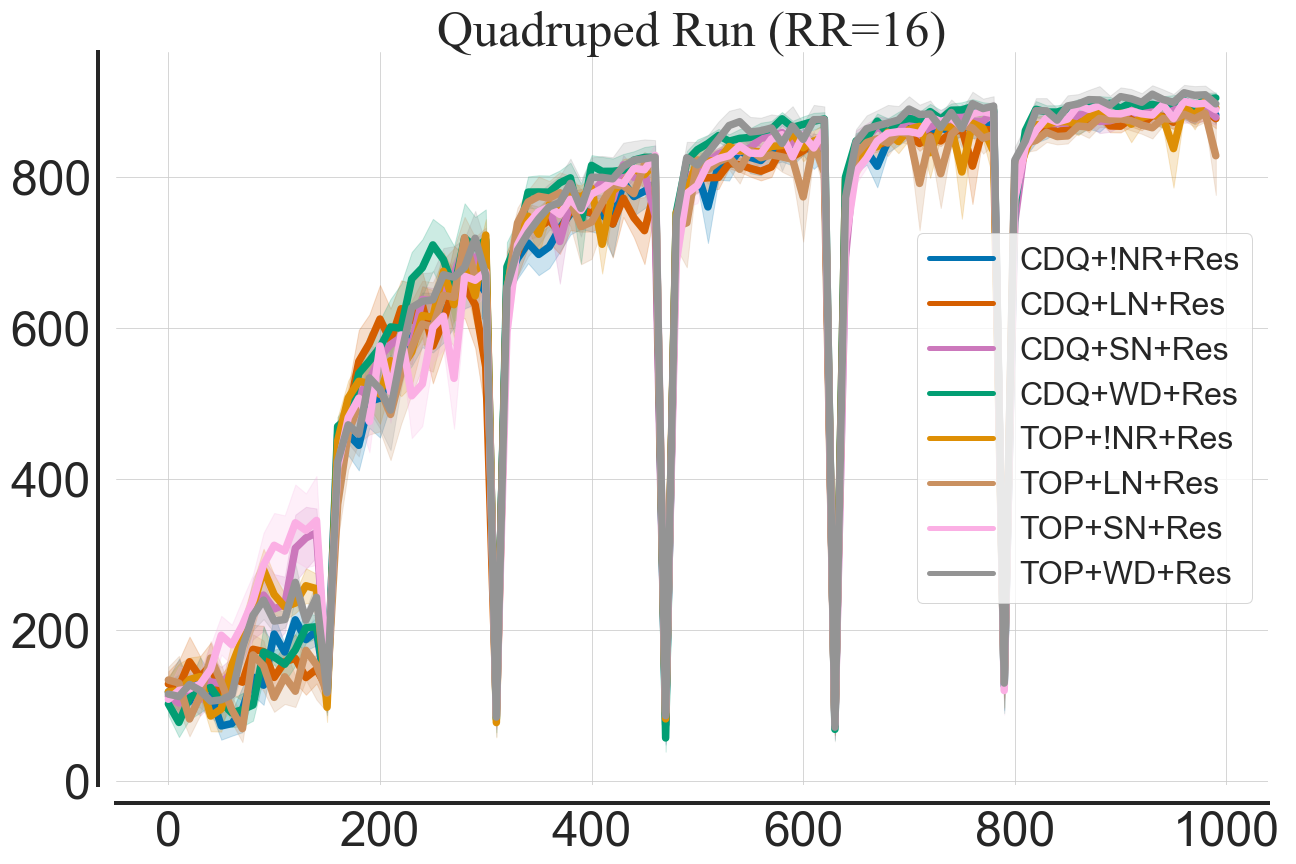}
    \hfill
    \includegraphics[width=0.19\linewidth]{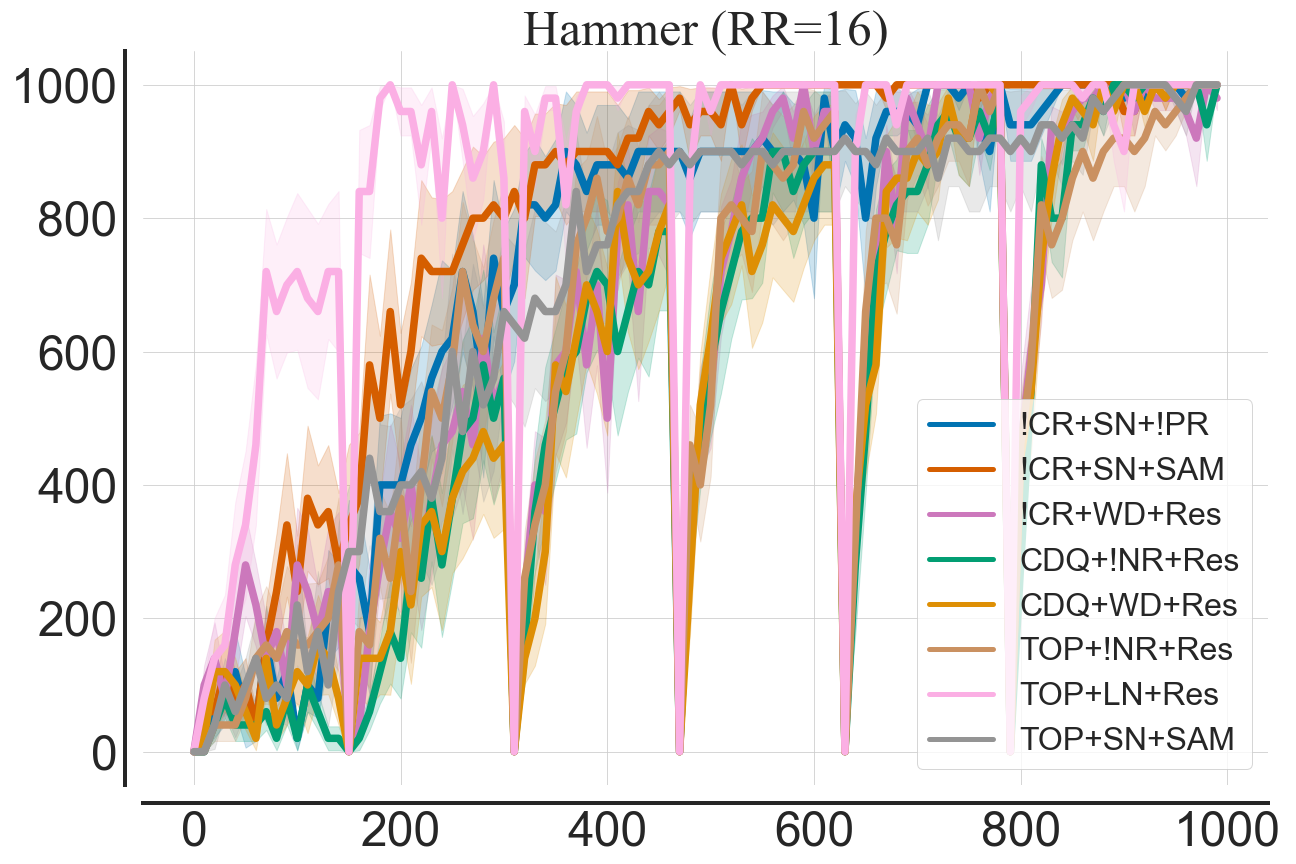}
    \hfill
    \includegraphics[width=0.19\linewidth]{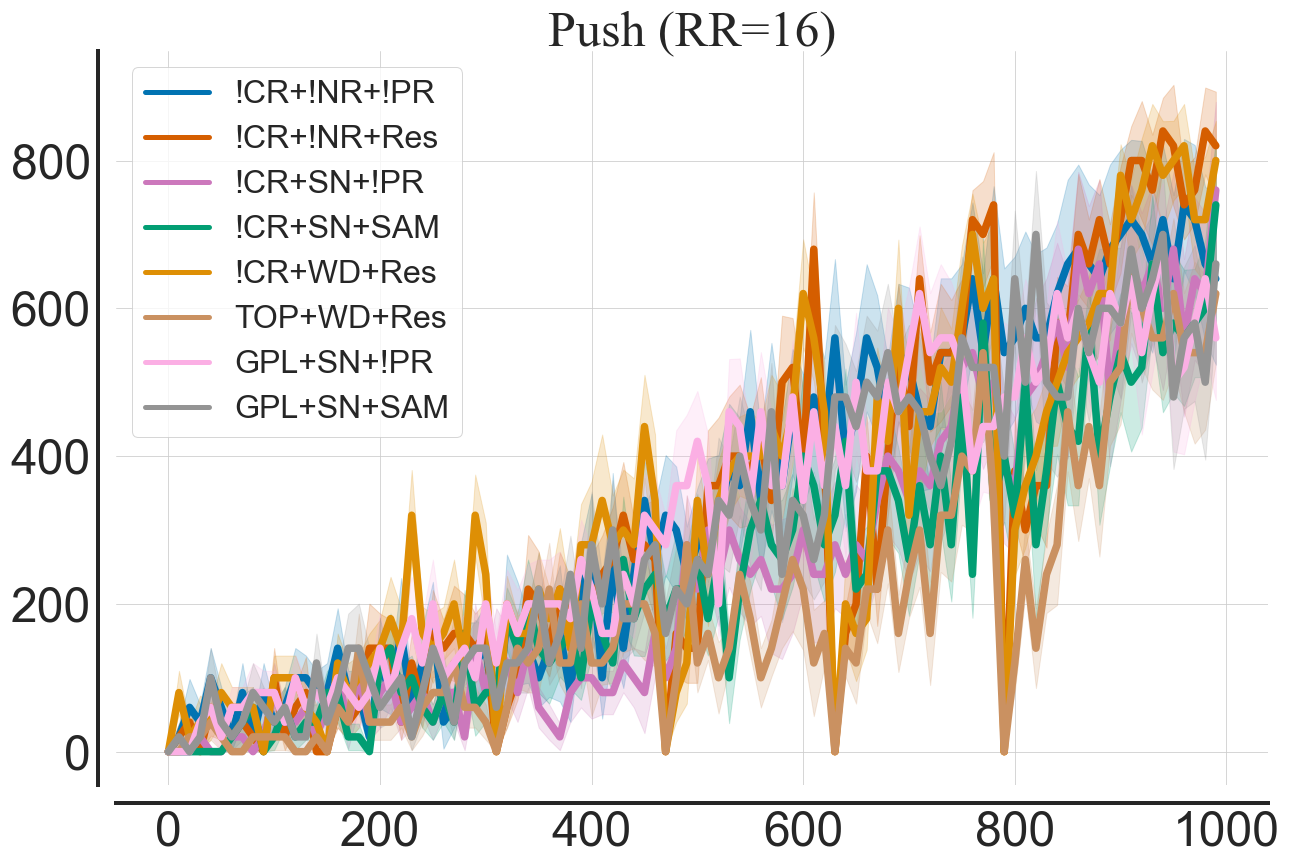}
    \hfill
    \includegraphics[width=0.19\linewidth]{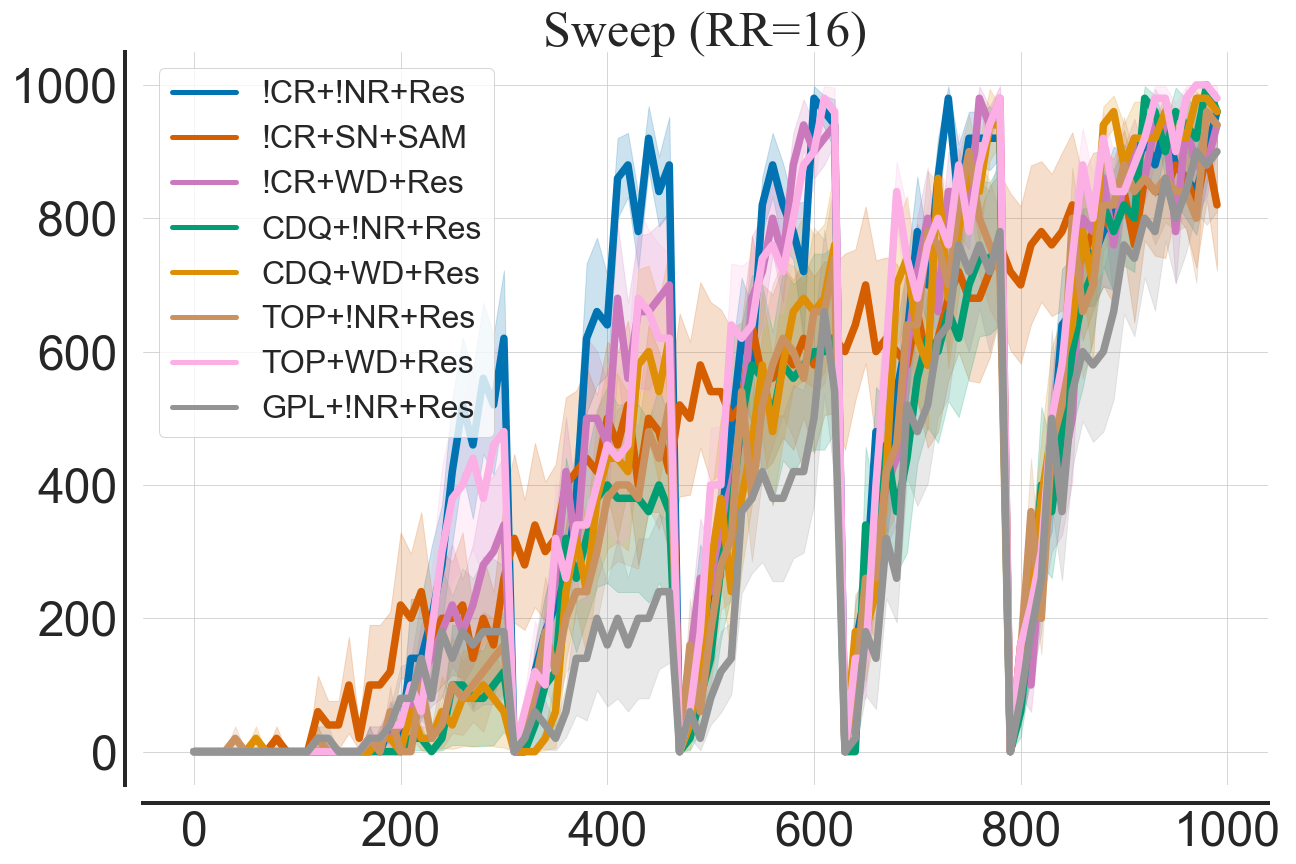}
    \end{subfigure}
\end{minipage}
\begin{minipage}[h]{1.0\linewidth}
    \begin{subfigure}{1.0\linewidth}
    \includegraphics[width=0.19\linewidth]{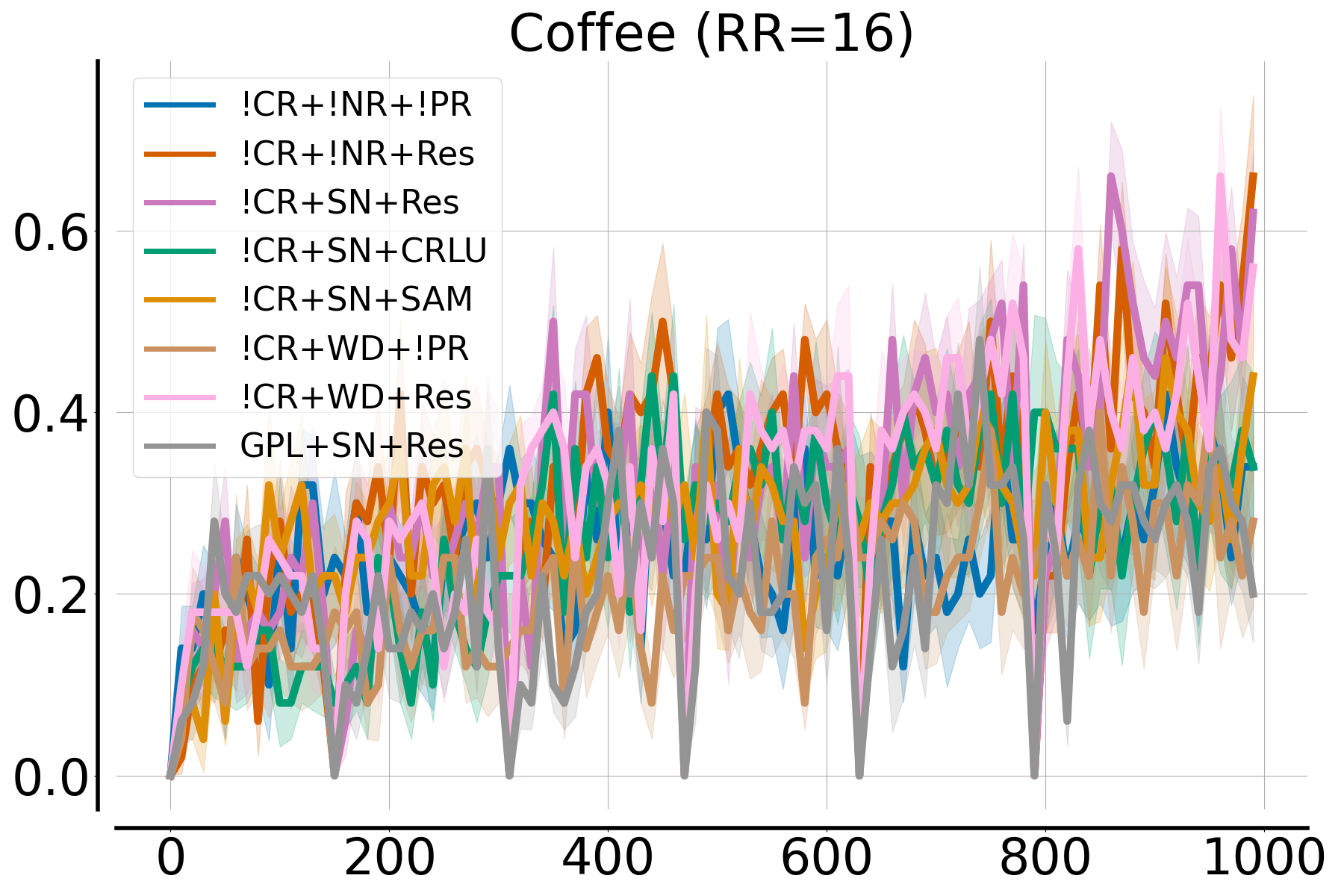}
    \hfill
    \includegraphics[width=0.19\linewidth]{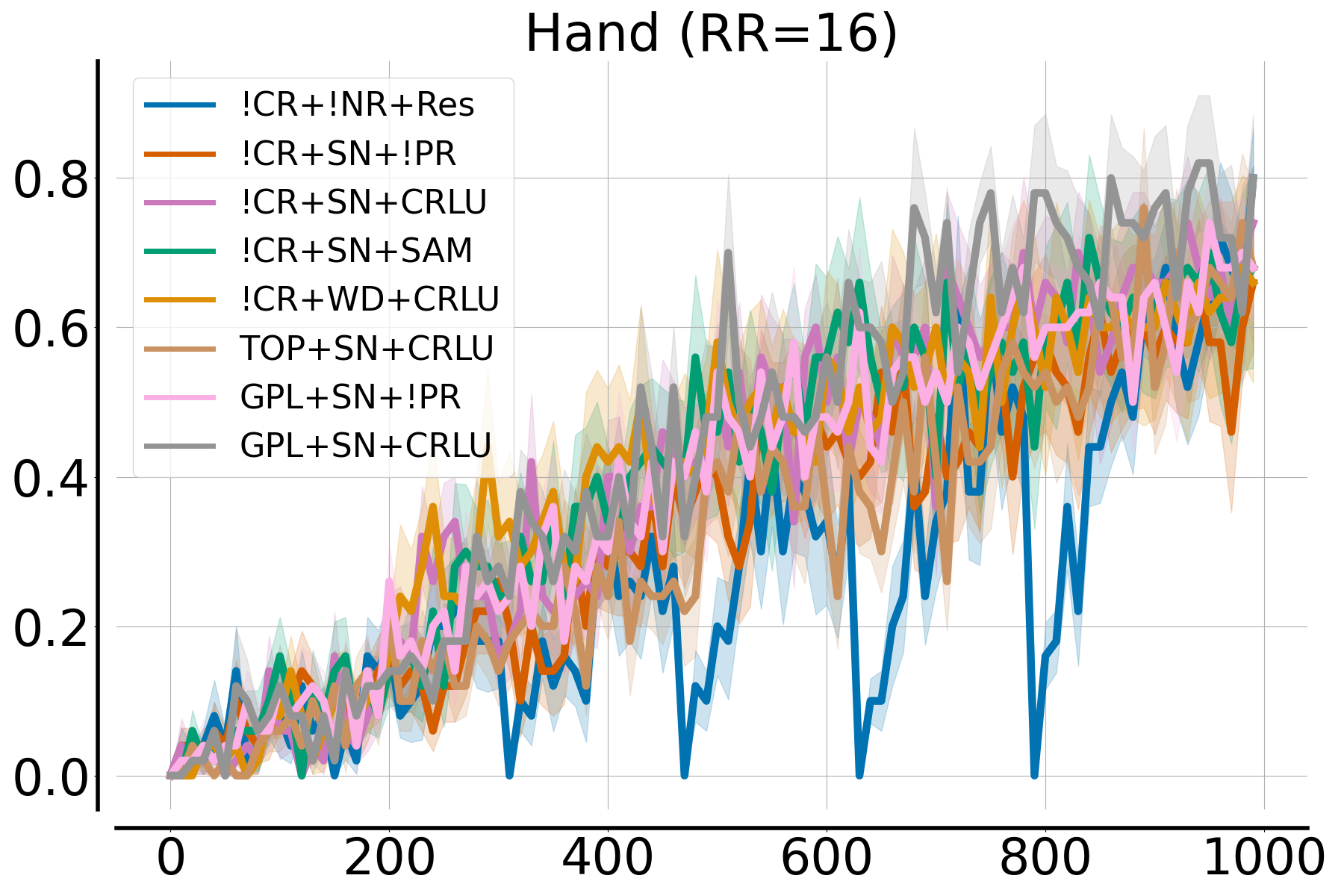}
    \hfill
    \includegraphics[width=0.19\linewidth]{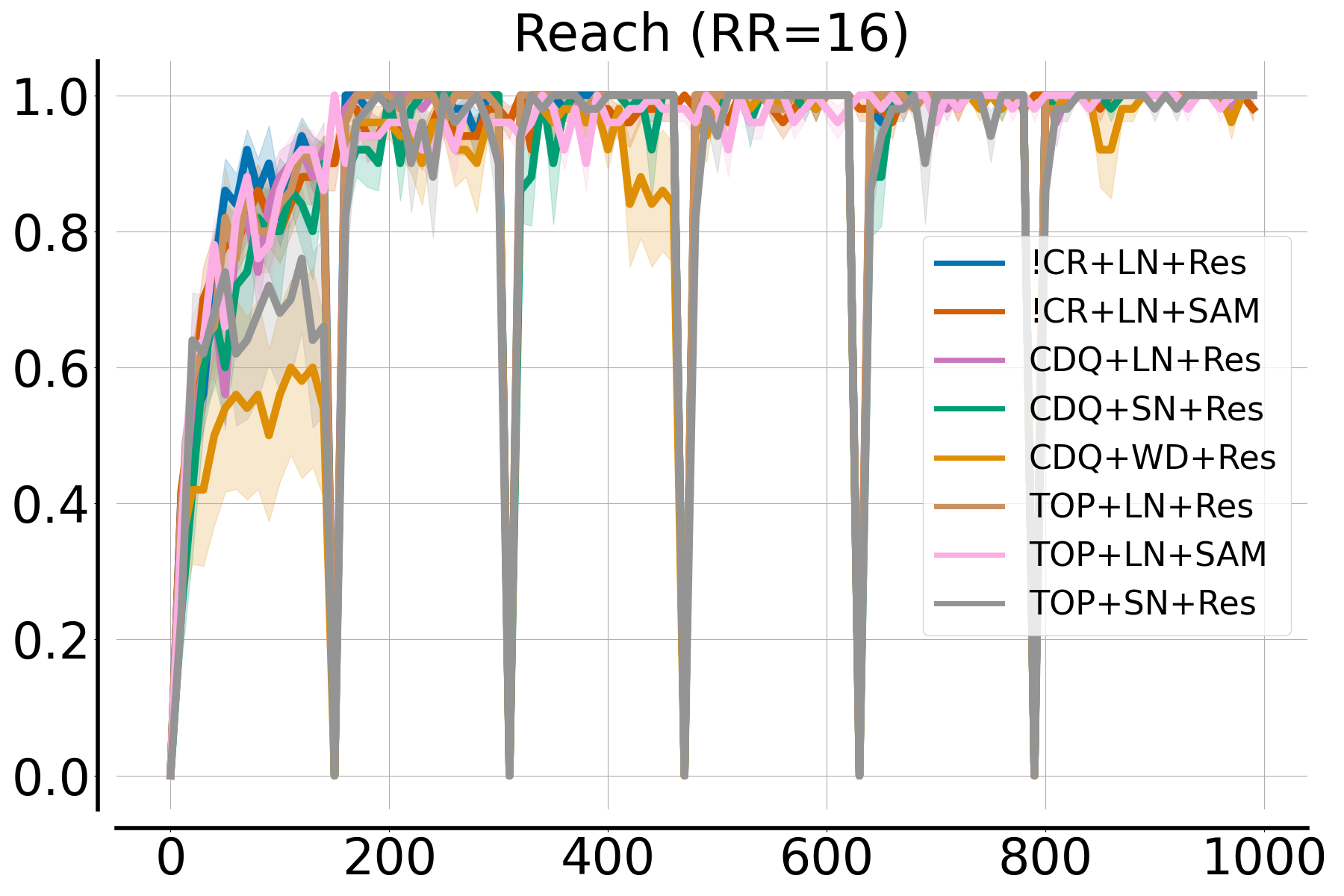}
    \hfill
    \includegraphics[width=0.19\linewidth]{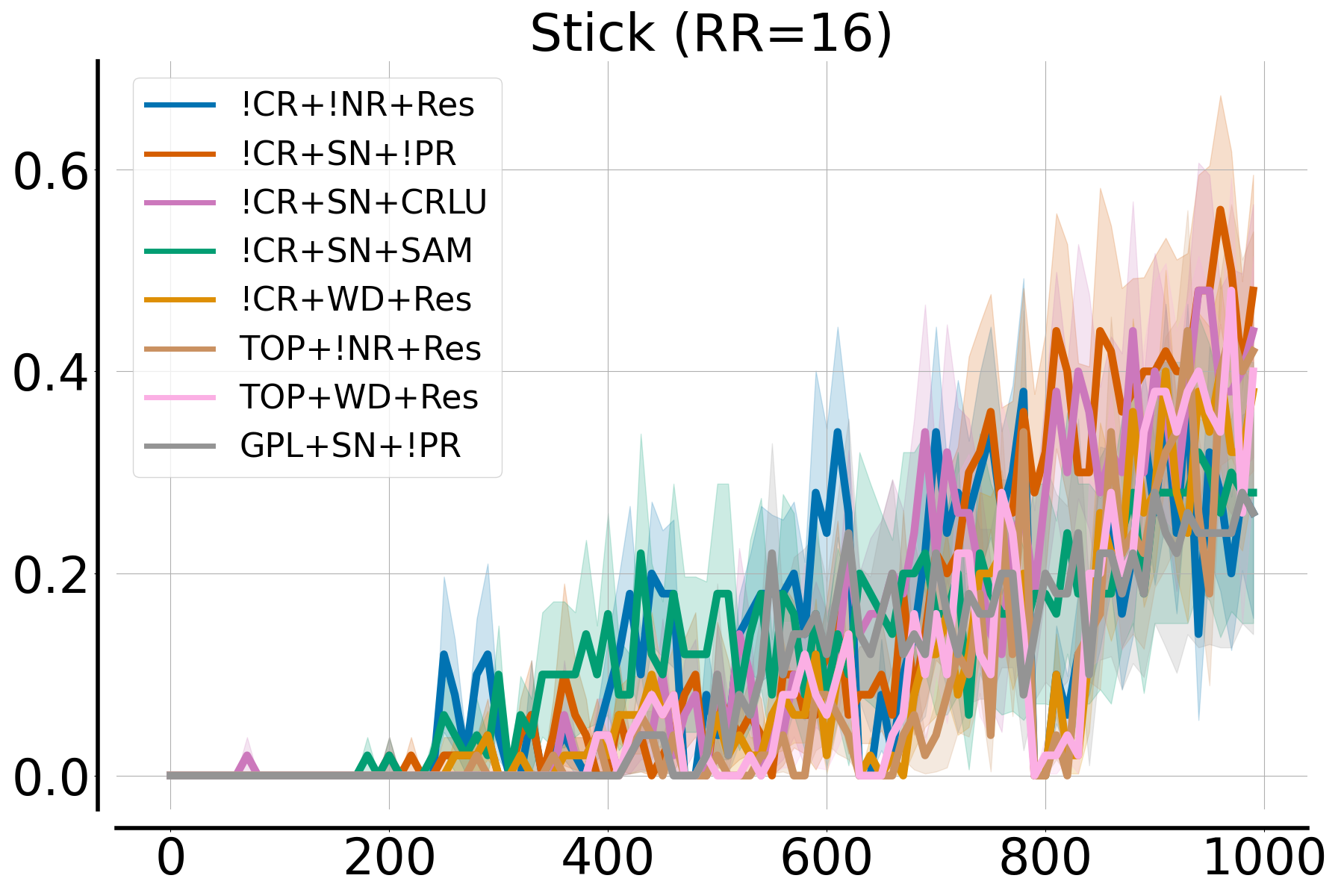}
    \end{subfigure}
\end{minipage}
\caption{Top performing configuration in the high replay regime. 10 seeds per task per algorithm.}
\label{fig:task_specific_rr16}
\end{center}
\end{figure*}

In a scenario with a high replay ratio, the resets are the most important intervention. They work best in combination with Layer norm, but other neural network regularization methods combined with resets are at the forefront regarding performance in the rr=16 scenario (see the top of Figure~\ref{fig:second_interv}). We present more results in Figures \ref{fig:task_specific_rr2} and \ref{fig:task_specific_rr16}.

\subsection{Gradient Norm Analysis}
\label{app:Grad_analysis}
\begin{figure}[ht!]
\centering
\begin{subfigure}{0.24\linewidth}
        \includegraphics[width=0.95\linewidth]{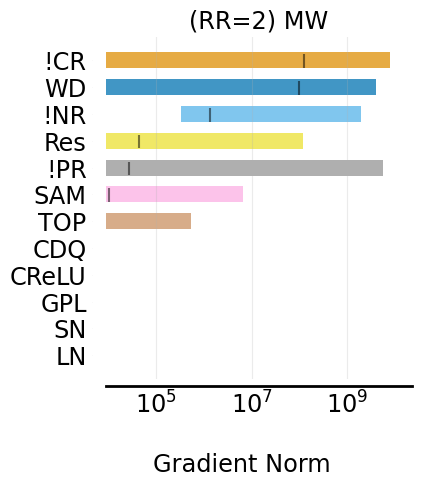}
\end{subfigure}
\begin{subfigure}{0.24\linewidth}
        \includegraphics[width=0.95\linewidth]{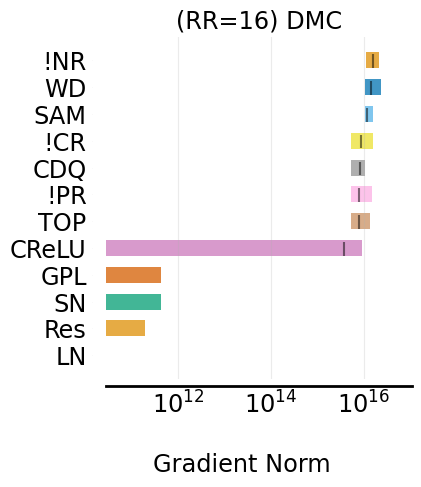}
\end{subfigure}
\begin{subfigure}{0.24\linewidth}
        \includegraphics[width=0.95\linewidth]{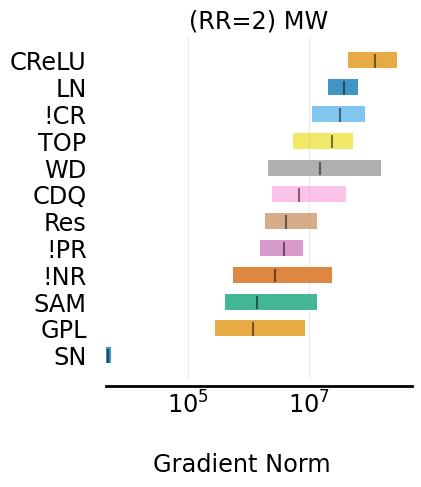}
\end{subfigure}
\begin{subfigure}{0.24\linewidth}
        \includegraphics[width=0.95\linewidth]{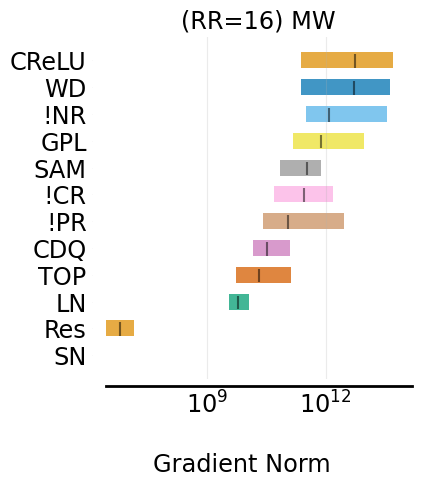}
\end{subfigure}
\caption{IQM gradient norm of first-order results. The IQM is calculated based on the average of the last ten evaluation points in each run, not the last evaluation point.}
\label{fig:iqm-GradNorms2}
\end{figure}

Drawing insights from the findings in section~\ref{sec:second_order}, where we highlighted the significance of spectral norm in enhancing agent performance on the MW benchmark, we now delve deeper into the behavior of the critic's gradient norm. In Figure~\ref{fig:iqm-GradNorms2}, one can compare orders of magnitude of IQM of gradient norm with respect to different replay ratio (RR) regimes and on different benchmarks. Referring to results from section~\ref{sec:second_order}, Figure~\ref{fig:iqm-GradNorms2} underscores that the gradient norm on MetaWorld, particularly in the RR=16 setup, exhibits orders of magnitude higher values compared to the DMC benchmark.

A similar phenomenon can be observed on the DMC benchmark, but the layer norm proves more robust in mitigating exploding gradients than the Spectral Norm. Interestingly, training on very complex environments such as Dog causes enormous gradient explosion, even in a small replay ratio regime~\ref{fig:iqm-GradNorms3}.

\begin{figure}[ht!]
\centering
\begin{subfigure}{0.24\linewidth}
        \includegraphics[width=0.95\linewidth]{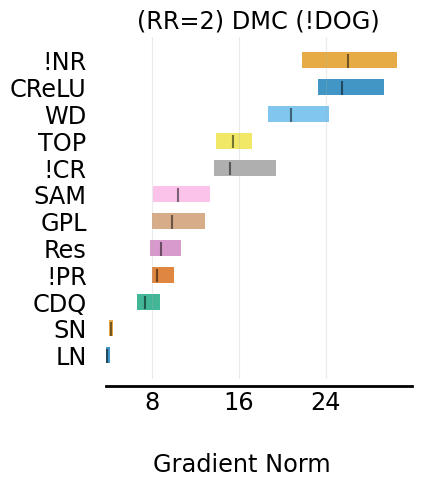}
\end{subfigure}
\begin{subfigure}{0.24\linewidth}
        \includegraphics[width=0.95\linewidth]{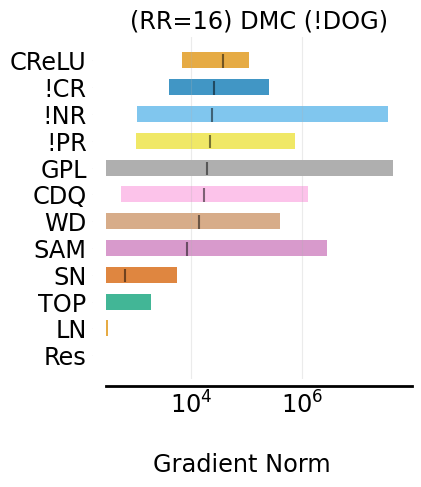}
\end{subfigure}
\begin{subfigure}{0.24\linewidth}
        \includegraphics[width=0.95\linewidth]{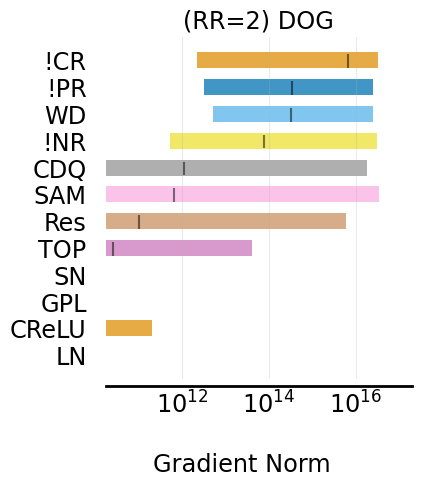}
\end{subfigure}
\begin{subfigure}{0.24\linewidth}
        \includegraphics[width=0.95\linewidth]{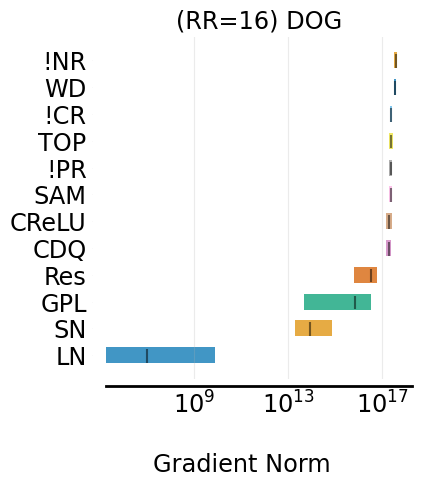}
\end{subfigure}
\caption{IQM gradient norm of first-order results. The IQM is calculated based on the average of the last ten evaluation points in each run, not the last evaluation point.}
\label{fig:iqm-GradNorms3}
\end{figure}

\subsection{Comparison of ReDO to other plasticity-inducing methods}
\label{sec:redo}
The ReDO method, as proposed by Sokar et al. \cite{sokar2023dormant}, is a technique for inducing network plasticity. It shares similarities with the full reset approach but involves more targeted interventions within the network. In particular, ReDo does not reset the full network; it only resets weights connected to dormant neurons. Specifically, incoming weights to dormant neurons are initialized as in full reset, but outgoing weights from dormant neurons are zeroed out, resulting in less severe network output changes. 

\begin{figure}[ht!]
\centering
\begin{subfigure}{0.33\linewidth}
        \includegraphics[width=0.95\linewidth]{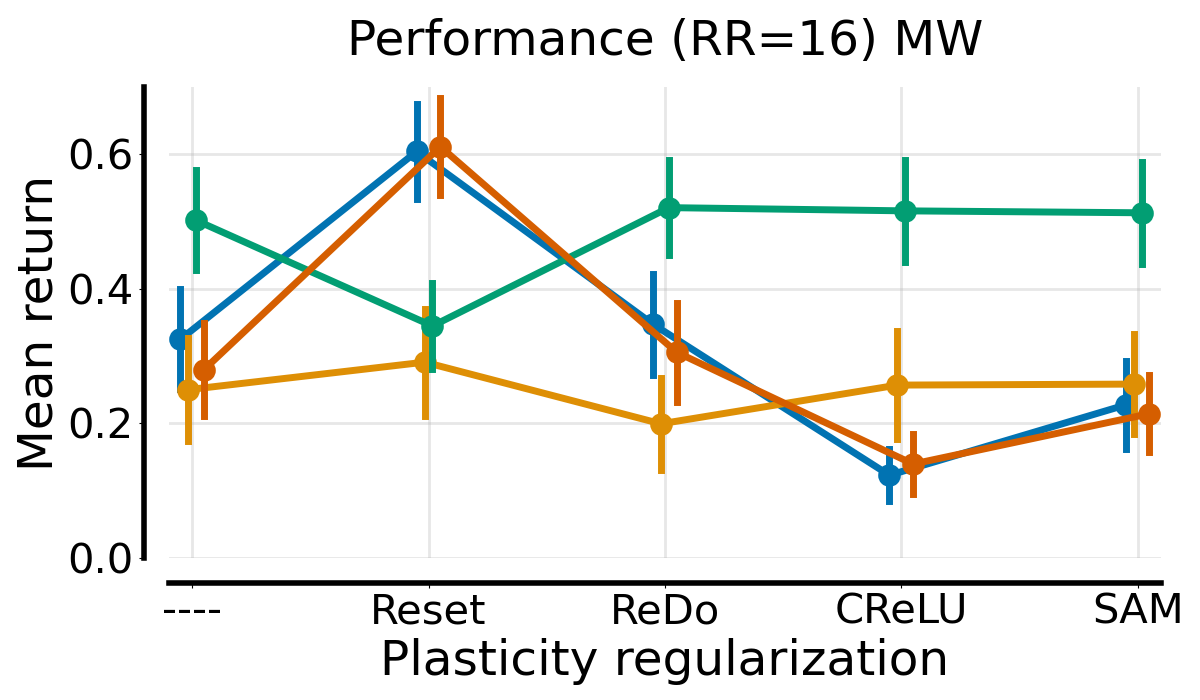}
\end{subfigure}
\begin{subfigure}{0.33\linewidth}
        \includegraphics[width=0.95\linewidth]{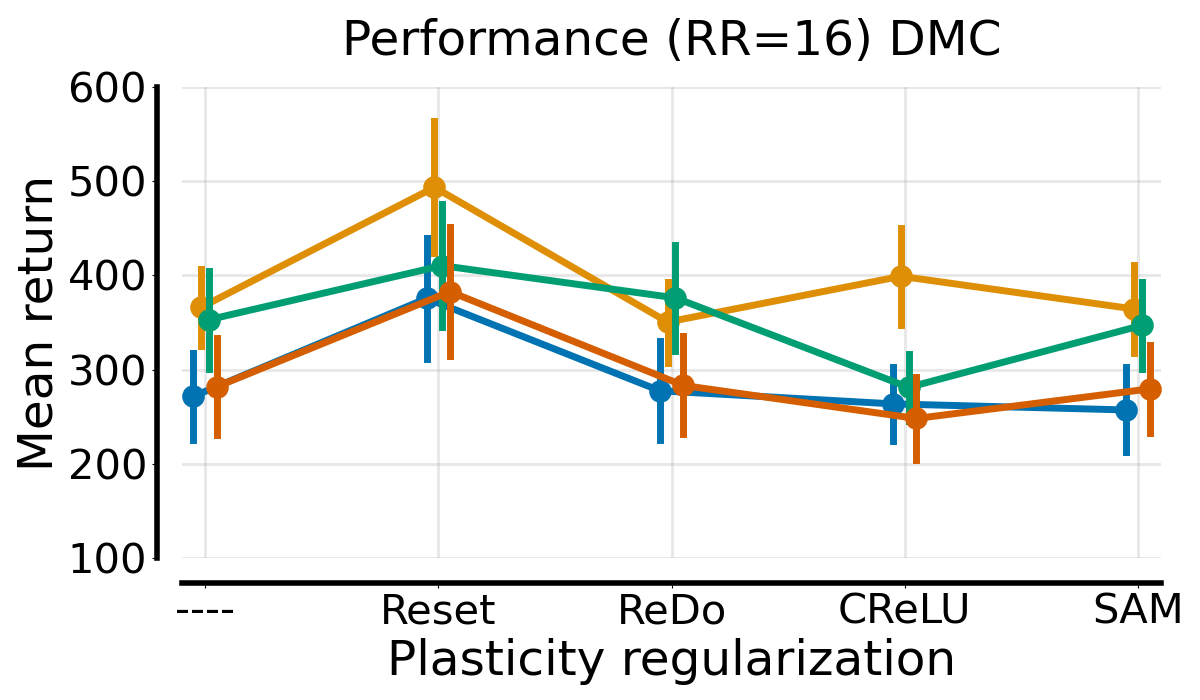}
\end{subfigure}
\begin{subfigure}{0.33\linewidth}
        \includegraphics[width=0.95\linewidth]{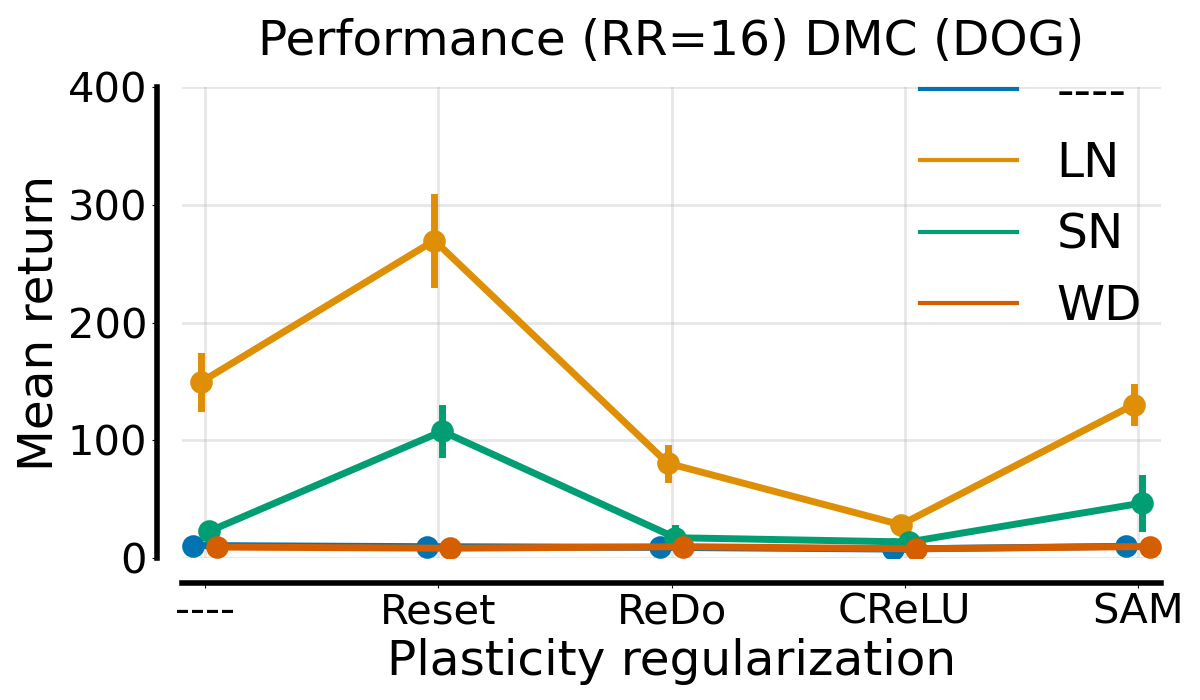}
\end{subfigure}
\caption{Second-order results marginalizing critic regularization methods with ReDO.}
\label{fig:redo}
\end{figure}

\begin{figure}[ht!]
\centering
\begin{subfigure}{0.99\linewidth}
        \includegraphics[width=0.24\linewidth]{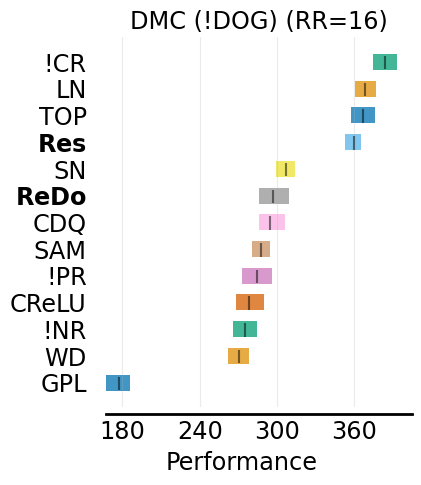}
        \includegraphics[width=0.24\linewidth]{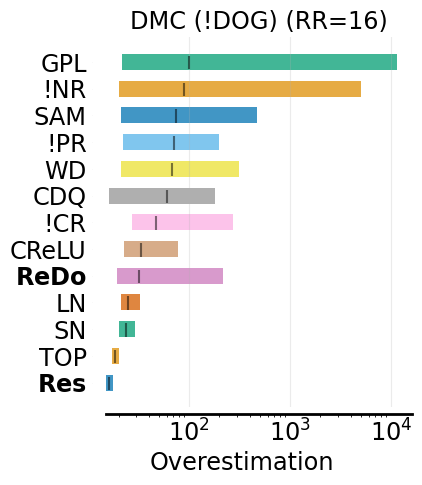}
        \includegraphics[width=0.24\linewidth]{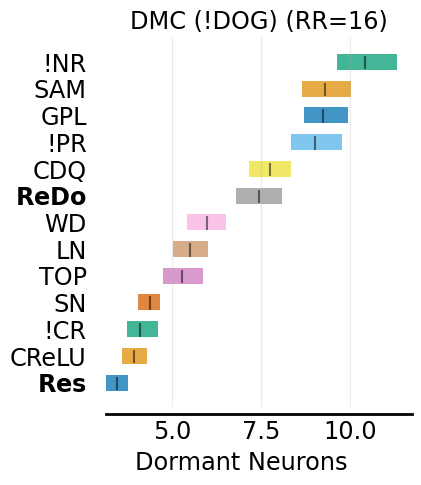}
        \includegraphics[width=0.24\linewidth]{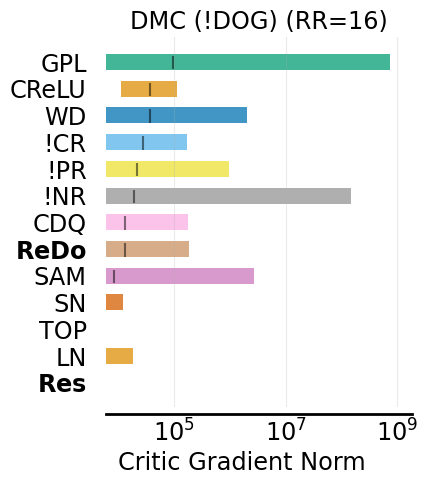}
\end{subfigure}
\begin{subfigure}{0.99\linewidth}
        \includegraphics[width=0.24\linewidth]{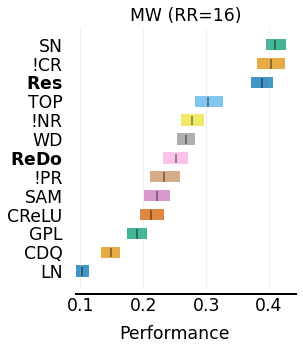}
        \includegraphics[width=0.24\linewidth]{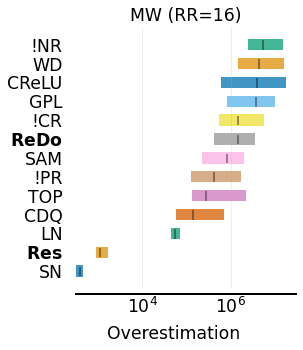}
        \includegraphics[width=0.24\linewidth]{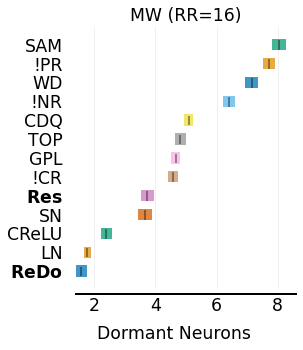}
        \includegraphics[width=0.24\linewidth]{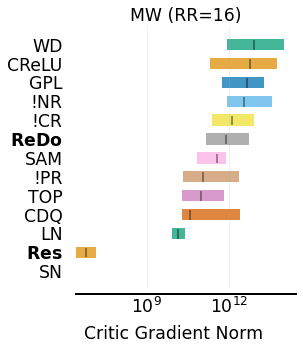}
\end{subfigure}
\begin{subfigure}{0.99\linewidth}
        \includegraphics[width=0.24\linewidth]{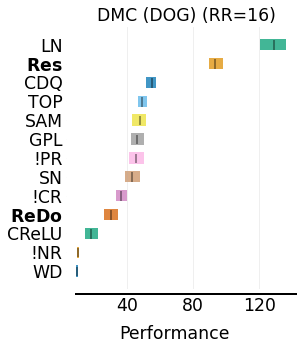}
        \includegraphics[width=0.24\linewidth]{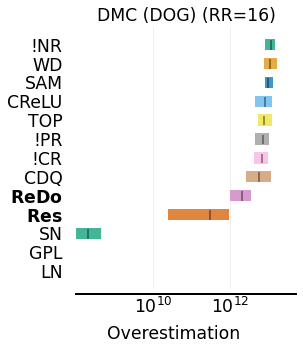}
        \includegraphics[width=0.24\linewidth]{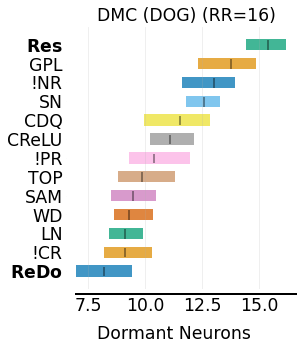}
        \includegraphics[width=0.24\linewidth]{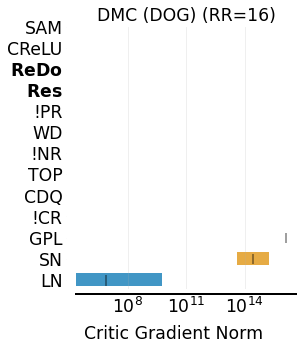}
\end{subfigure}
\caption{IQM of performance, overestimation, dormant neurons, and gradient norm, of first-order results. The IQM is calculated based on the average of the last ten evaluation points in each run, not the last evaluation point.}
\label{fig:redo_agg}
\end{figure}

Figure \ref{fig:redo} presents results from Figure\ref{fig:marginalized} updated with the ReDo method for both MetaWorld and DMC environments for a high replay ratio regime. ReDo does not perform as well as resets in the Metaworld and DMC suites. However, both methods effectively reduce the critic gradient norm, overestimation and the number of dormant neurons for both Metaworld and DMC without dog benchmarks, as shown in Figure \ref{fig:redo_agg}. In dog environments, we observed that ReDo was unstable for runs without SN or LN, and some runs crashed. We report results for the last ten timesteps before the crash for these runs. Interestingly, all methods except SN and LN cannot reduce the critic gradient norm, as shown in the bottom right plot in Figure \ref{fig:redo_agg}.


\subsection{Closer look on CDQL performance on Hopper and Quadruped environments}
\label{sec:CDQL_negative}
\begin{figure*}[ht!]
\begin{center}
\begin{minipage}[h]{1.0\linewidth}
    \begin{subfigure}{1.0\linewidth}
     \includegraphics[width=0.24\linewidth]{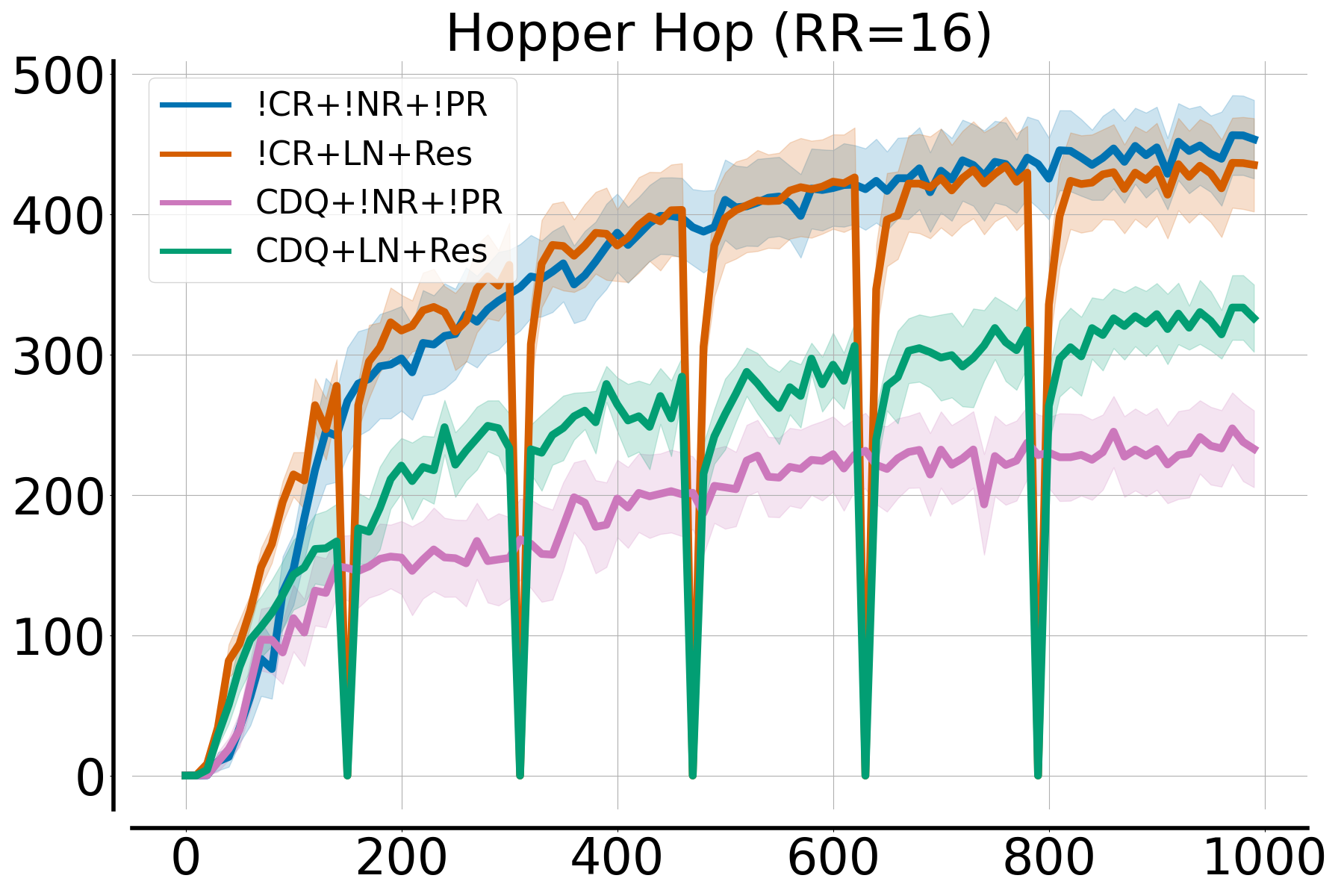}
    \hfill
    \includegraphics[width=0.24\linewidth]{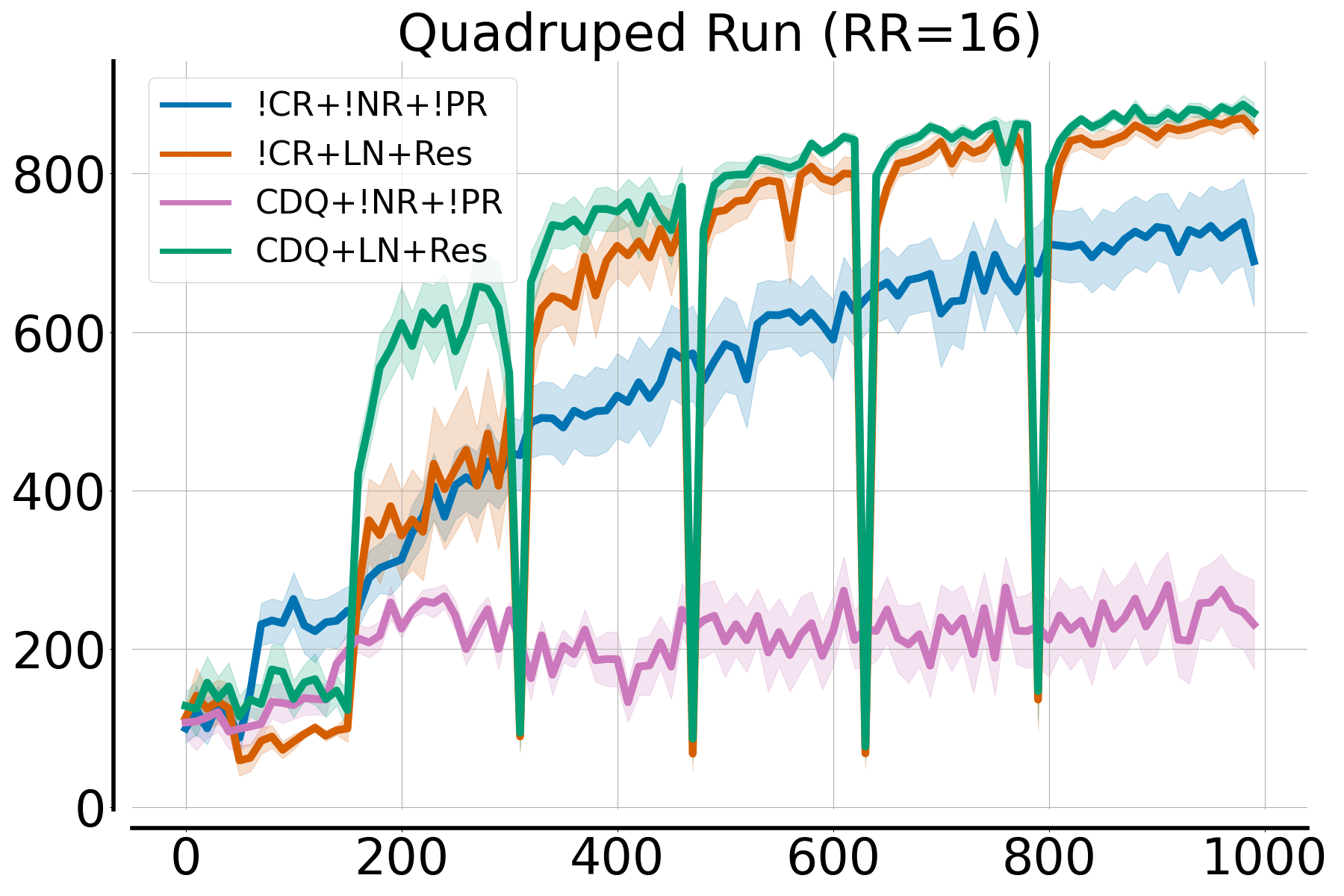}
      \hfill
    \includegraphics[width=0.24\linewidth]{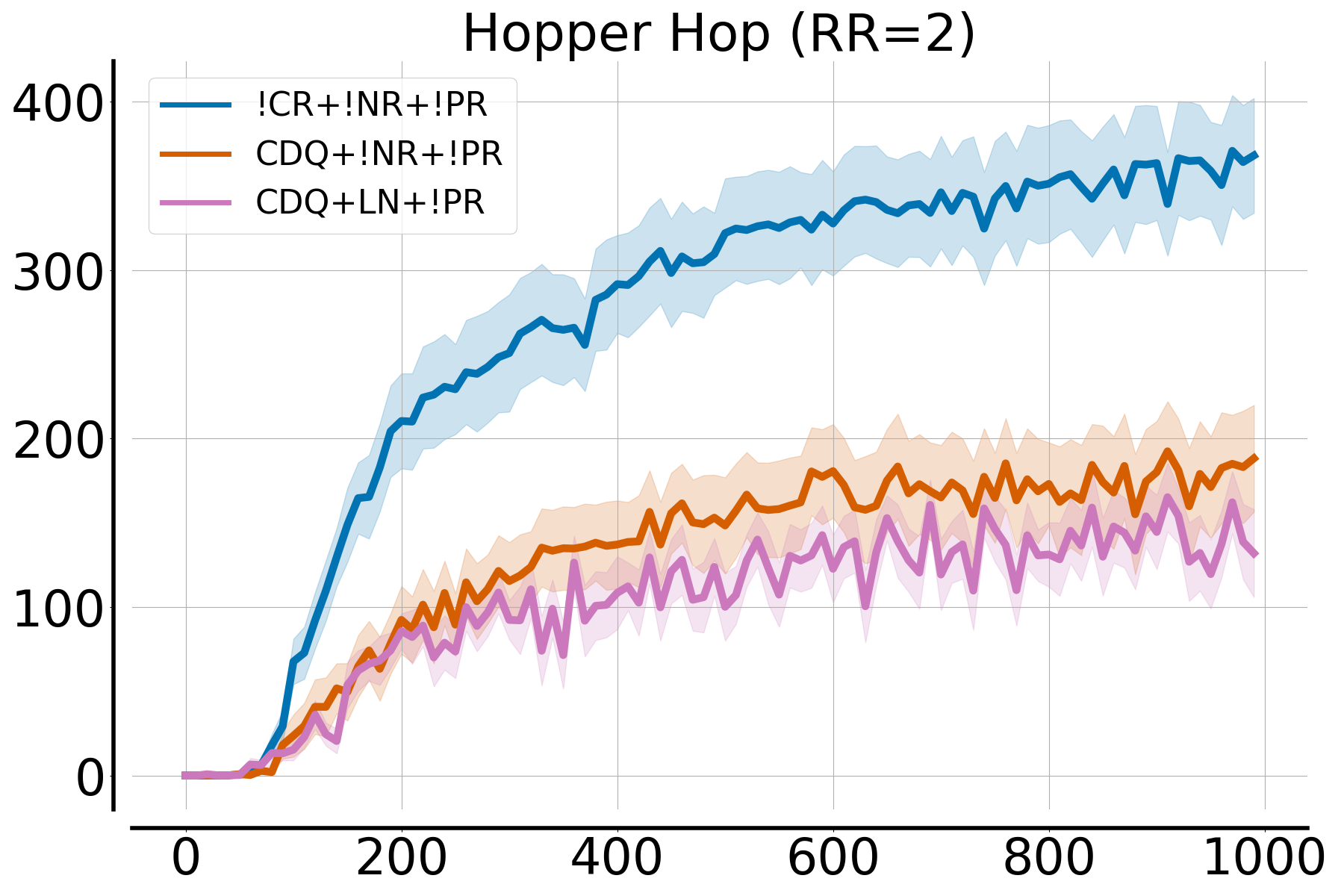}
    \hfill
    \includegraphics[width=0.24\linewidth]{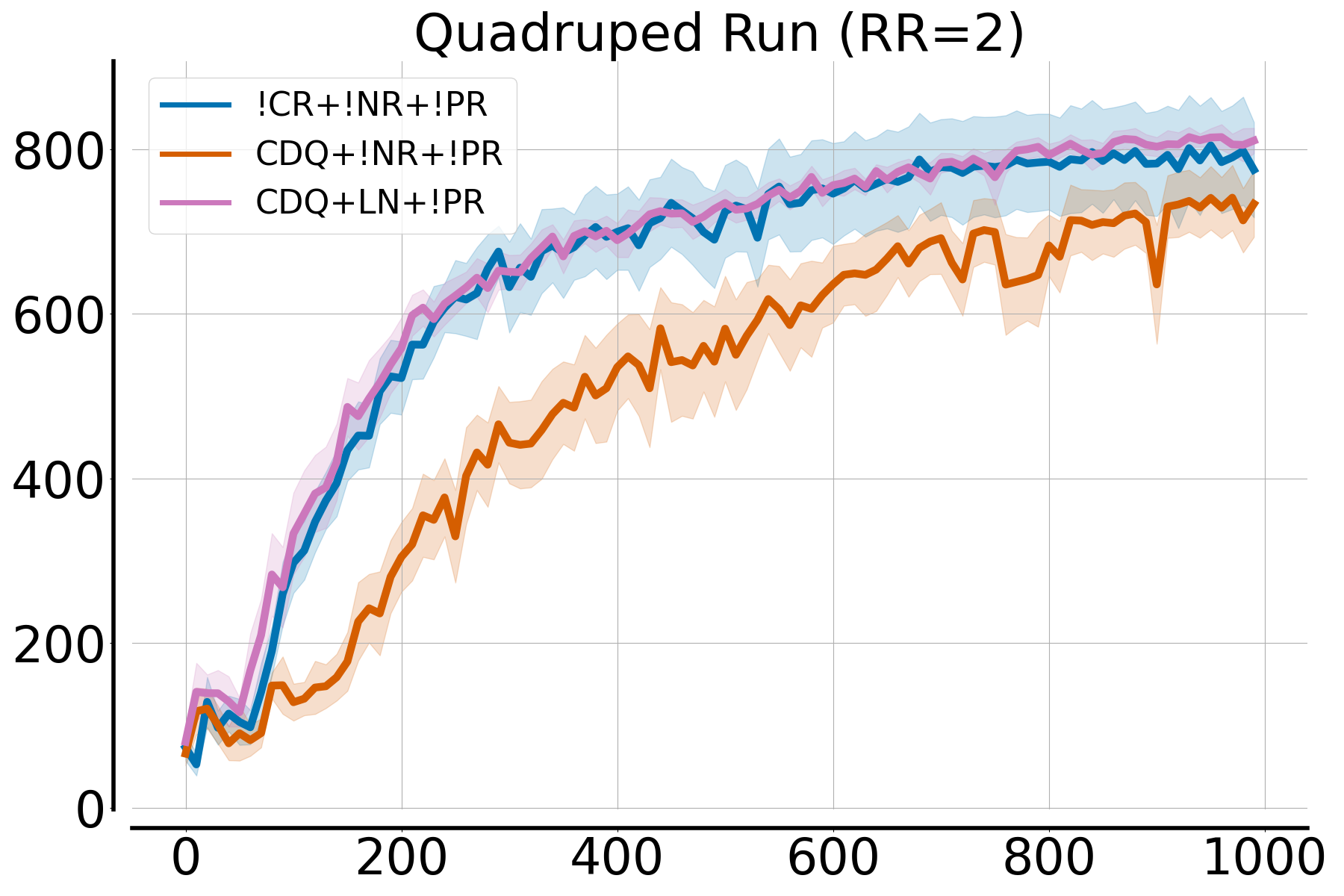}
    \end{subfigure}
\end{minipage}
\begin{minipage}[h]{1.0\linewidth}
    \begin{subfigure}{1.0\linewidth}
     \includegraphics[width=0.24\linewidth]{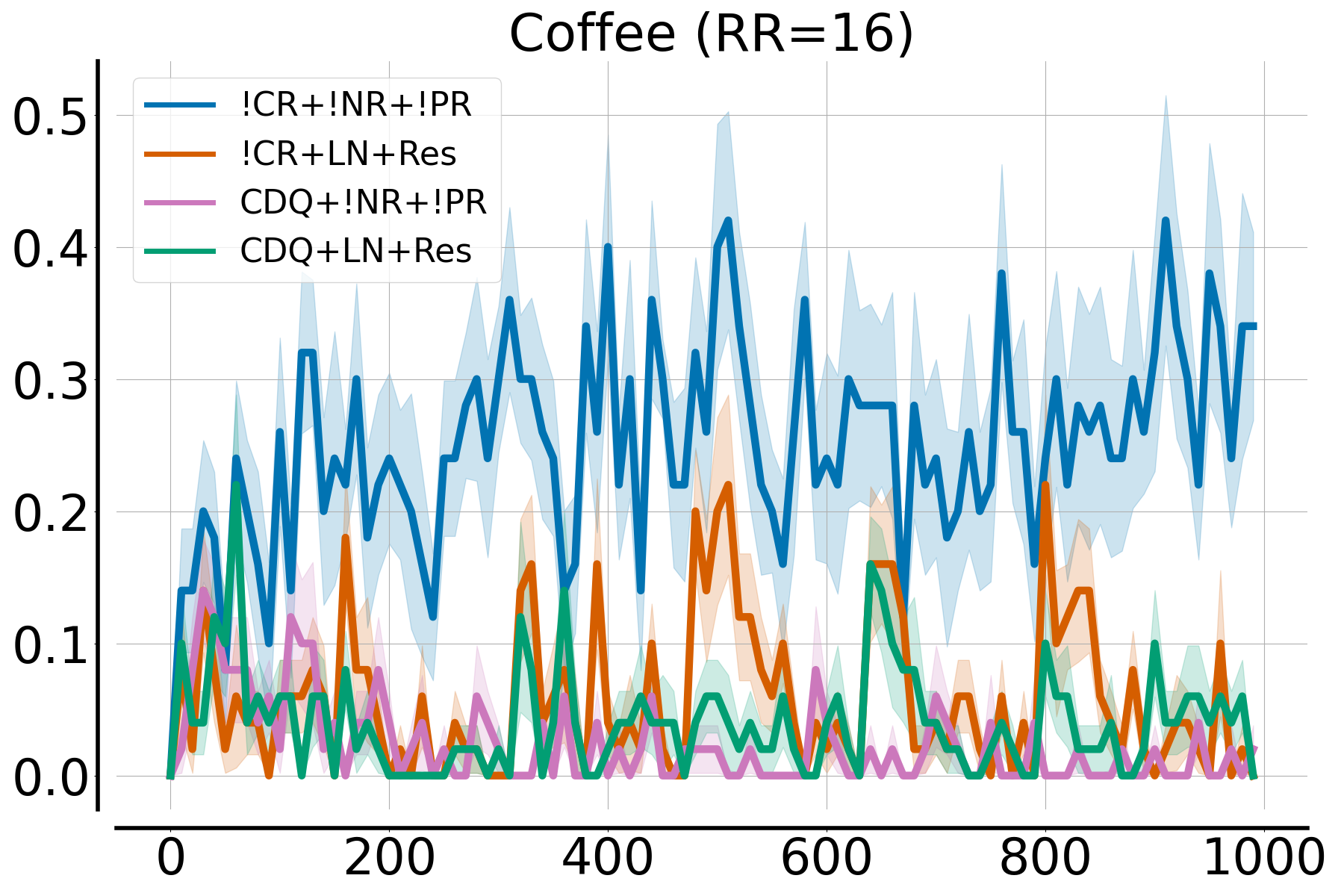}
    \hfill
    \includegraphics[width=0.24\linewidth]{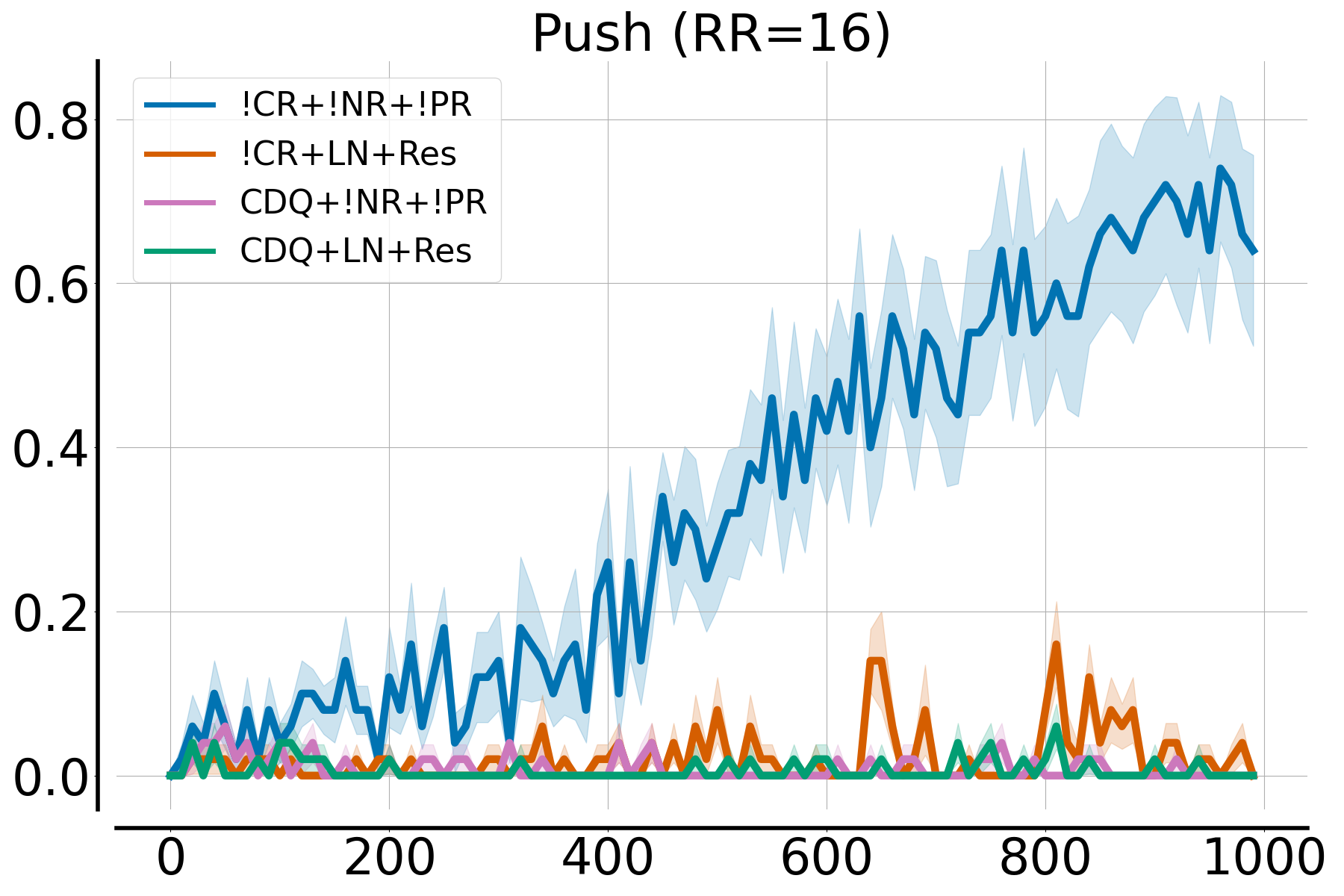}
      \hfill
    \includegraphics[width=0.24\linewidth]{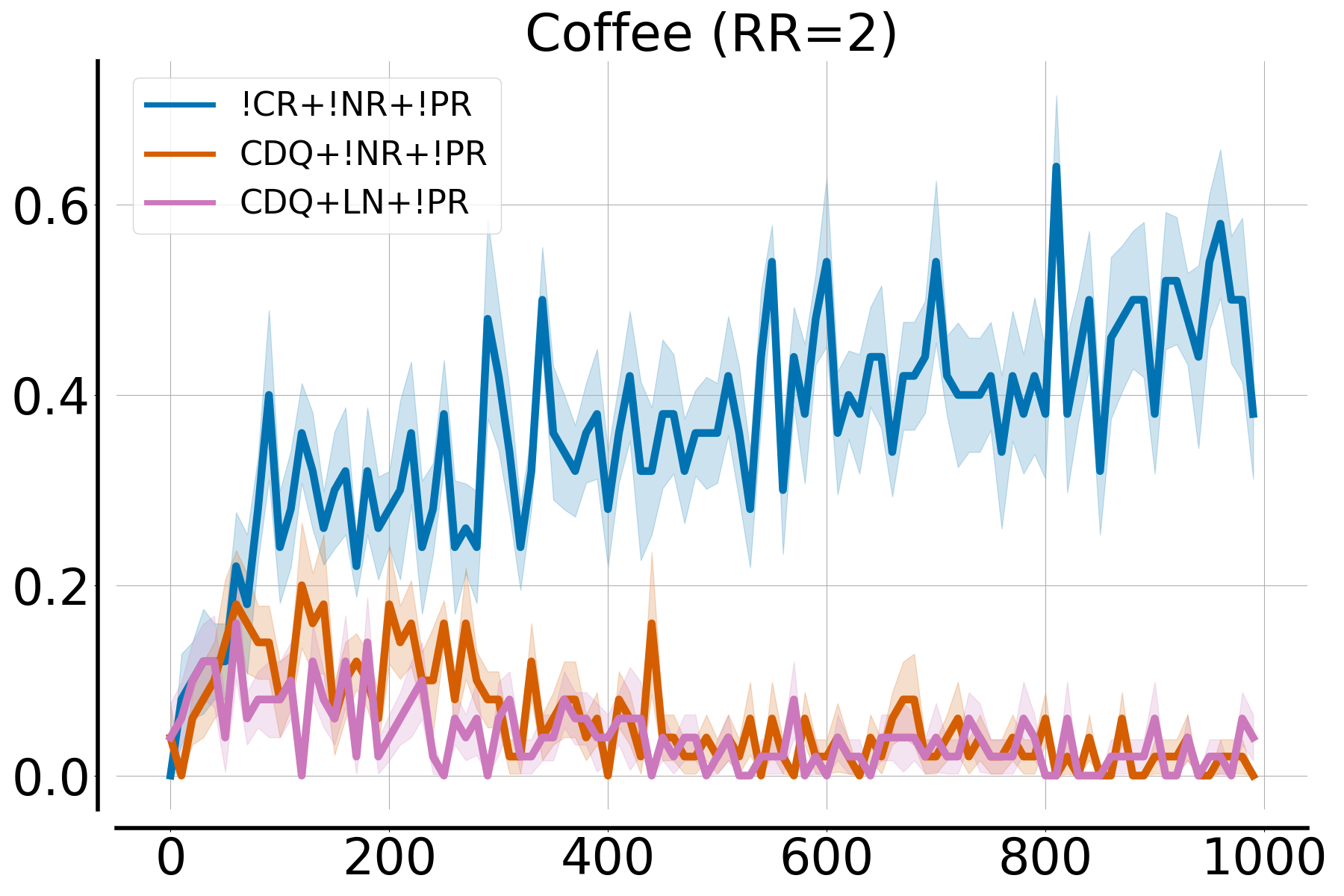}
    \hfill
    \includegraphics[width=0.24\linewidth]{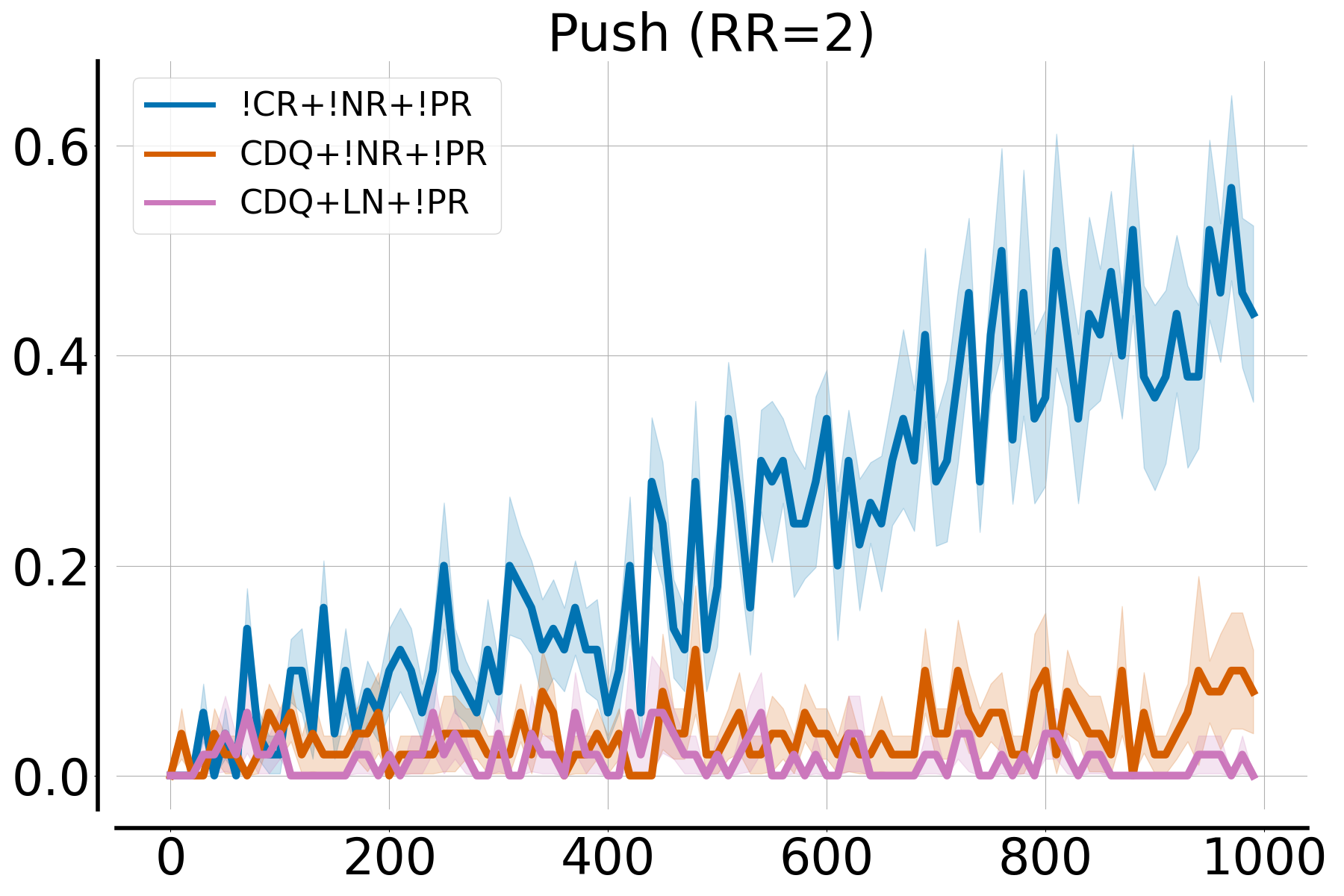}
    \end{subfigure}
\end{minipage}
\begin{minipage}[h]{1.0\linewidth}
    \begin{subfigure}{1.0\linewidth}
    \hfill
     \includegraphics[width=0.24\linewidth]{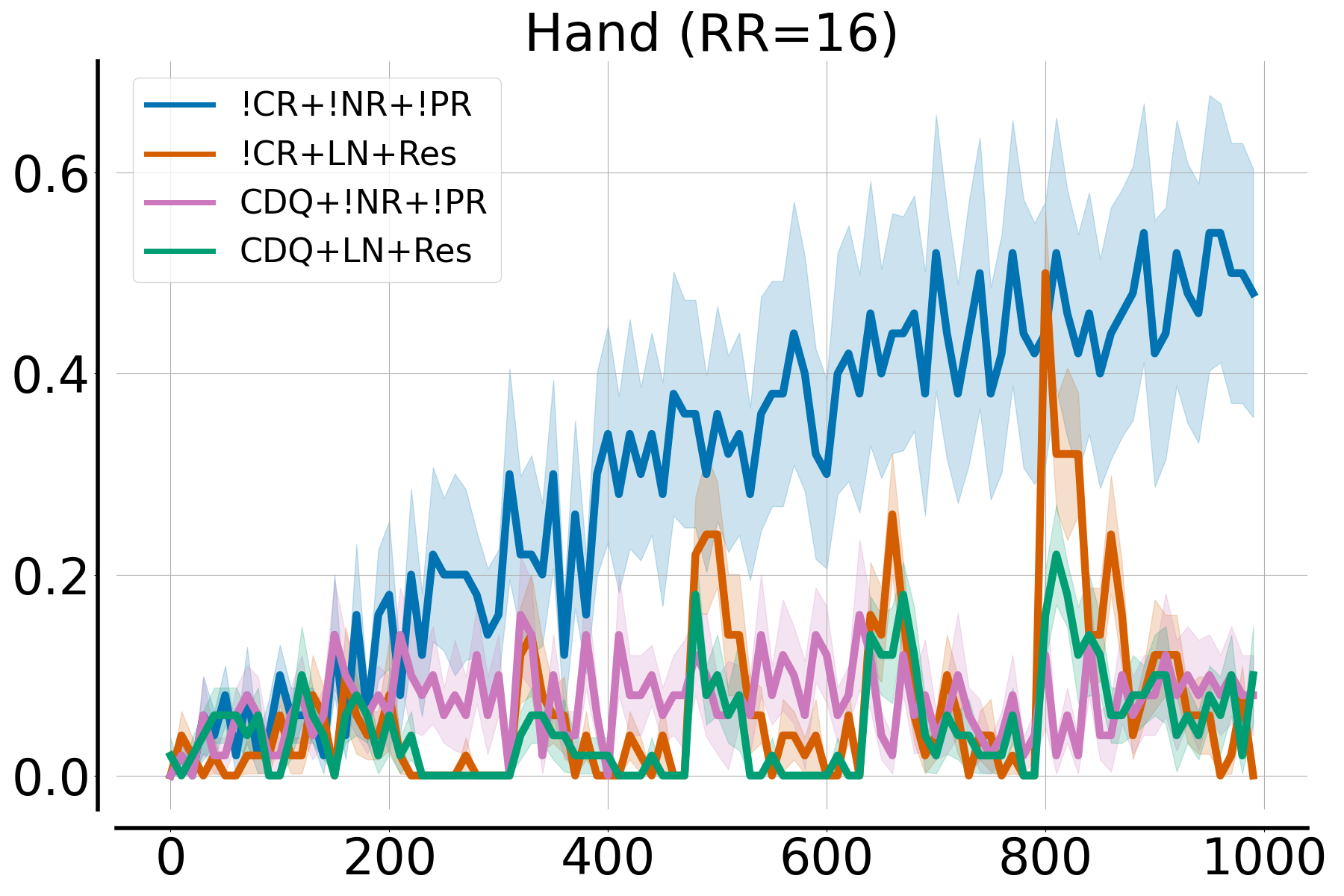}
    \includegraphics[width=0.24\linewidth]{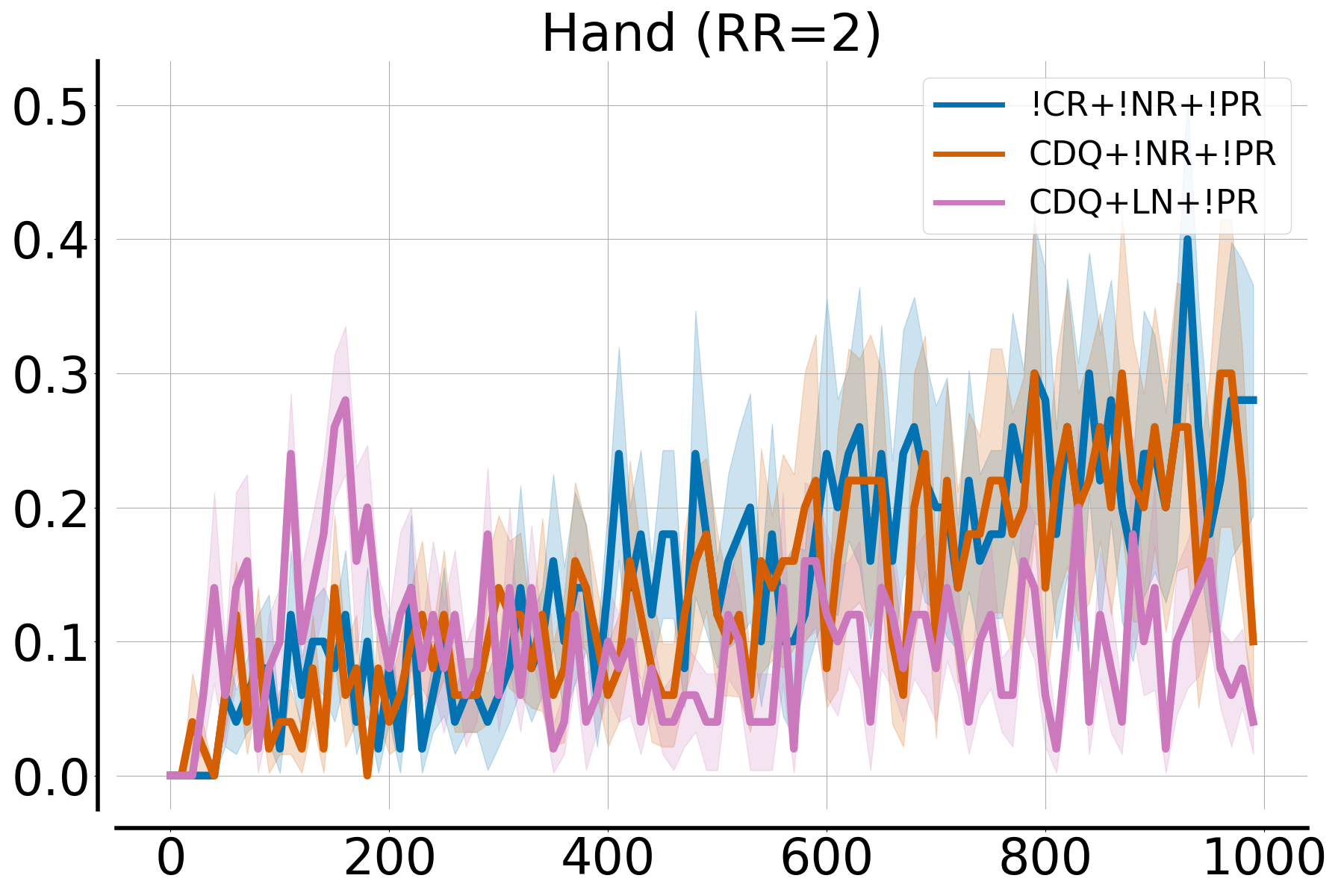}
\hfill
    \end{subfigure}
\end{minipage}
\caption{Examining the influence of CDQL on performance in the Hopper Hop and Quadruped Run environments within the DMC benchmark (top row), as well as the Push, Coffee, and Hand environments within the MW benchmark (bottom row).}
\label{fig:closer_cdql_vs_cr}
\end{center}
\end{figure*}

Figure~\ref{fig:closer_cdql_vs_cr} illustrates the outcomes of applying Clipped Double Q-learning (CDQL)~\citep{haarnoja2018soft,fujimoto2018addressing,hansen2022temporal}, a widely used and straightforward critic regularization method, across various environments. In the case of the hopper hop environment or some MW environments (bottom row), CDQL exhibits a detrimental effect on performance, with additional regularization techniques such as resets or layer normalization failing to alleviate this effect. Similarly, in the quadruped run task, CDQL demonstrates a comparable negative impact, although the application of additional regularization methods successfully mitigates it. 

\subsection{Regression and Spearman correlation analysis}
\label{app:Scatter_regression}

Deepening analysis from section \ref{sec:correlations}, we investigate the coefficients of Ordinary Least Square regression for overestimation, overfitting, dormant neurons and critic's gradient norm with return. Except for dormant neurons, we take the symmetrical logarithm of every metric. We perform regression fitting for every variable separately, which results in four different regression analyses. As previously stated, because these relations are highly non-linear, however, often monotonic, goodness of fit coefficients such as $R^2$ for linear regression are low. Besides regression coefficients ($\beta$), we report a standard error of regression ($se$), $R^2$, p-value $p$ in the plot legend. In the axis titles, the Spearman correlation and its p-value are shown. 

\begin{figure*}[ht!]
\begin{center}
    \begin{minipage}[h]{1.0\linewidth}
        \begin{subfigure}{1.0\linewidth}
            \includegraphics[width=0.99\linewidth]{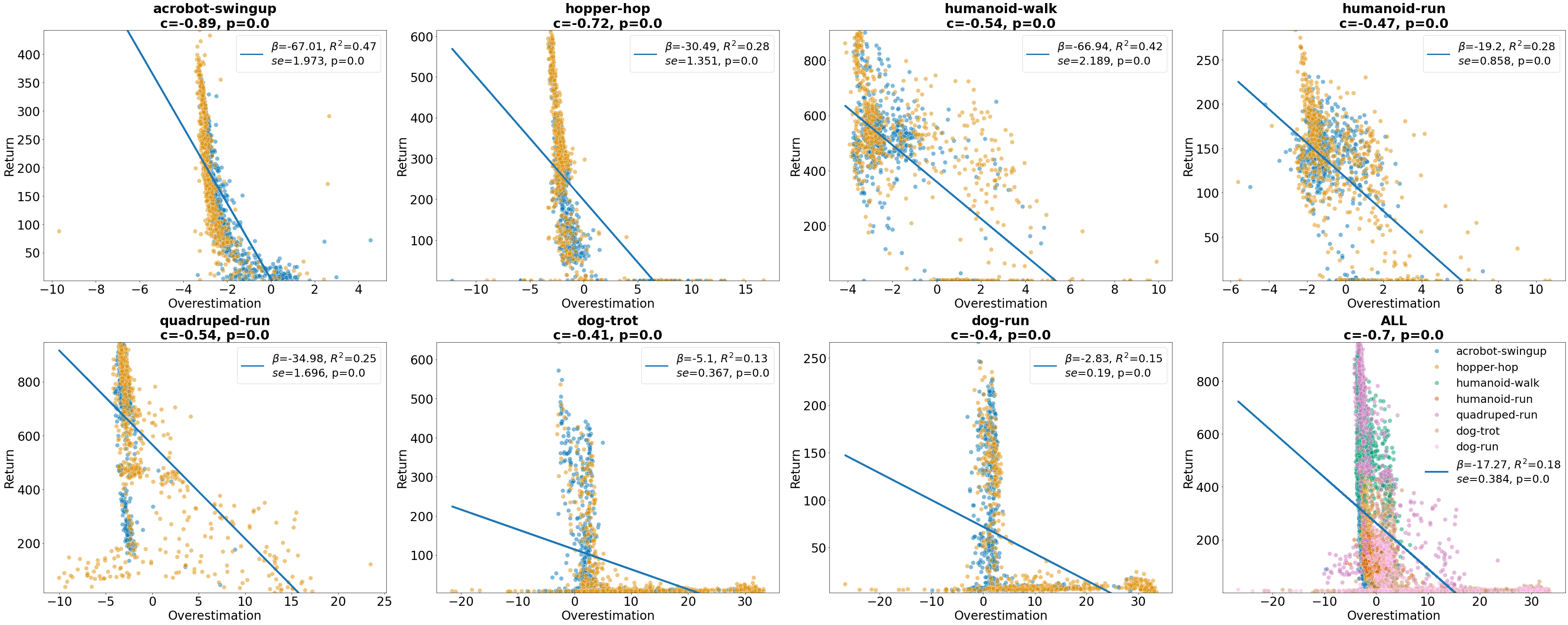}
        \end{subfigure}
    \end{minipage}
\caption{Overestimation logarithm scatter plots with regression line for DMC environments.}
\label{fig:dmc_reg_overest}
\end{center}
\end{figure*}

\begin{figure*}[ht!]
\begin{center}
    \begin{minipage}[h]{1.0\linewidth}
        \begin{subfigure}{1.0\linewidth}
            \includegraphics[width=0.99\linewidth]{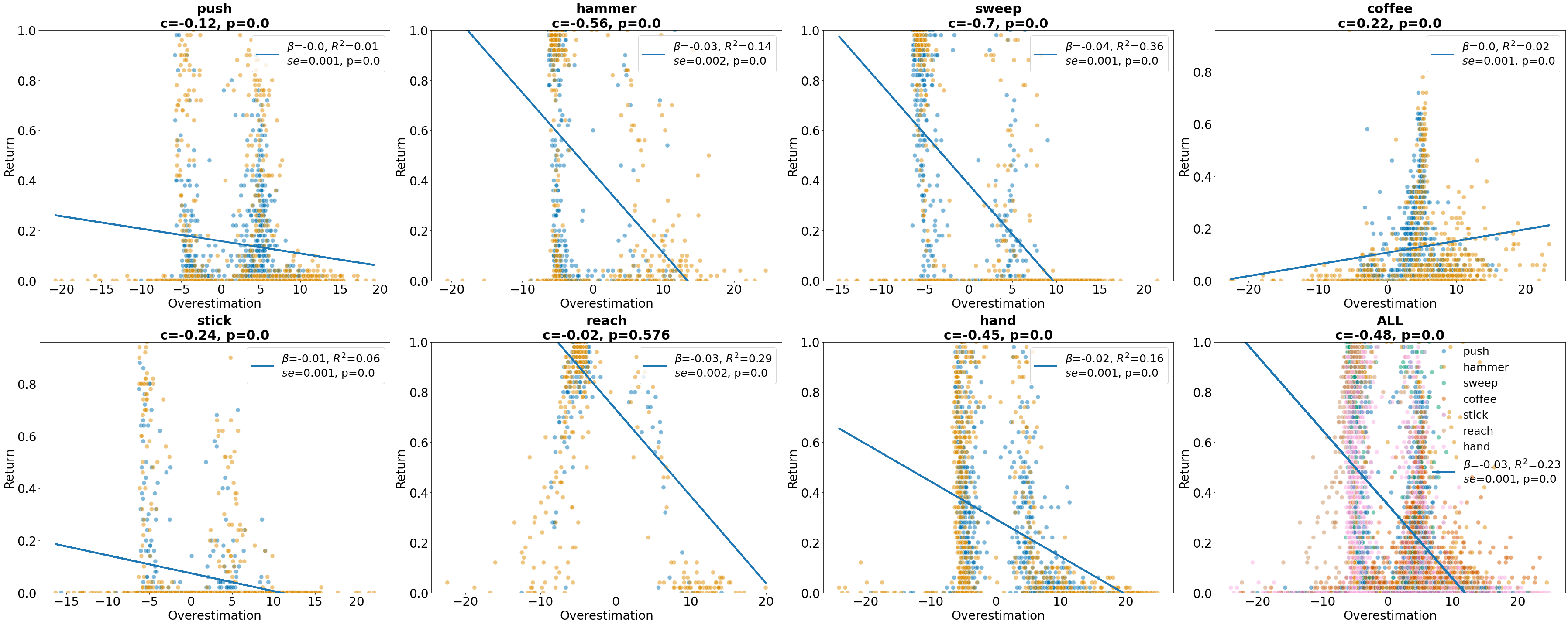}
        \end{subfigure}
    \end{minipage}
\caption{Overestimation logarithm scatter plots with regression line for MW environments.}
\label{fig:mw_reg_overest}
\end{center}
\end{figure*}

In figures \ref{fig:dmc_reg_overest} and \ref{fig:mw_reg_overest}, we observe consistent negative coefficients, except for the Coffee environment for overestimation and return. Interestingly, for the MetaWorld benchmark, most environments exhibit a symmetrical degree of overestimation and underestimation as two distinct clusters of well-performing runs are mirrored by the Y axis. For the DMC environments, we observe a much higher amount of runs in which overestimation is present.

\begin{figure*}[ht!]
\begin{center}
    \begin{minipage}[h]{1.0\linewidth}
        \begin{subfigure}{1.0\linewidth}
            \includegraphics[width=0.99\linewidth]{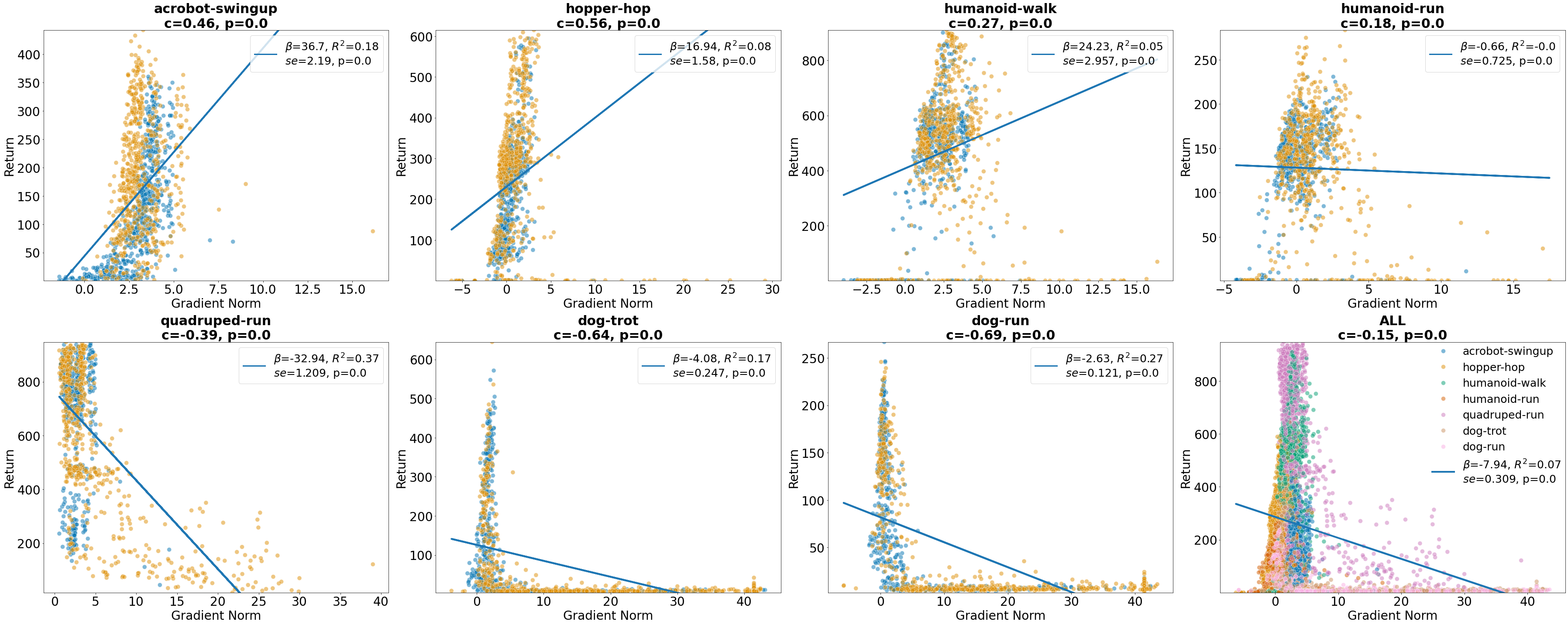}
        \end{subfigure}
    \end{minipage}
\caption{Gradient norm logarithm scatter plots with regression line for DMC environments.}
\label{fig:dmc_reg_gn}
\end{center}
\end{figure*}

\begin{figure*}[ht!]
\begin{center}
    \begin{minipage}[h]{1.0\linewidth}
        \begin{subfigure}{1.0\linewidth}
            \includegraphics[width=0.99\linewidth]{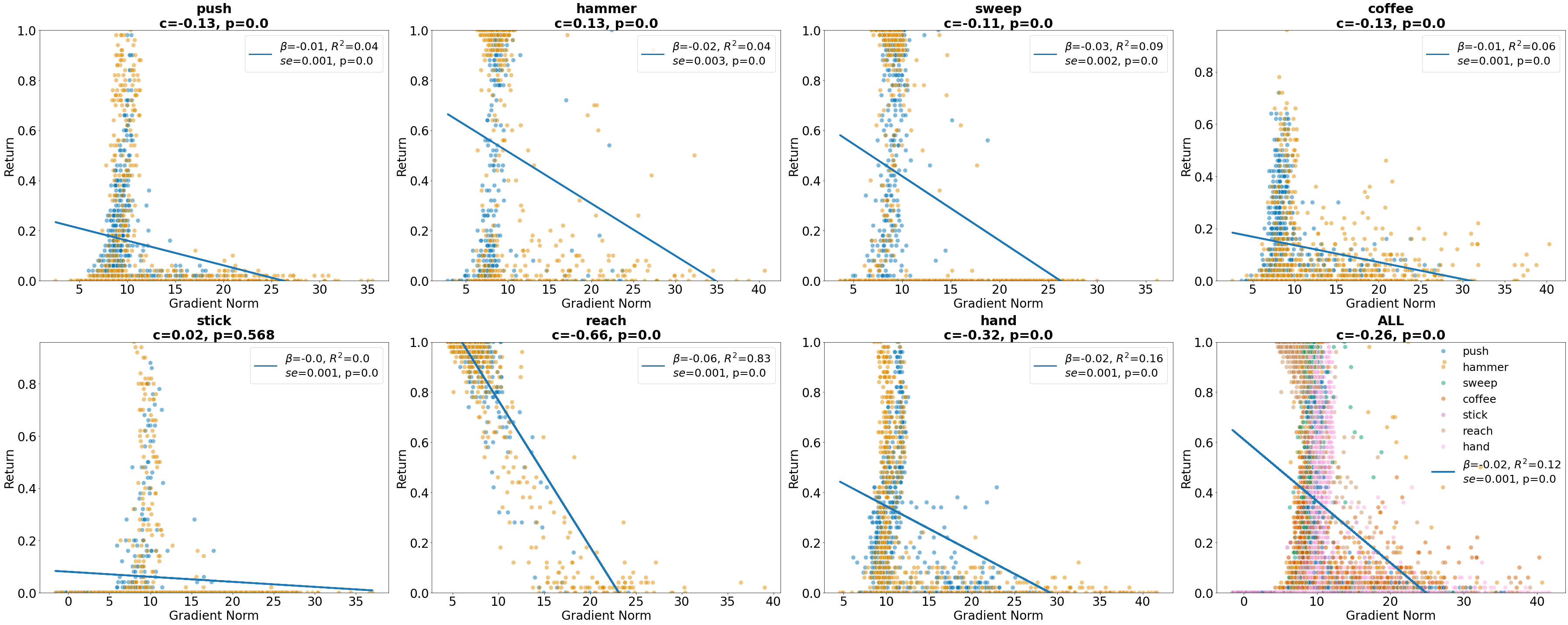}
        \end{subfigure}
    \end{minipage}
\caption{Gradient norm logarithm scatter plots with regression line for MW environments.}
\label{fig:mw_reg_gn}
\end{center}
\end{figure*}

For the gradient norm, we observe a higher value of this metric for runs with RR=16 both for DMC and MW benchmarks (Figures \ref{fig:dmc_reg_gn} and \ref{fig:mw_reg_gn}. There is consistency in coefficients for MW; however, in DMC, we observe that only four out of 7 environments exhibit negative signs of the coefficient. However, for most of the extremely high values of gradient norm, we systematically observe low returns regardless of the environment.

\begin{figure*}[ht!]
\begin{center}
    \begin{minipage}[h]{1.0\linewidth}
        \begin{subfigure}{1.0\linewidth}
            \includegraphics[width=0.99\linewidth]{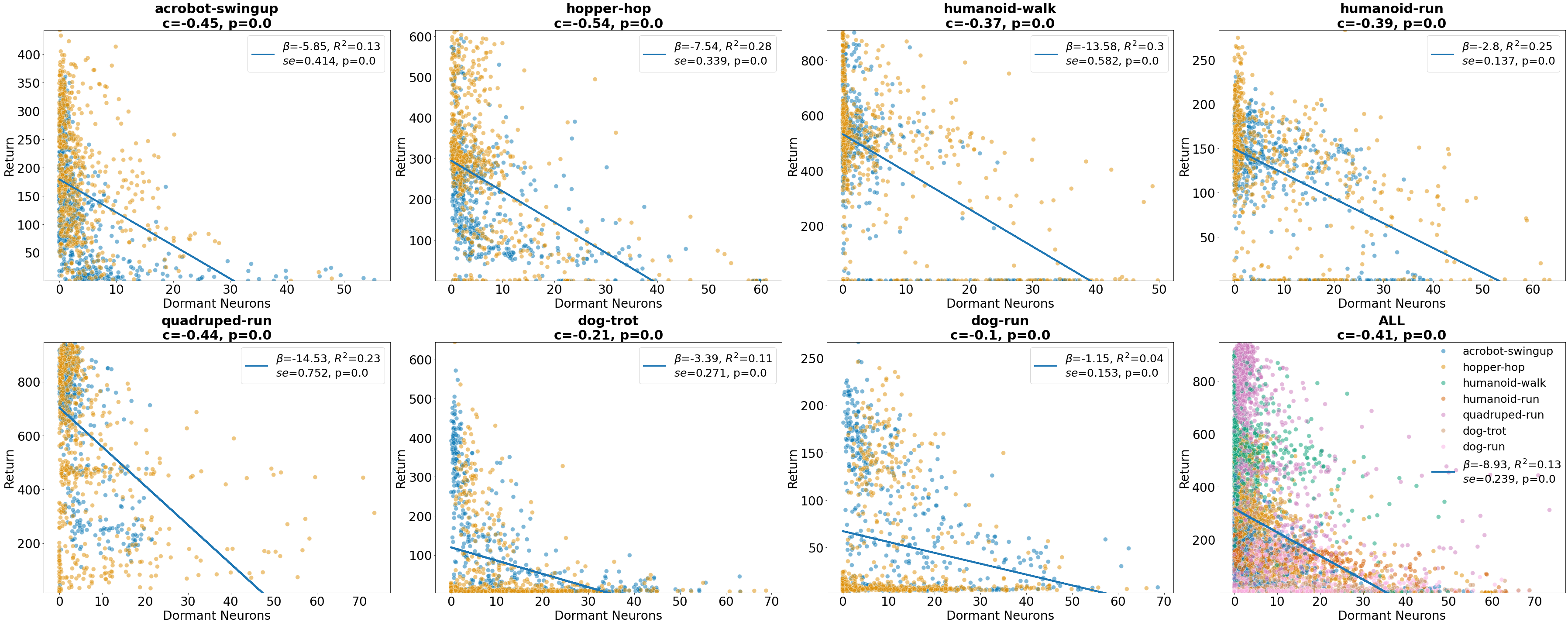}
        \end{subfigure}
    \end{minipage}
\caption{Dormant neurons scatter plots with regression line for DMC environments.}
\label{fig:dmc_reg_dn}
\end{center}
\end{figure*}

\begin{figure*}[ht!]
\begin{center}
    \begin{minipage}[h]{1.0\linewidth}
        \begin{subfigure}{1.0\linewidth}
            \includegraphics[width=0.99\linewidth]{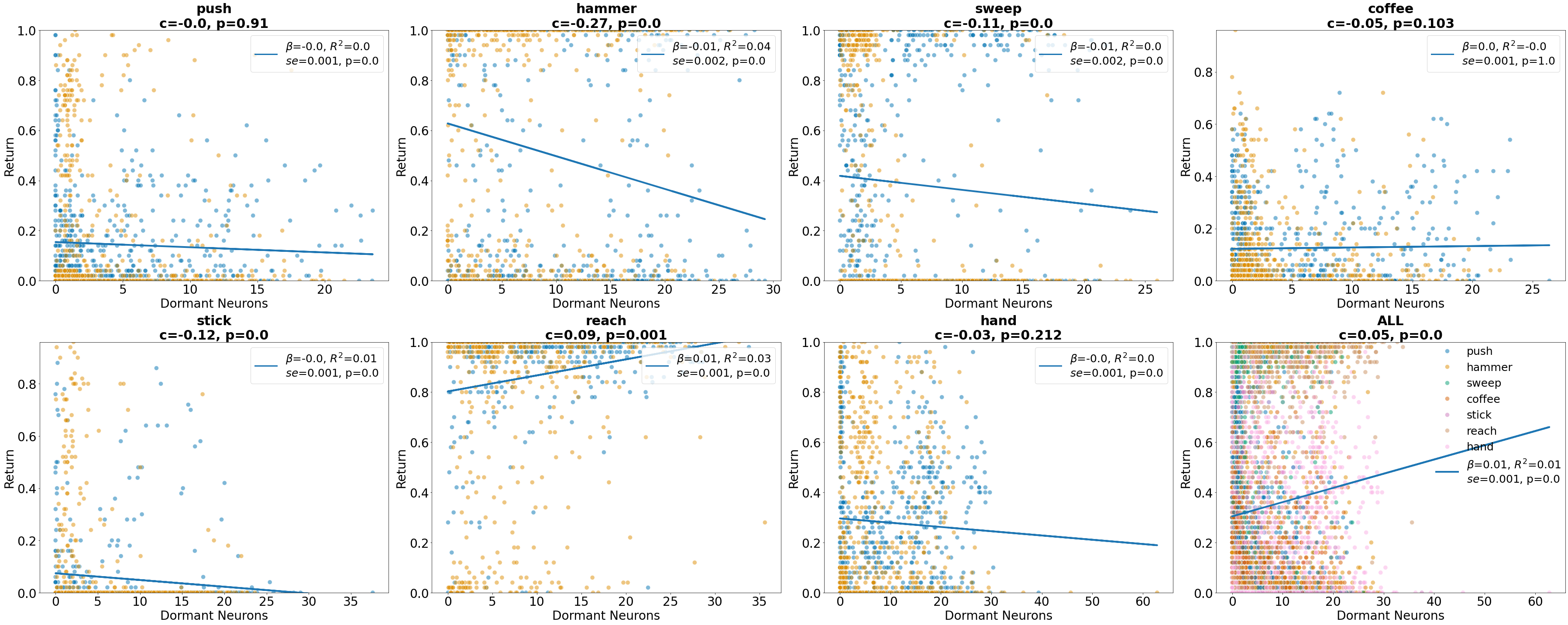}
        \end{subfigure}
    \end{minipage}
\caption{Dormant neurons scatter plots with regression line for MW environments.}
\label{fig:mw_reg_dn}
\end{center}
\end{figure*}

Dormant neurons, presented in Figures \ref{fig:dmc_reg_dn} and \ref{fig:mw_reg_dn}, clearly correlate negatively with the return, especially for DMC environments. However, there are cases, such as the Reach environment from MetaWorld, where a high percentage of dormant neurons benefit the agent, probably because this particular environment is especially easy.

\begin{figure*}[ht!]
\begin{center}
    \begin{minipage}[h]{1.0\linewidth}
        \begin{subfigure}{1.0\linewidth}
            \includegraphics[width=0.99\linewidth]{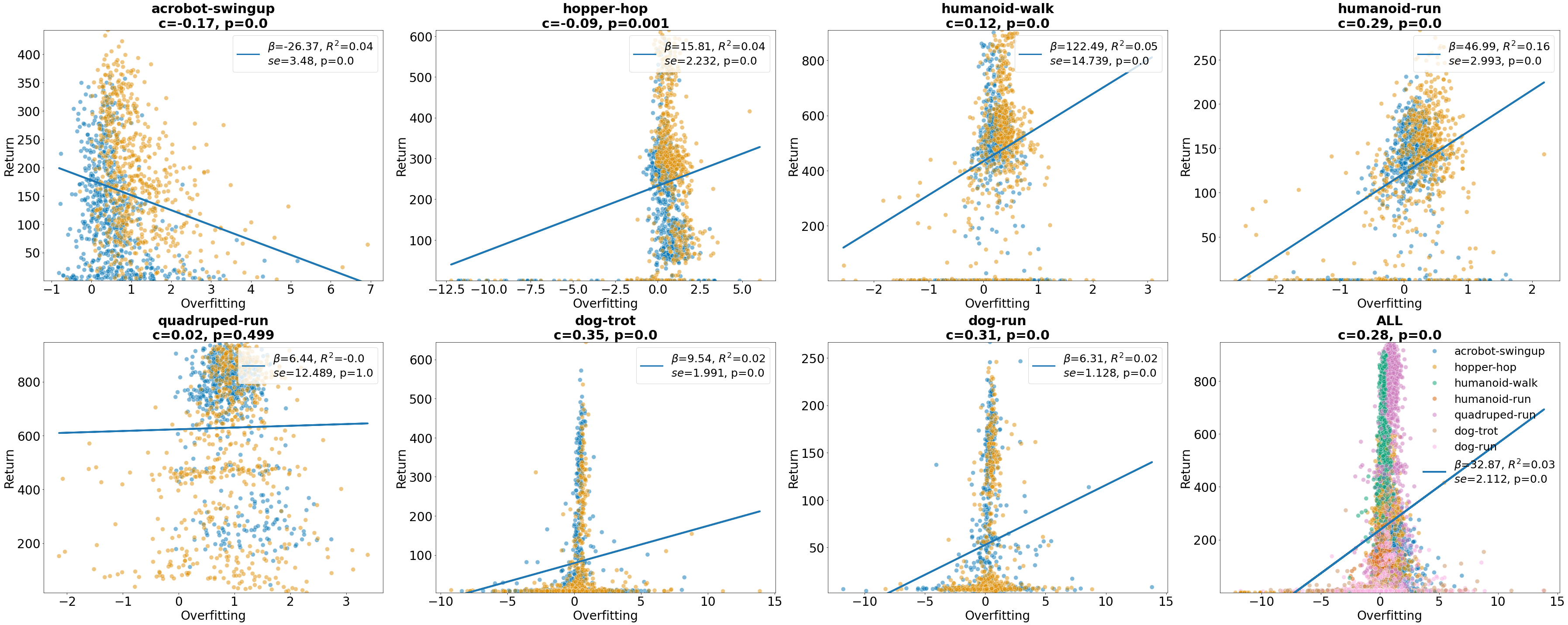}
        \end{subfigure}
    \end{minipage}
\caption{Overfitting logarithm scatter plots with regression line for DMC environments.}
\label{fig:dmc_reg_overf}
\end{center}
\end{figure*}

\begin{figure*}[ht!]
\begin{center}
    \begin{minipage}[h]{1.0\linewidth}
        \begin{subfigure}{1.0\linewidth}
            \includegraphics[width=0.99\linewidth]{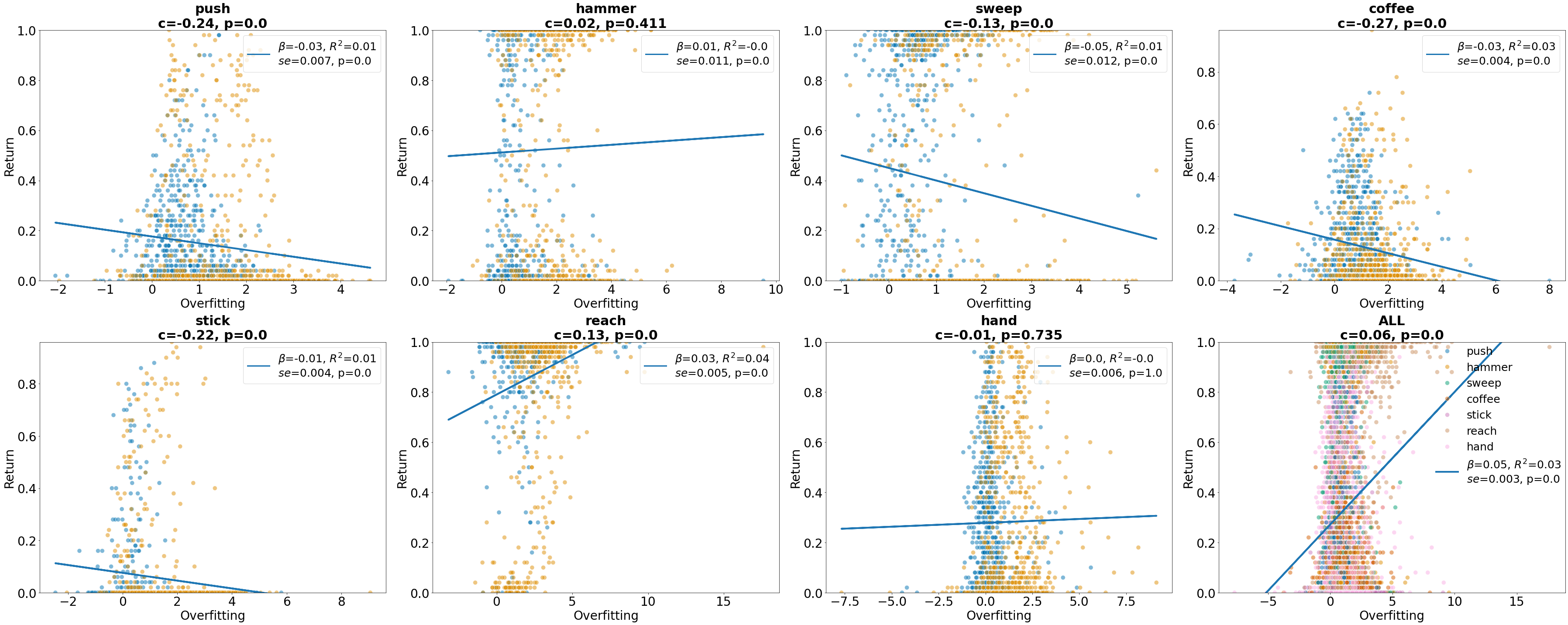}
        \end{subfigure}
    \end{minipage}
\caption{Overfitting logarithm scatter plots with regression line for MW environments.}
\label{fig:mw_reg_overf}
\end{center}
\end{figure*}

Overfitting is the metric that exhibits the highest variance in coefficient signs across environments and benchmarks. As shown in Figures \ref{fig:dmc_reg_overf} and \ref{fig:mw_reg_overf}, the best-performing runs are located for low absolute values of overfitting. In addition, RR=16 clearly results in higher overfitting values.

\subsection{Image-based DeepMind Control}

We test whether the results achieved on proprioceptive state representation transfer to image-based control. To this end, we run 4 versions of the DrQ agent \citep{yarats2020image}:

\begin{enumerate}
    \item Vanilla DrQ (DrQ)
    \item DrQ with layer normalization on the critic network (DrQ + LN)
    \item DrQ with full-parameter resets every 200k environment steps (DrQ + Res)
    \item DrQ with both normalization on the critic network and full-parameter resets every 200k environment steps (DrQ + LN + Res)
\end{enumerate}

We run these variations on 6 tasks from the DeepMind Control benchmark: Acrobot Swingup, Cheetah Run, Hopper Hop, Humanoid Run, Humanoid Stand, and Humanoid Walk. We run the humanoid tasks for 3mln frames and the other tasks for 1mln frames with a replay ratio of 1. We calculate the relationship between frames and environment steps according to the methodology presented in \citet{yarats2020image}. We present the results in the table below.

\begin{table}[h]
\centering
\caption{Final performance in image-based environments. 3 seeds per task.}
\vspace{0.1cm}
\label{table:size_explanation}
 \begin{tabular}{l | c | c | c | c } 
\toprule
 & \textbf{DrQ} & \textbf{DrQ + LN} & \textbf{DrQ + Res} & \textbf{DrQ + LN + Res} \\
 \midrule

  \textbf{Acrobot Swingup} & $172.9\pm22.1$ & $87.6 \pm 19.7$ & $52.3 \pm 11.2$ & $88.8 \pm 8.2$ \\
  \textbf{Cheetah Run} & $727.4\pm7.3$ & $680.0 \pm 2.3$  & $715.6 \pm 7.1$ & $653.9 \pm 2.1$ \\
  \textbf{Hopper Hop} & $74.6\pm34.9$ & $116.3 \pm 27.2$  & $135.7 \pm 24.1$ & $167.3 \pm 14.3$ \\
  \textbf{Humanoid Stand} & $7.8\pm0.2$ & $8.1 \pm 0.2$  & $7.5 \pm 0.4$ & $7.8 \pm 0.4$ \\
\bottomrule
 \end{tabular}
\end{table}   

Unfortunately, the humanoid agents were mostly unable to achieve non-random policies in the budget of 3mln frames. Interestingly, the proprioceptive results do not seem to directly transfer to image-based agents with a low-replay ratio. As such, we believe the image-based benchmark requires further studies.

\subsection{Best combinations of intervention performance plots}
\label{app:best_combination}

\subsection{Other}
\label{app:Other}

\begin{figure*}[ht!]
\begin{center}
\begin{minipage}[h]{1.0\linewidth}
    \begin{subfigure}{1.0\linewidth}
        \includegraphics[width=0.49\linewidth]{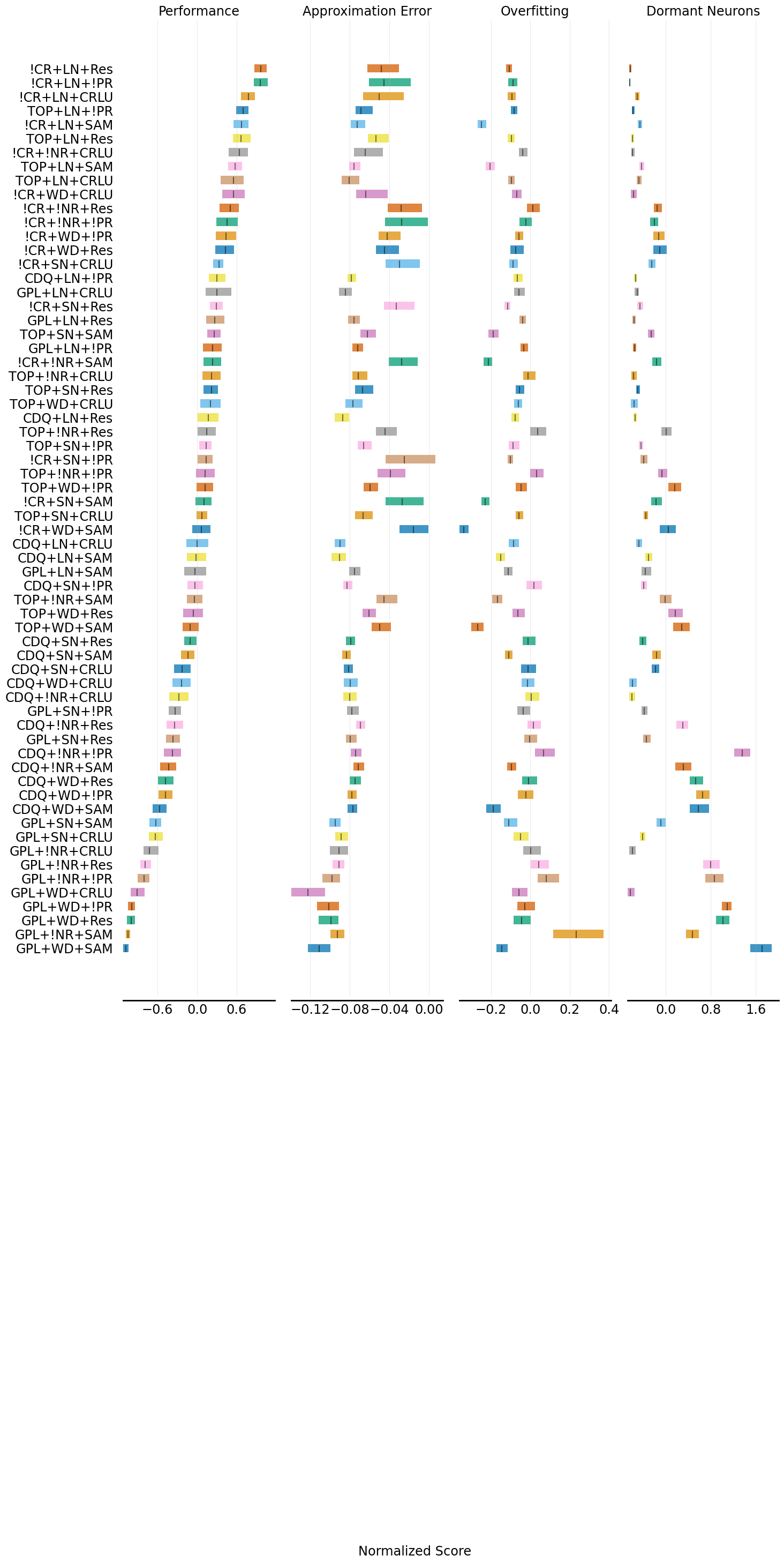}
    \hfill
        \includegraphics[width=0.49\linewidth]{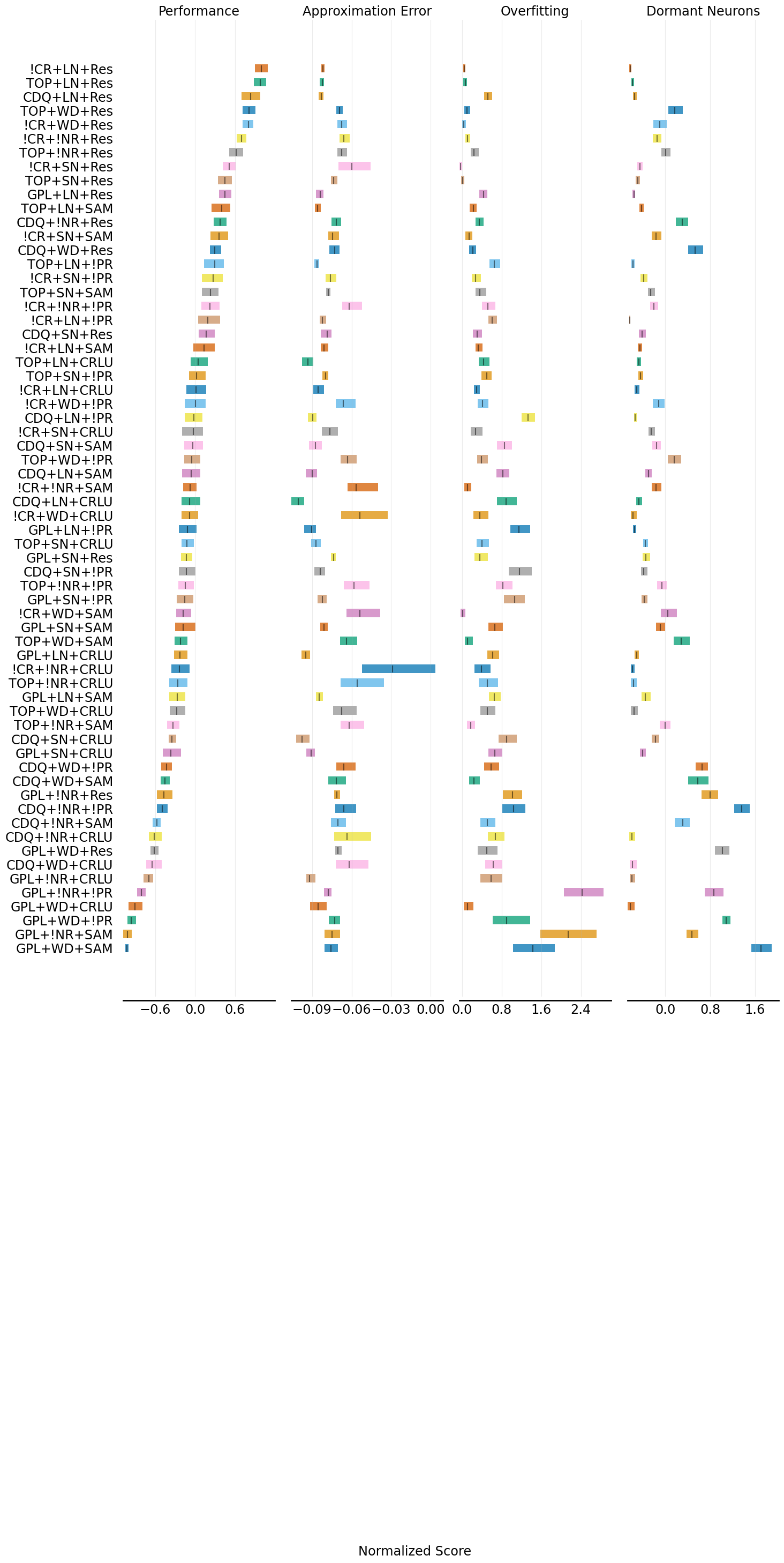}
    \end{subfigure}
\end{minipage}
\caption{Top performing configuration in the low (left) and high (right) replay regime. 10 seeds per task per algorithm.}
\label{fig:metrics}
\end{center}
\end{figure*}

\begin{figure}[ht!]
\centering
\begin{minipage}[h]{1.0\linewidth}
    \begin{subfigure}{1.0\linewidth}
        \includegraphics[trim={0 0 1.9cm 1.8cm},clip,width=0.48\linewidth]{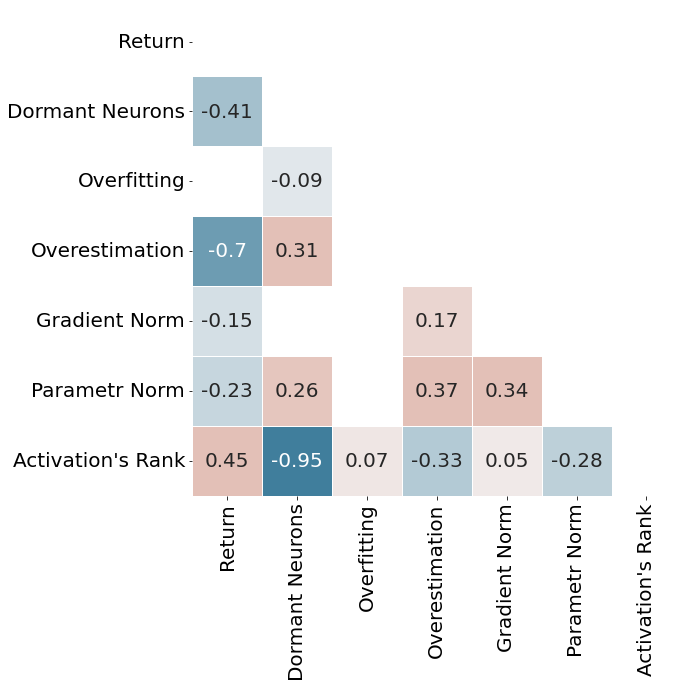}
    \hfill
        \includegraphics[trim={0 0 1.9cm 1.8cm},clip,width=0.48\linewidth]{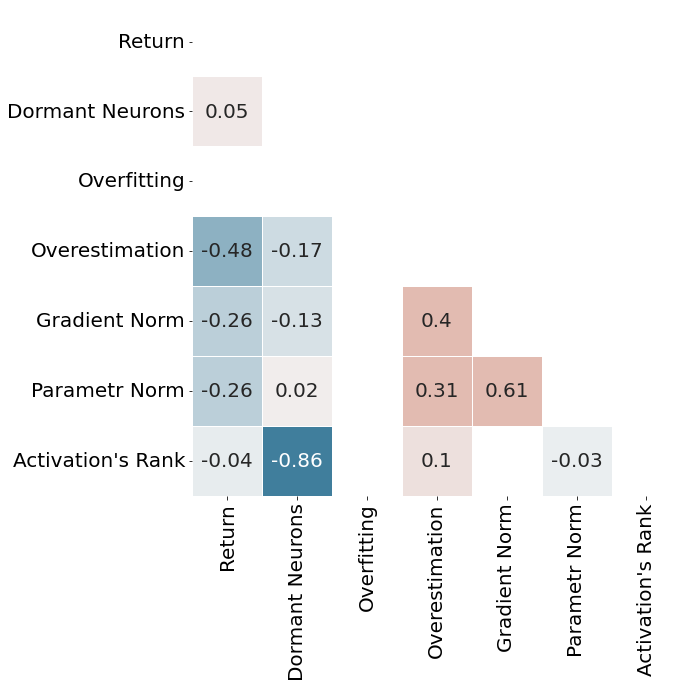}
    \end{subfigure}
\end{minipage}
\caption{Spearman correlation matrix for explanatory metrics on DMC benchmark (left) and on MetaWorld benchmark (roght plot). Blank spaces are correlations that do not meet the p-value.}
\label{fig:dmc_corr}
\end{figure}




\begin{figure}[ht!]
\centering
\begin{subfigure}{1\linewidth}
        \includegraphics[width=0.59\linewidth]{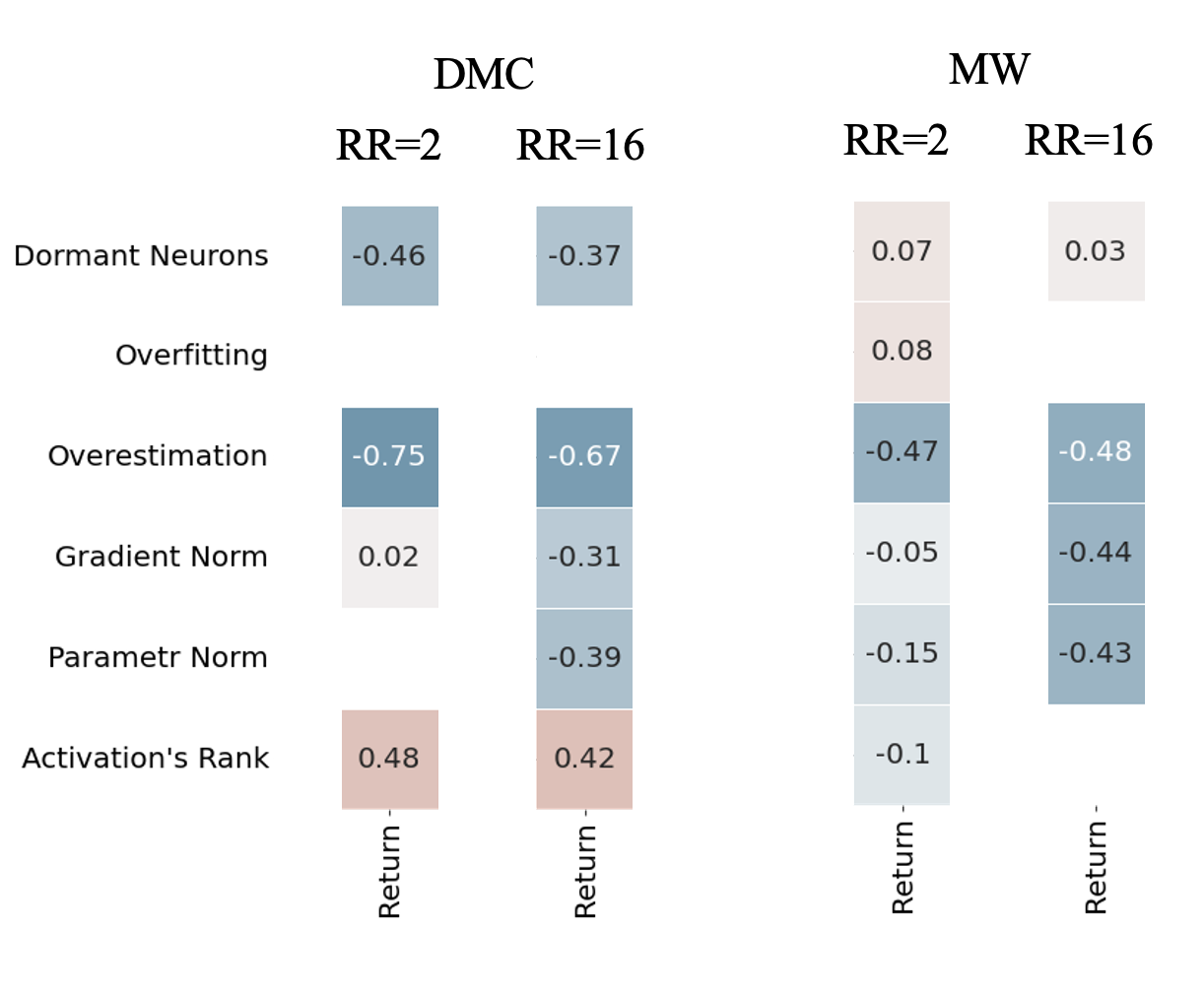}
\end{subfigure}
\caption{Spearman correlation for different replay ratios.}
\label{fig:rr_corr_both}
\end{figure}

\end{document}